\documentclass[letterpaper, 10 pt, journal, twoside]{ieeetran} 
\IEEEoverridecommandlockouts                              % This command is only
\author{First Author$^{1}$, Second Author$^{2}$, and Third Author$^{1}$%
\thanks{Manuscript received: Month, Day, Year; Revised Month, Day, Year;
Accepted Month, Day, Year.}%Use only for final RAL version
\thanks{This paper was recommended for publication by Editor Editor name upon
evaluation of the Associate Editor and Reviewers' comments. *Here you can
knowledge the organizations/grants which supported the work}%Use only for final
RAL version
\thanks{$^{1}$First Author and Third Author are with School of Engineering,
Robotics Department,
University of Somewhere, Someland
{\tt\small first.author@papercept.net}}%
\thanks{$^{2} $SecondAuthor is with School of Engineering, Automation
Department,
University of Anywhere, Anyland
{\tt\small second.author@papercept.net}}%
\thanks{Digital Object Identifier (DOI): see top of this page.}
}
\usepackage[utf8]{inputenc}
\usepackage[T1]{fontenc}
\usepackage{cite}
\usepackage{amsmath,amssymb,amsfonts}
\usepackage{algorithmic}
\usepackage{graphicx}
\usepackage{textcomp}
\usepackage{xcolor}
\usepackage{subfigure}
\usepackage{pifont}
\usepackage{cite}
\newcommand{\cmark}{\ding{51}}%
\newcommand{\xmark}{\ding{55}}%

\usepackage{colortbl}

\title{Real-time Fusion Network for RGB-D Semantic Segmentation Incorporating Unexpected Obstacle Detection for Road-driving Images}

\author{Lei Sun$^{1}$, Kailun Yang$^{2}$, Xinxin Hu$^{1}$, Weijian Hu$^{1}$ and Kaiwei Wang$^{3}$
\thanks{*This work has been partially funded through the AccessibleMaps project. This work has been supported by Hangzhou SurImage Technology Co., Ltd.}
\thanks{$^{1}$L. Sun, X. Hu and W. Hu are with State Key Laboratory of Modern Optical Instrumentation, Zhejiang University, China {\tt \{leo\_sun, hxx\_zju, huweijian\}@zju.edu.cn}}
\thanks{$^{2}$K. Yang is with Institute for Anthropomatics and Robotics, Karlsruhe Institute of Technology, Germany {\tt kailun.yang@kit.edu}}
\thanks{$^{3}$K. Wang is with National Optical Instrumentation Engineering Technology Research Center, Zhejiang University, China {\tt wangkaiwei@zju.edu.cn}}
}

\begin{document}

\maketitle
\thispagestyle{empty}
\pagestyle{empty}

%%%%%%%%%%%%%%%%%%%%%%%%%%%%%%%%%%%%%%%%%%%%%%%%%%%%%%%%%%%%%%%%%%%%%%%%%%%%%%%%
\begin{abstract}

Semantic segmentation has made striking progress due to the success of deep convolutional neural networks. Considering the demands of autonomous driving, real-time semantic segmentation has become a research hotspot these years. However, few real-time RGB-D fusion semantic segmentation studies are carried out despite readily accessible depth information nowadays.
In this paper, we propose a real-time fusion semantic segmentation network termed RFNet that effectively exploits complementary cross-modal information. Building on an efficient network architecture, RFNet is capable of running swiftly, which satisfies autonomous vehicles applications. Multi-dataset training is leveraged to incorporate unexpected small obstacle detection, enriching the recognizable classes required to face unforeseen hazards in the real world. A comprehensive set of experiments demonstrates the effectiveness of our framework. On \textit{Cityscapes}, Our method outperforms previous state-of-the-art semantic segmenters, with excellent accuracy and 22Hz inference speed at the full 2048$\times$1024 resolution, outperforming most existing RGB-D networks.

\end{abstract}
% Keywords appear just beneath the abstract. Use only for final RAL version.

%%%%%%%%%%%%%%%%%%%%%%%%%%%%%%%%%%%%%%%%%%%%%%%%%%%%%%%%%%%%%%%%%%%%%%%%%%%%%%%%
\section{INTRODUCTION}

% Safety is always the most important issue in autonomous driving, and unexpected road hazards like debris, bricks, stones and cargos have always been dangerous factors that cannot be ignored. According to AAA foundation for Traffic Safety, debris on the road led to more than 200,000 crashes on U.S. roadways between 2011 and 2014, resulting in approximately 39,000 injuries and more than 500 deaths~\cite{roaddebris}. Therefore, intelligent vehicles and robots have to detect and avoid these small obstacles. These obstacles are generally small in size but not fixed in shape and type, making detecting them a challenging subject that has aroused interest among the robotics and computer vision community.
% 原来的第一段

\IEEEPARstart{E}{nvironment} perception is a significant task for intelligent robots and systems in object classification, autonomous driving, and localization.
In recent years, this field has witnessed remarkable progress thanks to deep Convolutional Neural Networks (CNNs) based semantic segmentation methods\cite{long2015fully}\cite{chen2017deeplab}\cite{chen2018encoder}. As an environment perception method to be applied in autonomous driving, safety, accuracy, and efficiency are the vital factors in semantic segmentation for upper-level navigational tasks.
However, unexpected road hazards like debris, bricks, stones, and cargos become the most dangerous and difficult elements to detect in autonomous driving imagery. 
According to the AAA Foundation for Traffic Safety, debris on the road led to more than 200,000 crashes on U.S. roadways between 2011 and 2014, resulting in approximately 39,000 injuries and more than 500 deaths~\cite{roaddebris}.
These obstacles are generally small in size but not fixed in shape and type, making detecting them a challenging subject that has aroused interest among the robotics and computer vision community. For these reasons, it is desirable to develop a semantic segmentation based method incorporating pixel-wise unexpected obstacle detection.
%第一段修改的较多，主要想法是把文章重心拉回到ss上，文章的方法是基于ss的小障碍物检测
%杨总看看intro逻辑上连接

\begin{figure}
    \centering
    \subfigure[SwiftNet]{
    \centering
    \begin{minipage}[b]{0.48\textwidth}
    \includegraphics[width=1.0\textwidth]{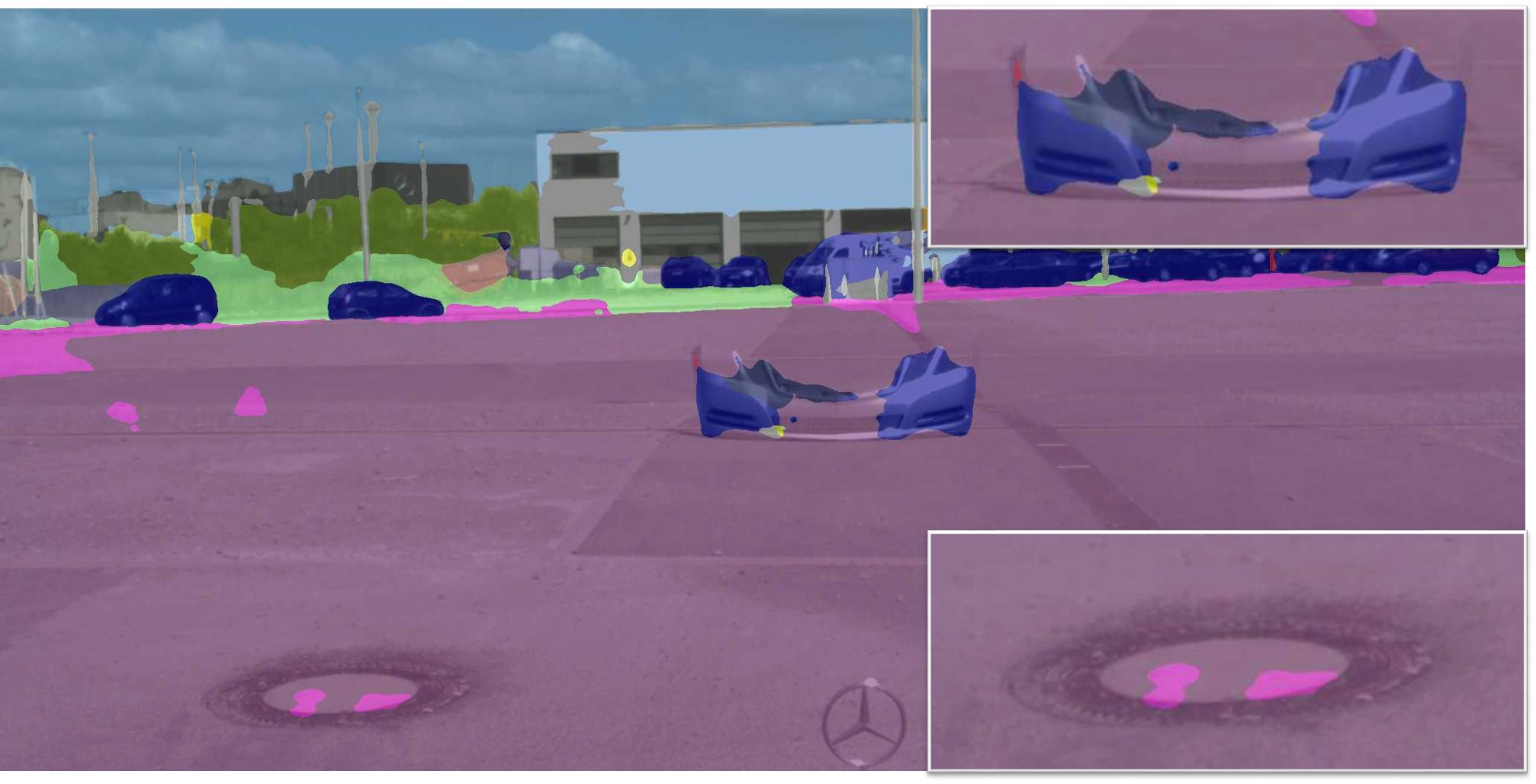}
    \end{minipage}
    }
    %\subfigure[SwiftNet with multi-dataset training]{
    %\centering
    %\begin{minipage}[b]{0.5\textwidth}
    %\includegraphics[width=0.9\textwidth]{./images/first/Swiftnet_multidataset/big.png}
    %\end{minipage}
    %}
    \subfigure[RFNet]{
        \centering
        \begin{minipage}[b]{0.48\textwidth}
        \includegraphics[width=1.0\textwidth]{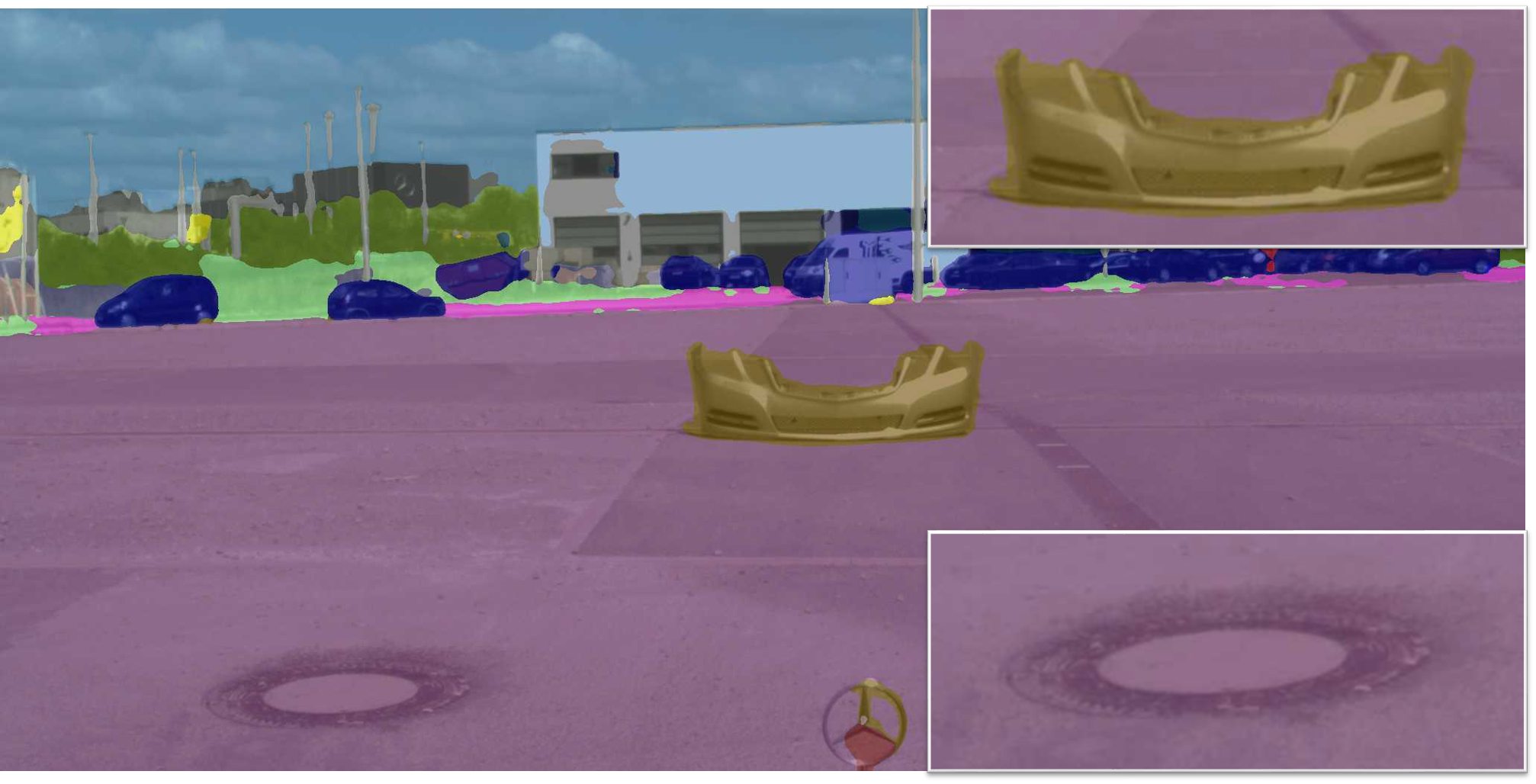}
        \end{minipage}
    }
    \caption{Examples from the \textit{Lost and Found dataset} and corresponding results of the methods: (a) SwiftNet (unexpected obstacle wrongly classified as car), (b) The proposed RFNet (clear and consistent segmentation).} 
    \label{fig:1}
    
\end{figure}

Compared to expensive 3D sensors like LIDAR, RGB camera is a much lower cost solution with higher resolution. Based on RGB stereo camera, there have been some attempts to detect small obstacles with the help of geometry cues and CNNs~\cite{pinggera2016lost}, but only relying on apparent information in the RGB image alone is not sufficient for obstacle detection~\cite{gupta2018mergenet}. For example, manhole covers, and small obstacles can both cause gradient changes in the image. The traversable areas and obstacles in the depth map vary vastly in depth maps. Depth maps contain more location and contour information that can be used as a critical indicator of objects in real-world driving scenarios. In this sense, appropriately combining of appearance and depth is promising to improving the performance~\cite{gupta2018mergenet}\cite{hu2019acnet}\cite{ramos2017detecting}. But most accuracy-oriented RGB-D semantic segmentation works focus on indoor scenes~\cite{hu2019acnet}\cite{hazirbas2016fusenet}\cite{jiang2018rednet}, without assuring a fast inference speed that is necessary for autonomous vehicles.

On the other hand, the outstanding capacity of CNNs is based on a large amount of annotated data, especially for semantic segmentation tasks~\cite{sun2019see}. Current mainstream autonomous driving datasets generally assume only some fixed categories of objects in the scene, ignoring unforeseen hazards like unexpected small obstacles in the real world. For instance, \textit{Cityscapes}~\cite{Cordts2016Cityscapes} only divides objects to 19 classes, without defining any unexpected class. Multi-source training has been proven to effectively increase recognizable semantics without having to relabel the dataset~\cite{meletis2018training}. However, previous multi-source training frameworks have only considered the heterogeneity in the label hierarchies of RGB data, missing the opportunity to leverage complementary depth information from different sources.

In this paper, we propose a framework that combines RGB-D semantic segmentation and obstacle detection. RFNet, a real-time fusion network for RGB-D semantic segmentation is elaborately designed. With our multi-dataset training strategy, our framework is able to classify 19 categories in \textit{Cityscapes} incorporating pixel-wise unexpected small obstacle detection (see Figure \ref{fig:1}). An extensive set of experiments shows the effectiveness and efficiency of the proposed framework for the semantic segmentation task. The main contributions of our work are threefold:

\begin{itemize}
\item We propose RFNet, a real-time fusion network for RGB-D semantic segmentation incorporating detection of unexpected obstacle, which achieves higher accuracy with fast inference compared to other state-of-the-art methods on the \textit{Cityscapes} dataset.
\item Depth complementary features are efficiently extracted in the proposed network, which improves the accuracy compared to the single RGB-stream architecture.
\item Multi-dataset training and the depth stream in the architecture enable the network to work remarkably effective in detecting unexpected small objects.
\end{itemize}

\section{RELATED WORKS}

\subsection{RGB-D Semantic Segmentation}

High-quality dense depth maps from depth sensors like Kinect and RealSense 
boost the development of indoor semantic segmentation. 
Early attempt like~\cite{couprie2013indoor} simply concatenated RGB and depth channels as a four-channel input, and fed it into a conventional RGB modal network. However, such method can not exploit complementary information from depth maps in most times~\cite{ngiam2011multimodal}. Wang et al.~\cite{depthawarecnn} introduced depth-aware CNN which augmented conventional CNN with a depth similarity term, but it only works well with dense depth maps. Schneider et al.~\cite{schneider2017multimodal} designed a lightweight depth branch with GoogLeNet~\cite{szegedy2015going} and explored different points for merging the depth and RGB networks.
In FuseNet~\cite{hazirbas2016fusenet} and RedNet~\cite{jiang2018rednet}, RGB images and depth maps are fed into two separate neural network branches respectively, which are fused before the upsampling. In~\cite{gupta2014learning}, depth maps are pre-processed as HHA features that encode horizontal disparity, height above ground and angle. Park et al.~\cite{park2017rdfnet} proposed a multilevel feature fusion scheme by introducing multi-modal feature fusion to the RefineNet blocks. ACNet~\cite{hu2019acnet} achieved a breakthrough by proposing an attention complementary module to exploit complementary depth information efficiently. 
These studies prove that RGB-D semantic segmentation can achieve better segmentation results than single RGB-based methods. The major reason for this is that compared to the single RGB images, depth maps contain more location and contour information that benefit the context-critical semantic segmentation. 

Compared to indoor scene depth maps from Kinect or RealSense, outdoor traffic scene depth maps are much more sparse. Li et al.~\cite{li2017traffic} simply stacked smoothed depth maps with RGB images as a 4-channel input.
Based on VGG~\cite{vgg}, Kreso et al.~\cite{krevso2016convolutional} introduced a scale selection layer and used the depth maps as a guidance to produce a scale-invariant representation to free appearance from the scale.
In~\cite{hung2019incorporating}, luminance information is used for depth map enhancement. Most recently, Deng et al.~\cite{deng2019rfbnet} proposed a Residual Fusion Block (RFB) to formulate the interdependencies of the encoders to extract cross-modal features based on ERFNet~\cite{romera2017erfnet}. Low latency is crucial in autonomous driving applications, but most of these methods cannot meet the real-time constraint. In this paper, we propose a real-time fusion network to achieve swift inference while retaining a highly competitive performance among the state of the art for RGB-D segmentation.

\subsection{Unexpected Obstacle Detection for Self-driving Cars}

Detecting unexpected small but potentially hazardous obstacles on the road is a vital task for autonomous driving, and this subject has always been a research hotspot. Generally these methods for detecting and localizing generic obstacles are based on stereo cameras integrated on self-driving cars. Among these methods, most are based on the generic geometric criteria. The Stixel algorithm \cite{pfeiffer2011towards} represents obstacles with a set of rectangular vertical obstacle segments, providing a robust representation of the 3D scene. Geometric point cluster methods like \cite{manduchi2005obstacle} and \cite{broggi2011stereo} exploit geometric relation between 3D points to detect and cluster obstacle points.

Because of the superiority in making use of visual appearance and context of images, CNNs are adopted in contemporary researches. Ramos et al.~\cite{ramos2017detecting} presented a principled Bayesian framework to fuse the semantic segmentation predicted from a convolutional neural network and stereo-based detection results from the Fast Direct Planar Hypothesis Testing (FPHT) method. MergeNet~\cite{gupta2018mergenet} was proposed with a multi-stage training procedure involving weight sharing, separating learning of low and high level features from the RGB-D input and a refining stage which learns to fuse the obtained complementary features. But all these methods can only predict three main classes: free-space, obstacle, and background. To meet the demands of autonomous driving, we need a more universal approach that can enrich the detectable semantics beyond simple roads/obstacles separation. In this work, we address unexpected obstacle detection by incorporating it in a multi-source semantic segmentation framework to provide a unified pixel-wise scene understanding.

\section{METHODOLOGY}

\subsection{Network Architecture}
\begin{figure}[t]
    \centerline{\includegraphics[width = 0.5\textwidth]{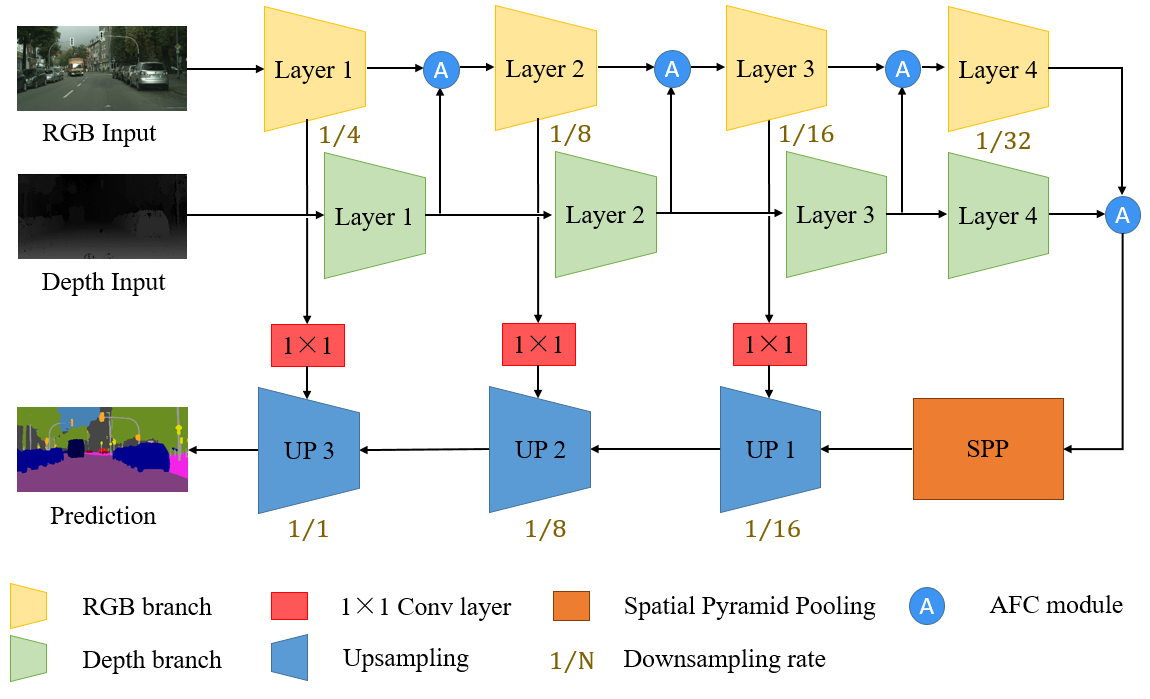}}
    \caption{Overview of RFNet: the proposed network architecture for real-time fusion-based RGB-D semantic segmentation.}
    \label{fig:RFNet}
\end{figure}

The entire network architecture of RFNet is shown in Figure~\ref{fig:RFNet}.
In the encoder part of the architecture, we design two independent branches to extract features for RGB and depth images separately—RGB branch as the main branch and Depth branch as the subordinate branch. In both branches, we choose ResNet-18~\cite{he2016identity} as the backbone to extract features from inputs because ResNet-18 has moderate depth and residual structure, and its small operation footprint is compatible with real-time operation. After each layer of ResNet-18, the output features from Depth branch are fused to RGB branch after the Attention Feature Complementary (AFC) module. The spatial pyramid pooling (SPP) block gathers the fused RGB-D features from two branches and produces feature maps with multi-scale information. Finally, referred to SwiftNet~\cite{orsic2019defense}, we design the efficient upsampling modules to restore the resolution of these feature maps with skip connections from the RGB branch.
%蓝色句想表达的意思是RGB是主分支，Depth是辅助分支
%depth branch 中Depth要不要大写？

%RFNet网络结构图

\begin{figure}[t]
    \centerline{\includegraphics[width = 0.5\textwidth]{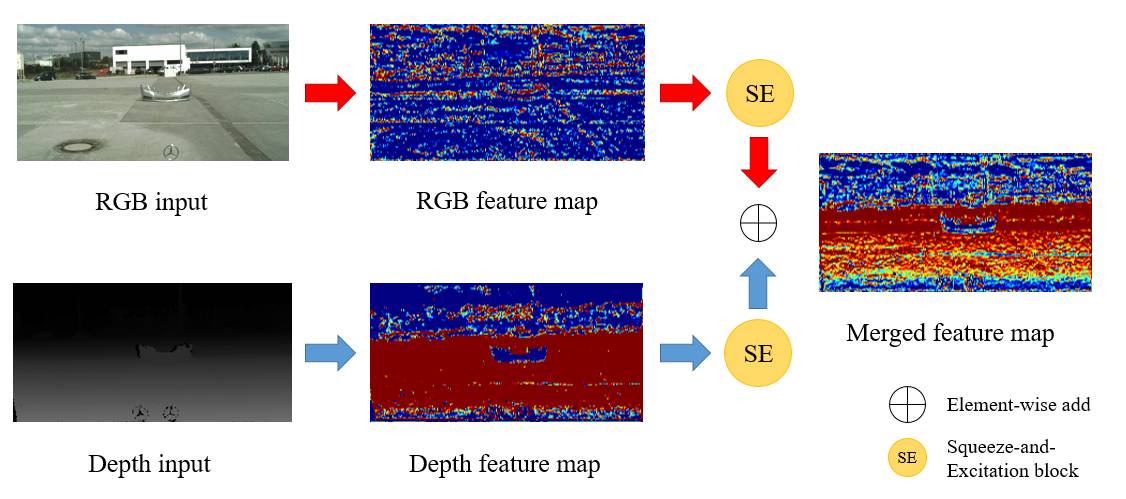}}
    \caption{AFC: Attention Feature Complementary module to exploit cross-model information from RGB and Depth inputs.}
    \label{fig:AFC}
\end{figure}

\textbf{RGB-D fusion module.}
As discussed in the last part, the depth maps contain more contour and location information that benefit RGB semantic segmentation. In order to fuse RGB and depth information effectively, we design an RGB-D fusion module termed Attention Feature Complementary (AFC) module (shown in Figure~\ref{fig:AFC}) to make the network focus on learning more complementary informative features from RGB and Depth branches. As shown in Figure~\ref{fig:AFC}, in the AFC module, we leverage a SE block~\cite{hu2018squeeze} as the channel attention method. SE block can learn to use global information to emphasize informative channels and suppress less useful channels, which helps the AFC module exploit informative features from both branches effectively.

With the multi-branch architecture, we have the RGB input feature maps $X=\left[ {{x}_{1}},\ldots ,{{x}_{C}} \right]\in {{\mathbb{R}}^{C\times H\times W}}$ and depth input feature maps $Y=\left[ {{y}_{1}},\ldots ,{{y}_{C}} \right]\in {{\mathbb{R}}^{C\times H\times W}}$. First we use global average pooling as a channel descriptor based on channel attention mechanism, then we add a 1$\times$1 convolution layer with the same channels as input. This 1$\times$1 convolution layer is able to excavate correlations between channels. The followed sigmoid function is applied to activate the convolution result and constrain the value of the weight vector between 0 and 1. Next, we do outer product for the weight vector and input feature maps in both branches. Finally, by adding results from RGB branch and Depth branch, we have the resulted feature map $Z\in {{\mathbb{R}}^{C\times H\times W}}$, expressed as:

\begin{equation}
    Z=X\otimes {{\sigma }_{\text{1}}}\left[ {{\phi }_{\text{1}}}\left( X \right) \right]+Y\otimes {{\sigma }_{2}}\left[ {{\phi }_{2}}\left( Y \right) \right]
    \label{AFC}
\end{equation}

Here, $\phi$ denotes global pooling and 1$\times$1 convolution. $\otimes$ and $\sigma$ denote outer product and sigmoid function respectively. By applying such attention mechanism in RGB-D fusion, more informative features obtain higher values of weights, which helps us exploit complementary information from depth maps more effectively.

% \textbf{Spatial pyramid pooling.}
% In order to increase the receptive field to cover pixels of large objects while maintaining real-time speed, we adopt Spatial Pyramid Pooling (SPP) to average features over aligned grids with different granularities. Our SPP module is based on a simplified version of pyramidal pooling module from PSPNet~\cite{zhao2017pyramid}. PSPNet uses it to provide multi-scale context and to condition features for linear classification with only one convolution layer between the SPP and the predictions. In our case, we upsample the feature pyramid to the original resolution and concatenate them with input features, following by convolutions to obtain output feature maps. In RFNet, SPP features must pass through the whole decoder before getting classified. Thus, it acts as an instrument to enlarge the receptive field.

%经过四个AFC module之后，融合后的特征图富含高层语义信息. In order to increase the receptive field to cover pixels of large objects while maintaining real-time speed, referred to ~, we adopt Spatial Pyramid Pooling (SPP) to average features over aligned grids with different granularities. 
After four ResNet blocks and AFC module, the fused feature maps contain rich high-level semantic information. In order to increase the receptive field to cover pixels of large objects while maintaining a real-time speed, referred to ~\cite{orsic2019defense}~\cite{ERF-PSPNet}~\cite{zhao2017pyramid}, we adopt Spatial Pyramid Pooling (SPP) to average features over aligned grids with different granularities before the upsampling.

\textbf{Efficient upsampling module.}
The purpose of the decoder is to upsample semantically rich visual features in coarse spatial resolution to the input resolution. We adopt a simple decoder that contains three simple upsampling modules with skip connections from the encoder. In the first two upsampling modules, low-resolution feature maps from the former block are upsampled with bilinear interpolation to the same resolution as feature maps from skip connection, then these two streams of feature maps are element-wisely added and finally mixed with a 3$\times$3 convolution. The third upsampling module is slightly different because we add a convolution layer and a 4-times bilinear interpolation at last to restore to the same resolution as the input. More precisely, the skip connection is routed before the second ReLU of the residual block because the current study shows that skip connection from any other stage impairs the accuracy~\cite{orsic2019defense}.

\subsection{Multi-Dataset Learning}
\begin{figure}[ht]
    \centering%改为单边
    \subfigure[RGB]{
    \begin{minipage}[b]{0.32\linewidth}
    \includegraphics[width=1\linewidth]{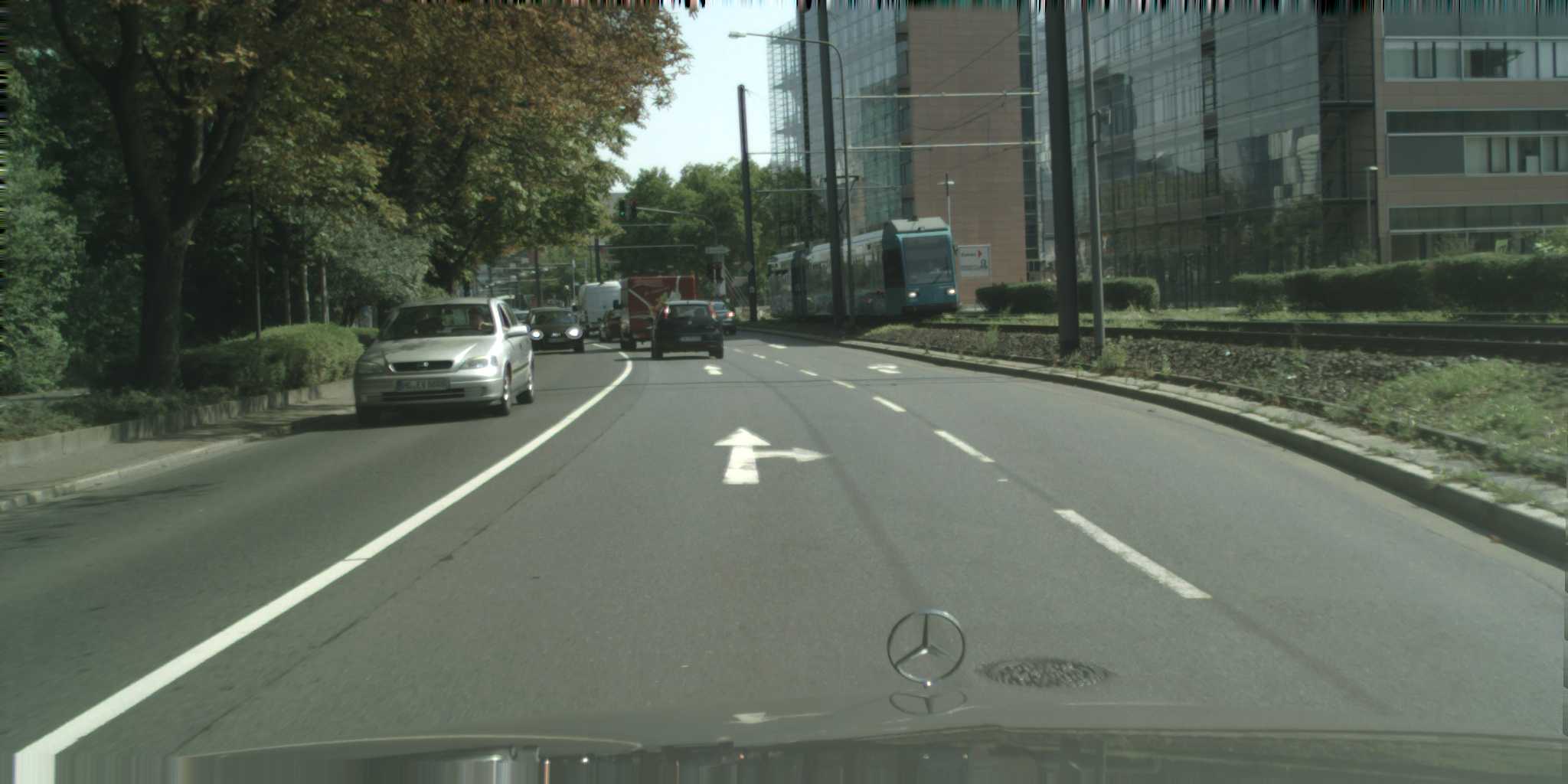}\vspace{0pt}  %控制垂直距离
    \includegraphics[width=1\linewidth]{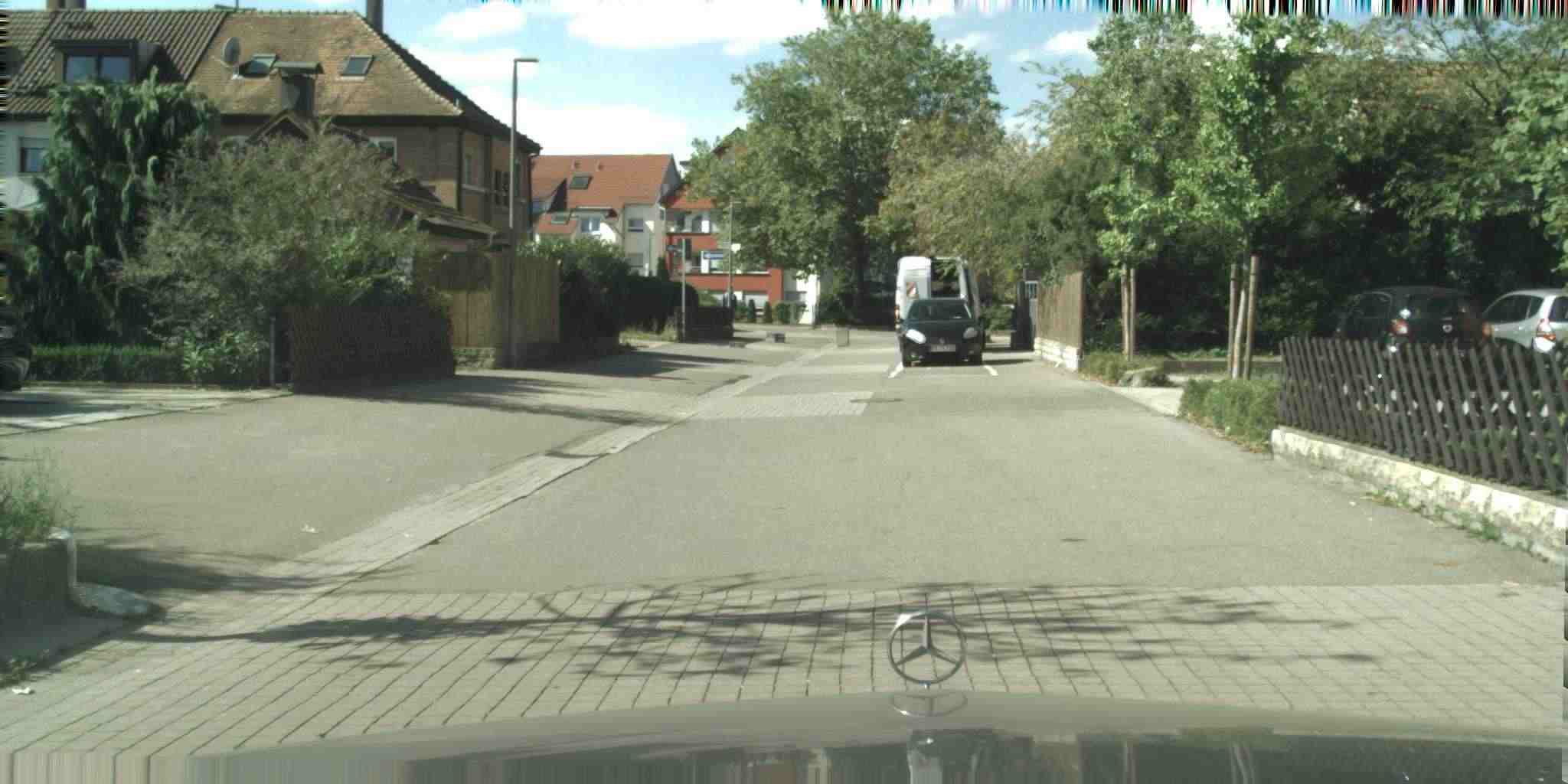}
    \end{minipage}}\hspace{-6pt}  % 控制subfigure之间横向距离
    \subfigure[GT]{
    \begin{minipage}[b]{0.32\linewidth}
    \includegraphics[width=1\linewidth]{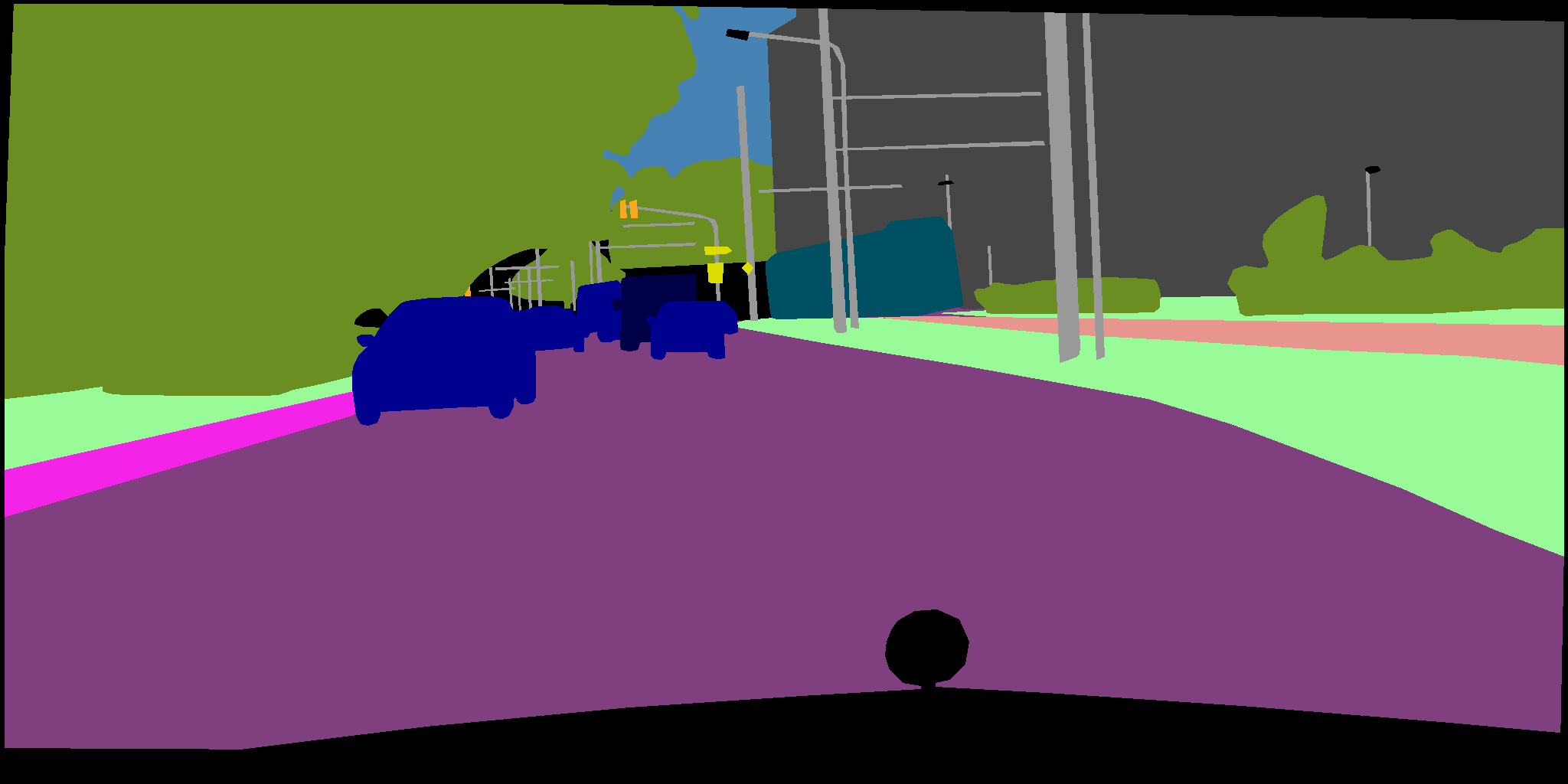}\vspace{0pt}  %控制垂直距离
    \includegraphics[width=1\linewidth]{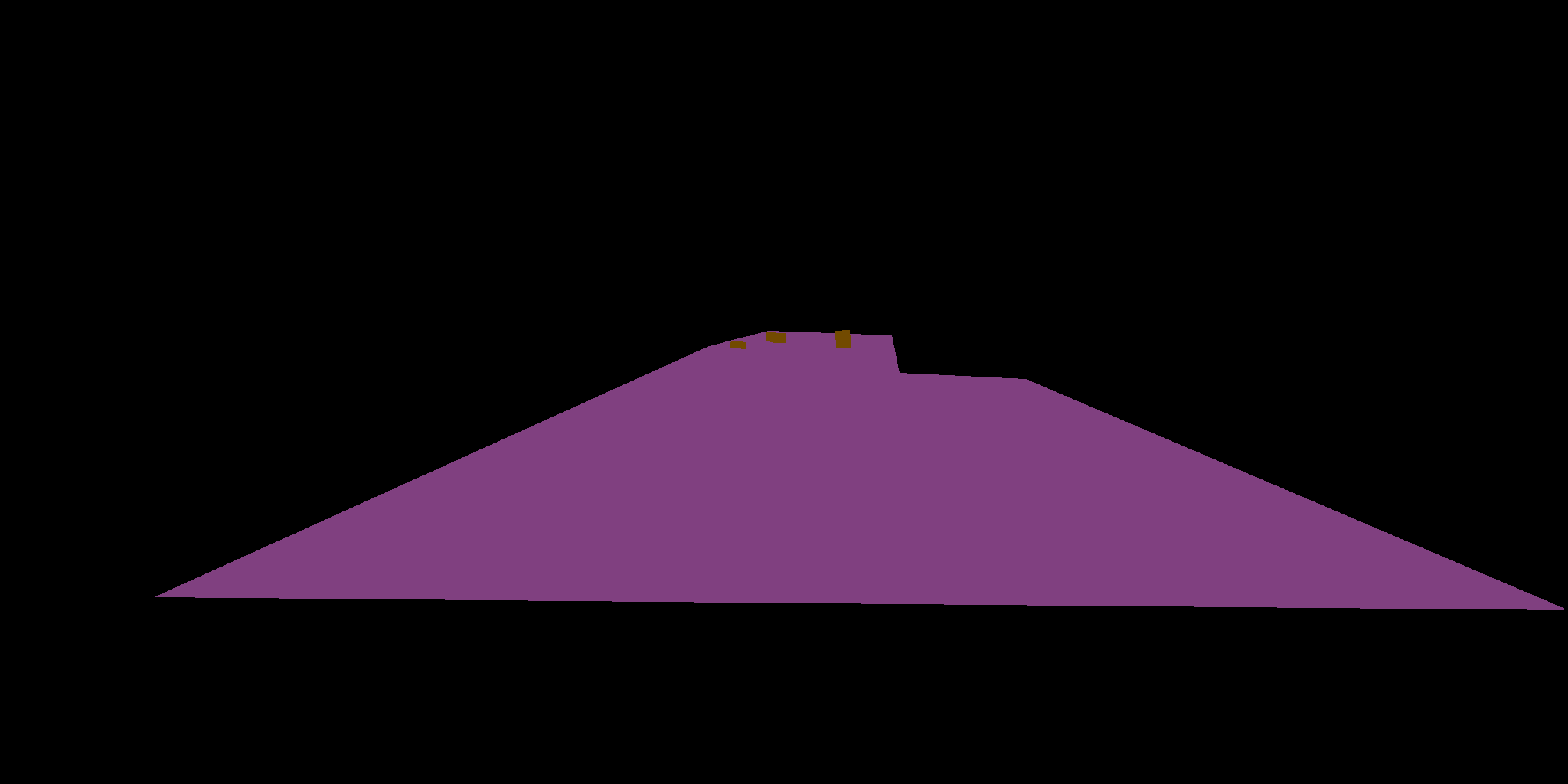}
    \end{minipage}}\hspace{-6pt}
    \subfigure[Result]{
    \begin{minipage}[b]{0.32\linewidth}
    \includegraphics[width=1\linewidth]{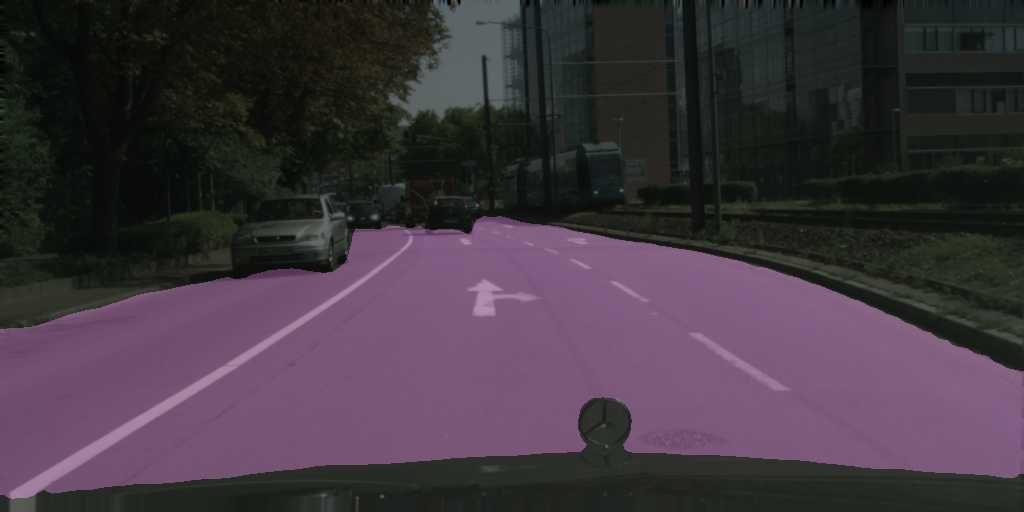}\vspace{0pt}  %控制垂直距离
    \includegraphics[width=1\linewidth]{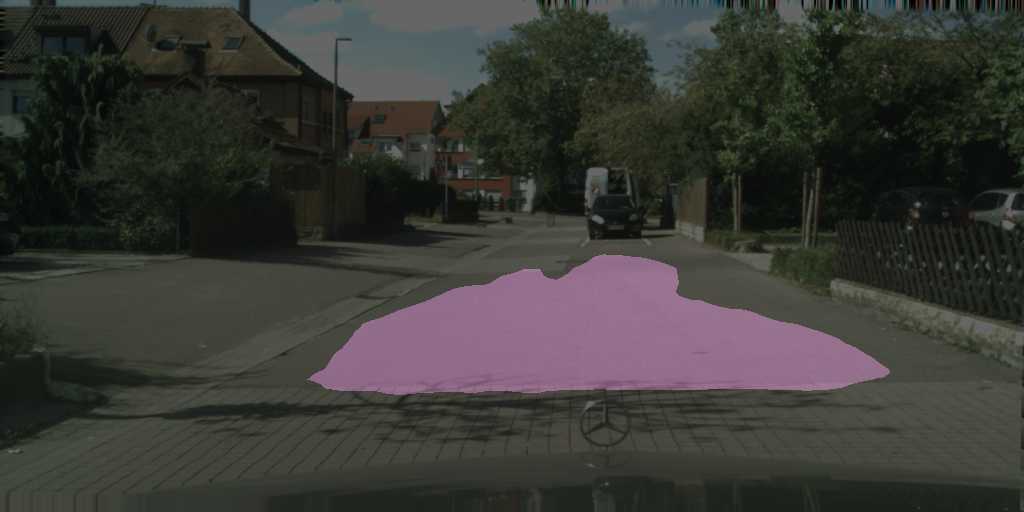}
    \end{minipage}}\hspace{-6pt}
    % \caption[mix directly]
    \caption{RGB images and ground truth from \textit{Cityscapes} (first row) and \textit{Lost and Found} (second row) respectively. The last column shows the inference result if simply training on two datasets without consideration of the heterogeneity in the annotation style.}
    \label{fig:mix_directly}
\end{figure}

As a data-driven technology, annotated labels are essential for semantic segmentation, but we can not annotate all classes in the real world. In order to utilize as much and diverse training data as possible and increase the number of recognizable classes from a few dozens to virtually anything that a scene can contain, multi-source learning is an effective method.
However, simply mixing two or more datasets for training may cause some problems.
As shown in Figure~\ref{fig:mix_directly}, the heterogeneity in the annotation type and sample amount may cause overfitting to one of the data sources, leading to incomplete segmentation when simply mixing the datasets. 

This is because classes in different datasets may conflict with each other. For example, the annotation type of these classes is different, or a certain class is a subclass of a class in another dataset. To facilitate multi-source learning with such heterogeneity, we design some training strategies. Formally, we have datasets ${{D}_{1}},\ldots ,{{D}_{c}},\ldots ,{{D}_{n}}$. Class set $A$ contains classes that do not conflict with each other, and class set $B$ contains the rest. For these conflicted classes, we refer to dataset ${{D}_{c}}$ as a standard annotation.
Let us denote an image by $x$, and the corresponding human annotation for $x$ is provided and denoted by $y$, where $y(m,n) \in {1,\ldots,C}$ is the label of pixel $x(m,n)$, and $C$ is the total number of classes. $l$ denotes the total number of images in all datasets and $\phi$ denotes the segmentation model.
We train on the joint multi-datasets with the loss function shown below:

\begin{equation}
    loss=\frac{1}{l}\sum\limits_{i=1}^{l}{\left[ {{L}_{A}}\left( \phi \left( {{x}_{i}} \right),{{y}_{i}} \right)+\lambda {{L}_{B}}\left( \phi \left( {{x}_{i}} \right),{{y}_{i}} \right) \right]}
\end{equation}

% \[
%     loss=\frac{1}{l}\sum\limits_{i=1}^{l}{\left[ {{L}_{A}}\left( \phi \left( {{x}_{i}} \right),{{y}_{i}} \right)+\lambda {{L}_{B}}\left( \phi \left( {{x}_{i}} \right),{{y}_{i}} \right) \right]} \\ 
% \]
% \[
%     \lambda = \{ 
%     \begin{matrix}
%        1 & {{x}_{i}}\in {{D}_{c}}  \\
%        0 & {{x}_{i}}\notin {{D}_{c}}  \\
%     \end{matrix}
% \]

where ${{L}_{A}}\left( .,. \right)$ and ${{L}_{B}}\left( .,. \right)$ denotes cross entropy loss function for class set $A$ and $B$ respectively. $\lambda$ is a hyper-parameter balancing the weights of different classes, and in our work we set $\lambda$ as following:

\begin{equation}
    \lambda =\left\{ \begin{matrix}
        1, & {{x}_{i}}\in {{D}_{c}}  \\
        0, & {{x}_{i}}\notin {{D}_{c}}  \\
     \end{matrix} \right. 
\end{equation}

% \begin{equation}

% \end{equation}

% \begin{equation}
%     \begin{align}
%         & loss=\frac{1}{l}\sum\limits_{i=1}^{l}{\left[ {{L}_{A}}\left( \phi \left( {{x}_{i}} \right),{{y}_{i}} \right)+\lambda {{L}_{B}}\left( \phi \left( {{x}_{i}} \right),{{y}_{i}} \right) \right]} \\ 
%        & \lambda =\left\{ \begin{matrix}
%          1 & {{x}_{i}}\in {{D}_{c}}  \\
%          0 & {{x}_{i}}\notin {{D}_{c}}  \\
%       \end{matrix} \right. \\ 
%     \end{align}
%     \label{loss}
% \end{equation}

%这里的类别名也应该大写吧
For instance, in this work we leverage \textit{Cityscapes}~\cite{Cordts2016Cityscapes} and \textit{Lost and Found}~\cite{pinggera2016lost}. Note that \textit{Cityscapes} has annotations on 19 classes except for those unexpected small obstacles. We make some modifications to the loss function while training. The road class in \textit{Cityscapes} and small obstacle class in \textit{Lost and Found} do not conflict with classes in other datasets, which belong to class set $A$. The rest classes in \textit{Cityscapes} are conflicted with the background class in \textit{Lost and Found}, so they are divided into class $B$. In this situation, we assume \textit{Cityscapes} as standard dataset ${{D}_{c}}$. In the training stage, background class and free-space in \textit{Lost and Found} should not be counted in the loss function. In our situation, the ignorance of background class makes coarse-annotated free-space class in \textit{Lost and Found} helpful for improving the training data amount, so we also include free-space in the final loss.
%这里想表达的意思是忽略background class后free-space有了利用的意义，可以增加数据量
With the presented multi-dataset training strategy, our RFNet learns to predict 19 classes from \textit{Cityscapes} and the critical unexpected small obstacle class from the \textit{Lost and Found} dataset.

%以下解释为什么固定的标签可以学到探测 unexpected obstacle
%我觉得可以放到别的地方比如intro
Although unexpected small obstacle is a generalized conception, which is not limited to obstacle types in \textit{Lost and Found}, the definition of this particular set of classes allows us to meet the demand by exploiting the power of deep learning methods. For example, learning that all kinds of obstacles have some common contextual property, being of small dimensions and surrounded at least partly by free-space. Thereby, the network is able to generalize far beyond its training data with the multi-source learning strategy when facing innumerable possible corner cases.

\section{EXPERIMENTS}

\subsection{Datasets}
In this work, two RGB-D semantic segmentation datasets: \textit{Cityscapes} and \textit{Lost and Found} are exploited.
 
\textit{Cityscapes}~\cite{Cordts2016Cityscapes} is a large-scale RGB-D dataset that focuses on semantic understanding of urban street scenes. It contains 2975/500/1525 images in the training/validation/testing subsets, both with finely annotated labels on 19 classes. The images cover 50 different cities with a full resolution of 2048$\times$1024.
 
The \textit{Lost and Found}~\cite{pinggera2016lost} dataset consists of 2014 annotated frames from 112 stereo video sequences, along with coarse annotations of free-space areas and fine-grained annotations of the small obstacles on the road. Among them, training set and validation set contain 814 and 1200 images with a resolution of 2048$\times$1024, covering different small obstacles present at long distance with non-uniform road textures/appearances and pathways with many non-obstacle class objects acting as distractors.
 
Both disparity images from \textit{Cityscapes} and \textit{Lost and Found} are obtained by using the semi-global matching algorithm\cite{hirschmuller2005accurate}, which is a sophisticated method for the estimation of a dense disparity map from a rectified stereo image pair.

\subsection{Implementation Details}

The models were implemented on a single 2080Ti GPU with CUDA 10.0, CUDNN 7.6.0, and PyTorch 1.1. Adam~\cite{kingma2014adam} is used for optimization with the learning rate set to 4$\times$10$^{-4}$, where cosine annealing learning rate scheduling policy~\cite{loshchilov2016sgdr} is adopted to adjust learning rate with a minimum value of 1$\times$10$^{-6}$ in the last epoch. The weight decay is set to 1$\times$10$^{-4}$. We initialize the ResNet-18 in both RGB branch and Depth branch with pre-trained weights from ImageNet~\cite{russakovsky2015imagenet}, and initialize the rest part of the model with kaiming initialization~\cite{he2015delving}. More precisely, we average the weights for RGB inputs to match the shape of one-channel depth image in the Depth branch, as research works~\cite{hu2019acnet}~\cite{schneider2017multimodal} show that RGB pre-trained weights also boost depth image feature extraction. For pre-trained parameters, we update them with a 4 times smaller learning rate and apply 4 times smaller weight decay. Because the left and bottom part of the disparity images are not applicable due to the restrictions of semi-global matching algorithm, we crop these pixels and resize images back to the original resolution with bilinear upsampling. The rest of the data augmentation operations consist of scaling with random factors between 0.5 and 2, random horizontal flipping, and random cropping with an output resolution of 768$\times$768. We train all the models for 200 epochs with a batch size of 8.

\subsection{Results and Analysis}

\begin{table*}[htbp]
    \centering
    \caption{Performance of RFNet on the Cityscapes and Lost and Found validation set with different design choices.}
    \label{ablationtab1}
    \begin{tabular}{l|clll|ll}
    
    \textbf{Method}                  & \multicolumn{1}{l}{\textbf{RGB-D Fusion}} & \textbf{Dual-branch}    & \textbf{Concatenation} & \textbf{Element-wise add} & \textbf{mIoU(\%)} & \textbf{Params} \\ \hline \hline
    Single RGB                       &                                           &                           &                                    &                                       &        69.20\%     & 12.17M     \\
    RGB-D-Stack                      & \checkmark                                     &                           &                                    &                                       &        65.20\%  & 12.17M        \\ \hline
    RGB-D-Fusion (concatenation)      & \checkmark                                     & \multicolumn{1}{c}{\checkmark} & \multicolumn{1}{c}{\checkmark}          &                                       &        68.67\%    & 25.08M          \\
    RGB-RGB-Fusion (element-wise add) & \multicolumn{1}{l}{}                      & \multicolumn{1}{c}{\checkmark} &                                    & \multicolumn{1}{c|}{\checkmark}            &      69.37\%        & 23.69M         \\ \hline
    RFNet (without multi-dataset training strategy)   & \checkmark   & \multicolumn{1}{c}{\checkmark} &     & \multicolumn{1}{c|}{\checkmark}    &   53.83\%   & 23.69M       \\ 
    RFNet   & \checkmark   & \multicolumn{1}{c}{\checkmark} &     & \multicolumn{1}{c|}{\checkmark}    &   72.22\%   & 23.69M       \\ 
    \end{tabular}
\end{table*}

\begin{table*}[htbp]
    %\footnotesize
    \scriptsize
    \centering
    \caption{Per-class IoU(\%) results of three networks on the blended validation set of Cityscapes and Lost and Found Dataset. List of classes(from left to right): Road, Sidewalk, Building, Wall, Fence, Pole, Traffic Light, Traffic Sign, Vegetation, Terrain, Sky, Pedestrian, Rider, Car, Truck, Bus, Train, Motorbike, Bicycle and Small Obstacle.}
    \label{benchmarking}
    \begin{tabular}{m{5.3em}<{\centering}|m{1.3em}<{\centering} m{1.3em}<{\centering} m{1.3em}<{\centering} m{1.3em}<{\centering} m{1.3em}<{\centering} m{1.3em}<{\centering} m{1.3em}<{\centering} m{1.3em}<{\centering} m{1.3em}<{\centering} m{1.3em}<{\centering} m{1.3em}<{\centering} m{1.3em}<{\centering} m{1.3em}<{\centering} m{1.3em}<{\centering} m{1.3em}<{\centering} m{1.3em}<{\centering} m{1.3em}<{\centering} m{1.3em}<{\centering} m{1.3em}<{\centering} m{1.5em}<{\centering}| m{1.8em}<{\centering}}
    %\begin{tabular}{l|cccccccccccccccccccc|c}
    % \hline

    \textbf{Network} & \textbf{Roa} & \textbf{Sid} & \textbf{Bui} & \textbf{Wal} & \textbf{Fen} & \textbf{Pol} & \textbf{TLi} & \textbf{TSi} & \textbf{Veg} & \textbf{Ter} & \textbf{Sky} & \textbf{Ped} & \textbf{Rid} & \textbf{Car} & \textbf{Tru} & \textbf{Bus} & \textbf{Tra} & \textbf{Mot} & \textbf{Bic} & \textbf{SOb} & \textbf{mIoU} \\ \hline \hline
    ERF-PSPNet                            & 89.3         & 65.1         & 82.1         & 43.4         & 39.8         & 46.9         & 48.2         & 46.1         & 86.4         & 45.2         & 88.4         & 68.5         & 57.2         & 90.1         & 55.6         & 62.1         & 65.6         & 56.9         & 64.8  & 60.3&63.1       \\
    SwiftNet                              & 95.7         & 60.6         & 89.1         & 50.9         & 53.6         & 56.9         & 61.1         & 71.4         & 90.7         & 55.0         & 92.2         & 75.2         & 58.5         & 92.7         & 65.3         & 81.3         & 70.0         & 56.2         & 72.1 &62.8&70.6        \\ \hline
    Our RFNet                           & 96.0         & 60.6         & 90.8         & 50.2         & 59.9         & 60.0         & 62.6         & 72.8         & 91.1         & 57.3         & 92.5         & 76.1         & 57.9         & 93.3         & 73.8         & 82.3         & 73.2         & 54.0         & 72.7 &67.9&72.2\\
    % \hline
    \end{tabular}
\end{table*}

\textbf{Ablation Study.}
We perform the ablation study on our RFNet to explore the influence of different architecture variants and fusion schemes on the network accuracy where the results are shown in Table \ref{ablationtab1}. Results in this section are obtained by evaluating on the blended validation set of \textit{Cityscapes} and \textit{Lost and Found}, which includes all images from both validation datasets. All backbones in these models are initialized with ImageNet pre-trained weights.
%这里的数据都是基于两个数据集的val set混合在一起的验证集

In the table, the single RGB method only exploits the RGB branch of RFNet. Here, compared to SwiftNet~\cite{orsic2019defense}, the only difference is the SE block after each block of the ResNet-18. It is a control group to determine if the depth information helps improve the accuracy, which achieves a mean Intersection over Union (mIoU) of 69.20\%. In the RGB-D-Stack method, we stack depth maps with respective RGB images to form a 4-channel input to the single branch of RFNet. The low accuracy of the method (65.20\% in mIoU) proves that depth information is not exploited effectively in this way. We also design RGB-D-Fusion (concatenation), where the only difference of this method to the RGB-D-Fusion (element-wise add) in our RFNet is that RGB feature maps and depth feature maps are concatenated to a higher dimension feature maps and restore to the original dimension after a 1$\times$1 convolution. Results show that this method (68.67\%) performs clearly worse than RFNet (72.22\%). This is because in a compact network like RFNet, concatenation is a more inefficient way to make use of the depth information.

To eliminate the cause that more parameters in two-branch RFNet make it perform better, we design and train the RGB-RGB-Fusion method. The difference of the RGB-RGB-Fusion method to RFNet is that inputs are duplicate RGB images instead of RGB-D images, and after each AFC module, the element-wise added feature maps are divided by 2. The accuracy (69.37\%) is much lower than RFNet and approximately the same as the single RGB method, proving the benefit of fusion in RFNet is not simply owing to the increased parameters.
We also perform an experiment to explore the influence of the proposed multi-dataset training strategy. It turns out that without multi-dataset training strategy, our RFNet gets nearly 20\% lower IoU because of the class conflictions in two datasets. Finally the proposed RFNet with the proposed multi-dataset training strategy achieves a mIoU of 72.22\%, which is significantly better than the baseline (single RGB architecture) and other fusion-based variants, demonstrating the effectiveness of our fusion scheme bridged by the designed attention complementary modules.

\begin{figure}[ht]
    \centering
    \subfigure[RGB]{
    \begin{minipage}[b]{0.235\linewidth}
    \includegraphics[width=1\linewidth]{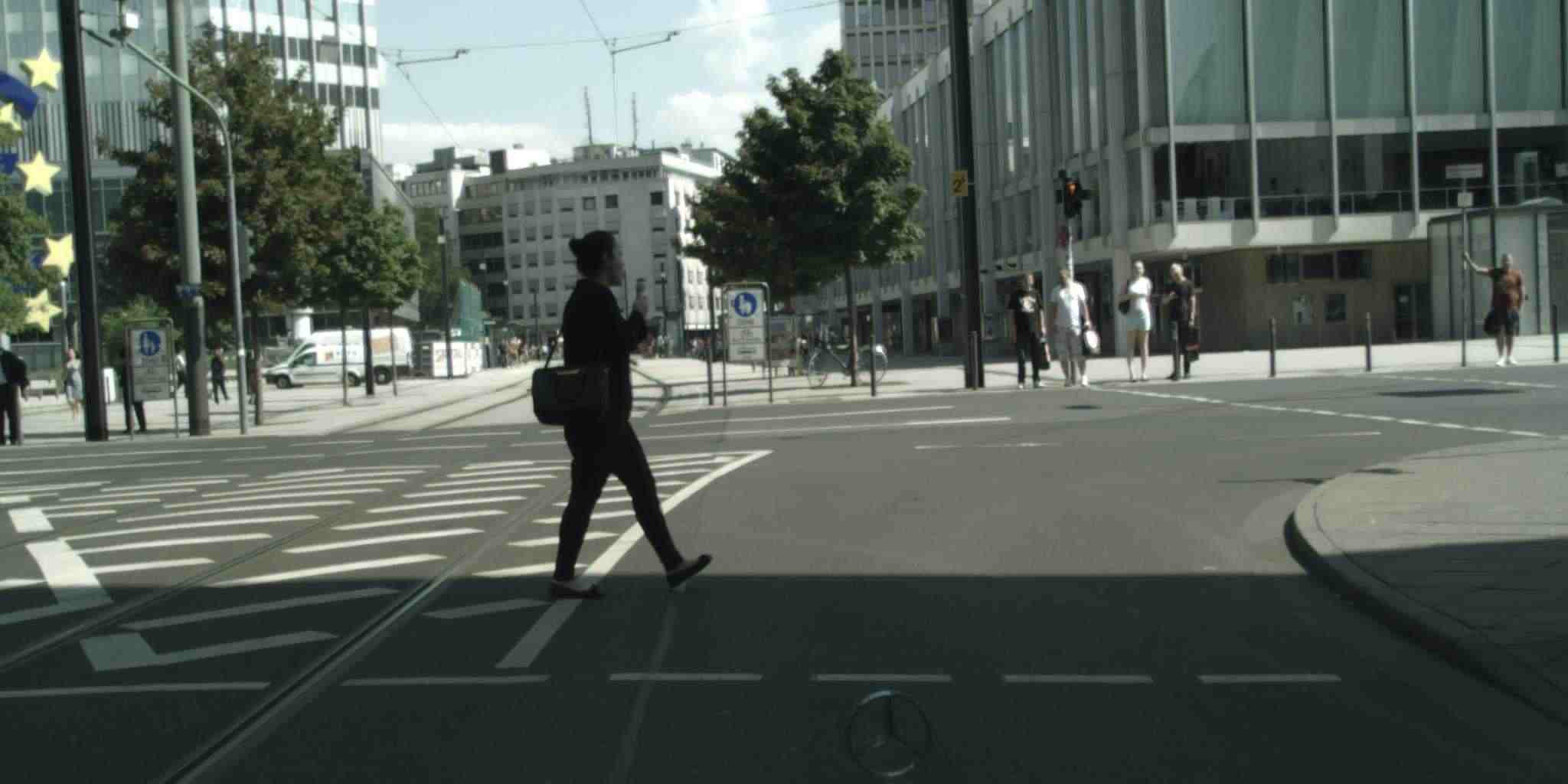}\vspace{0pt}  %控制垂直距离
    \includegraphics[width=1\linewidth]{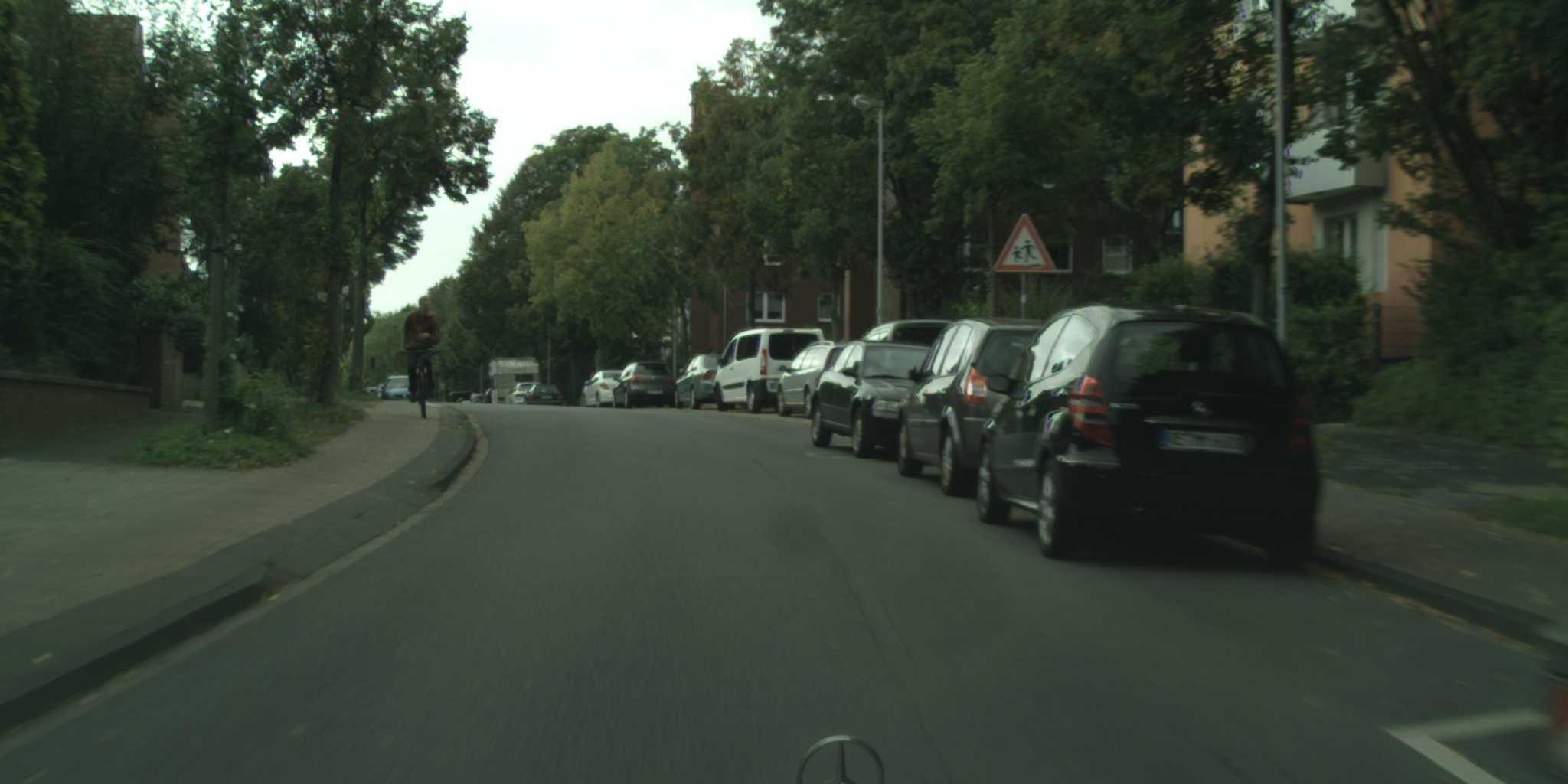}\vspace{0pt}
    \includegraphics[width=1\linewidth]{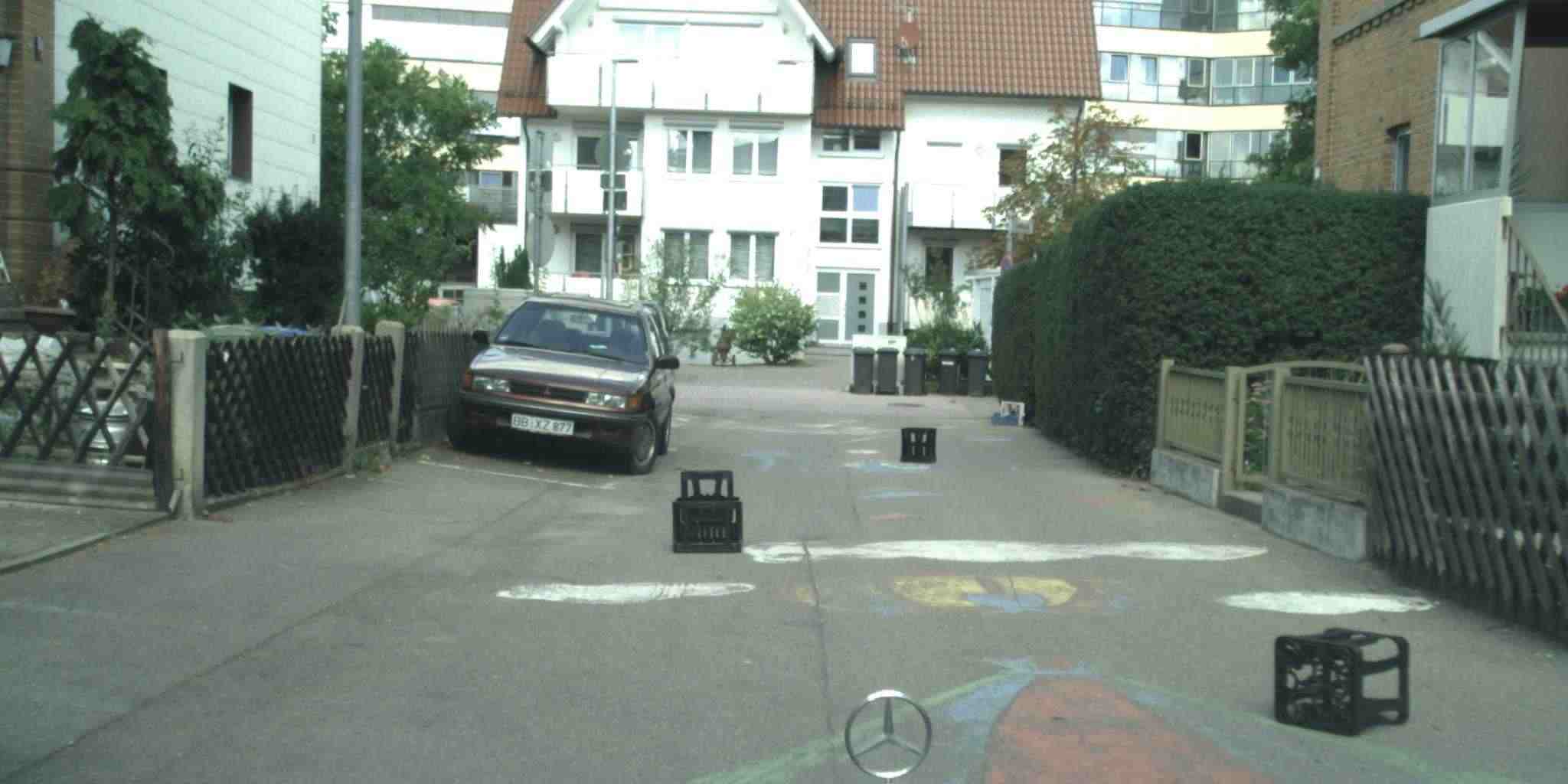}\vspace{0pt}
    \includegraphics[width=1\linewidth]{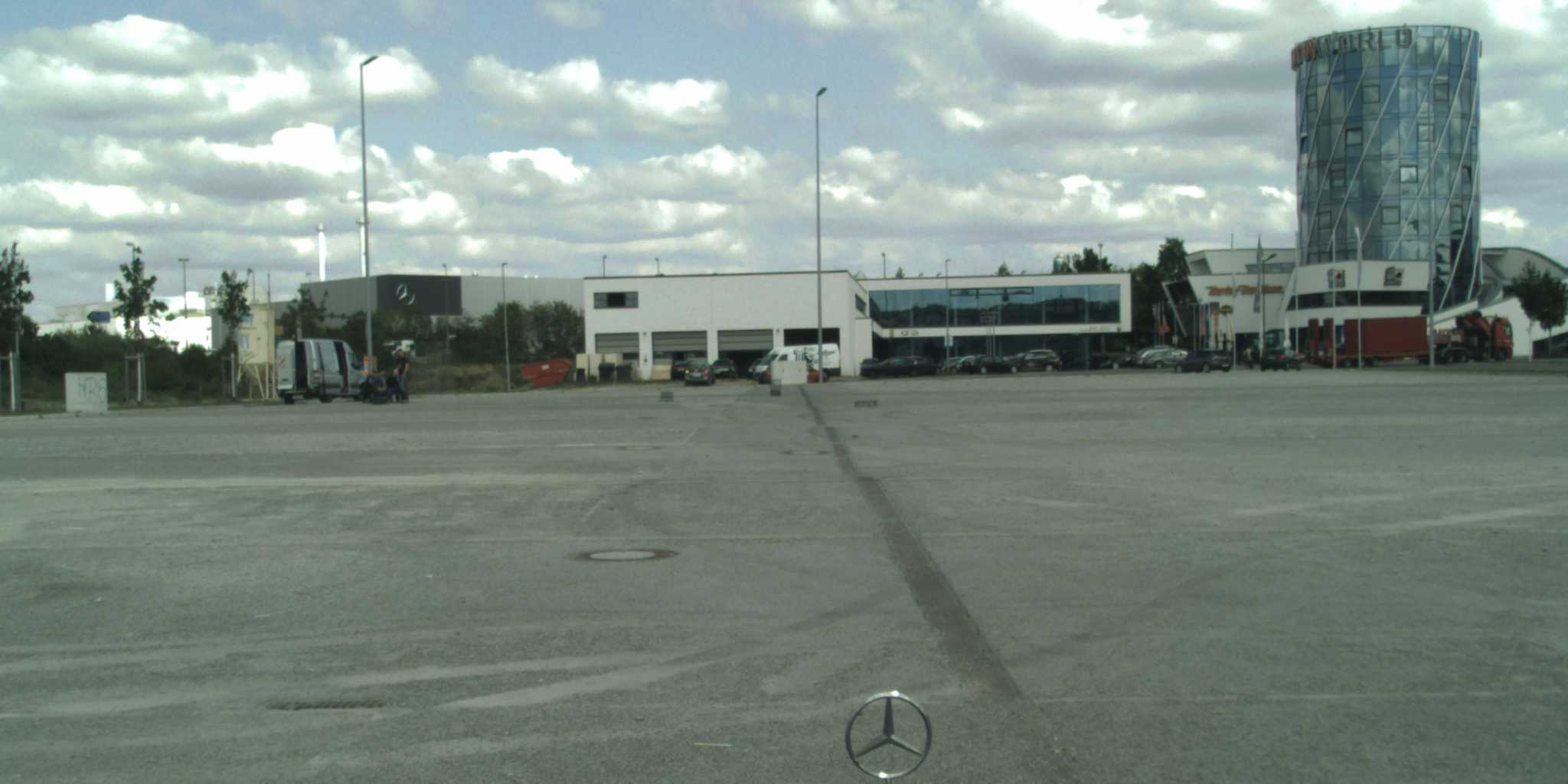}
    \end{minipage}}\hspace{-6pt}  % 控制subfigure之间横向距离
    \subfigure[Disparity]{
    \begin{minipage}[b]{0.235\linewidth}
        \includegraphics[width=1\linewidth]{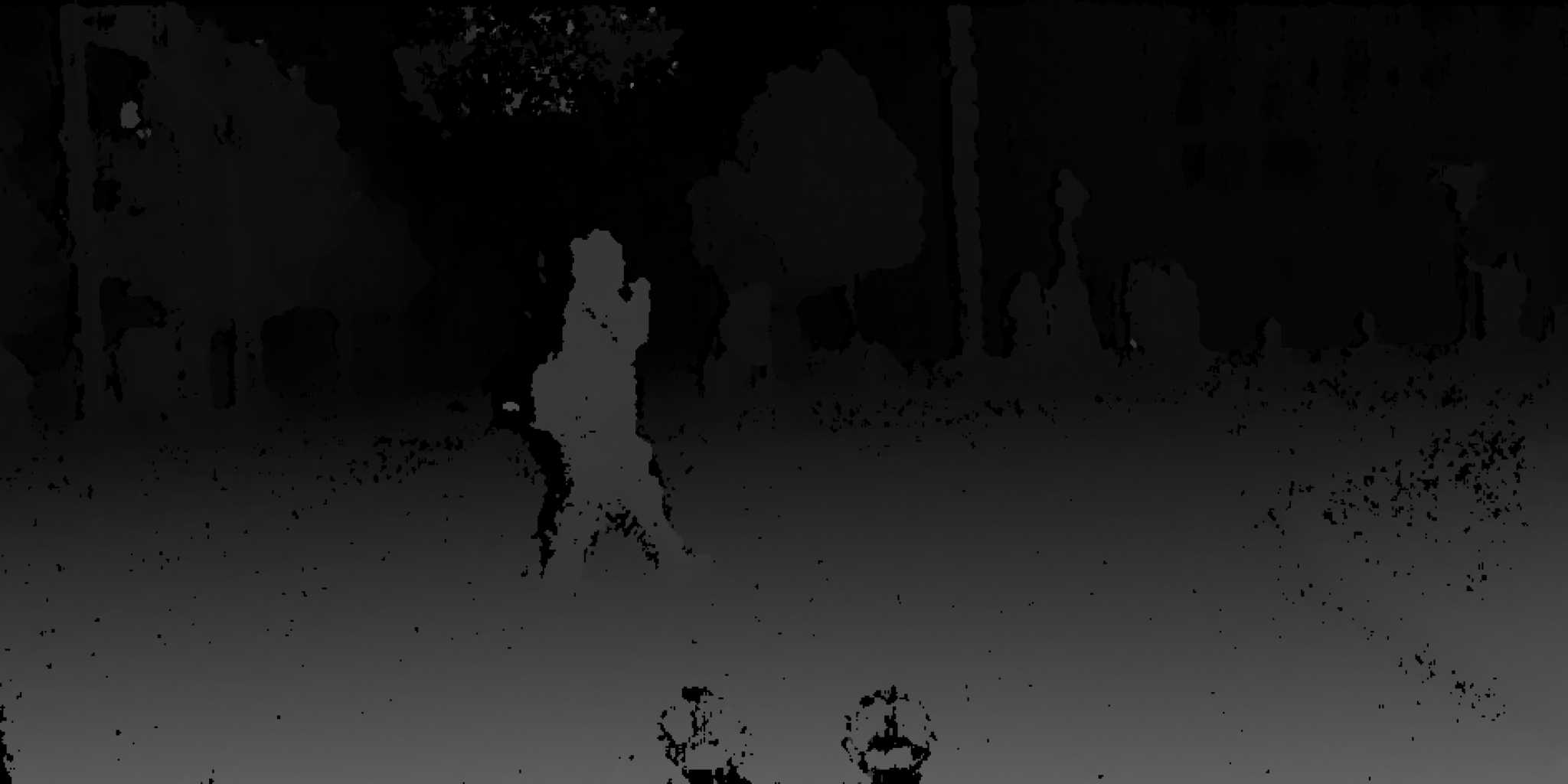}\vspace{0pt}  %控制垂直距离
        \includegraphics[width=1\linewidth]{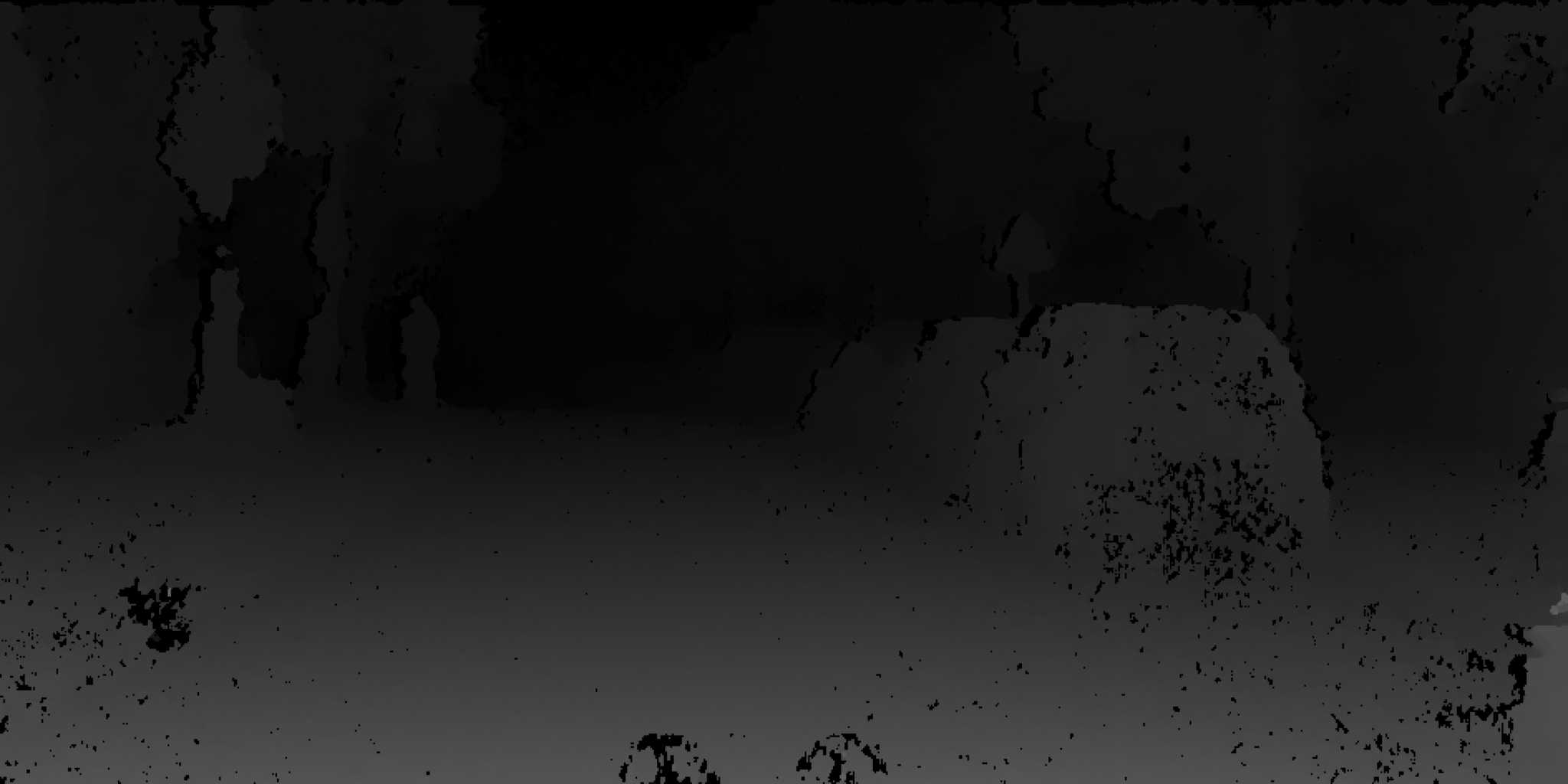}\vspace{0pt}
        \includegraphics[width=1\linewidth]{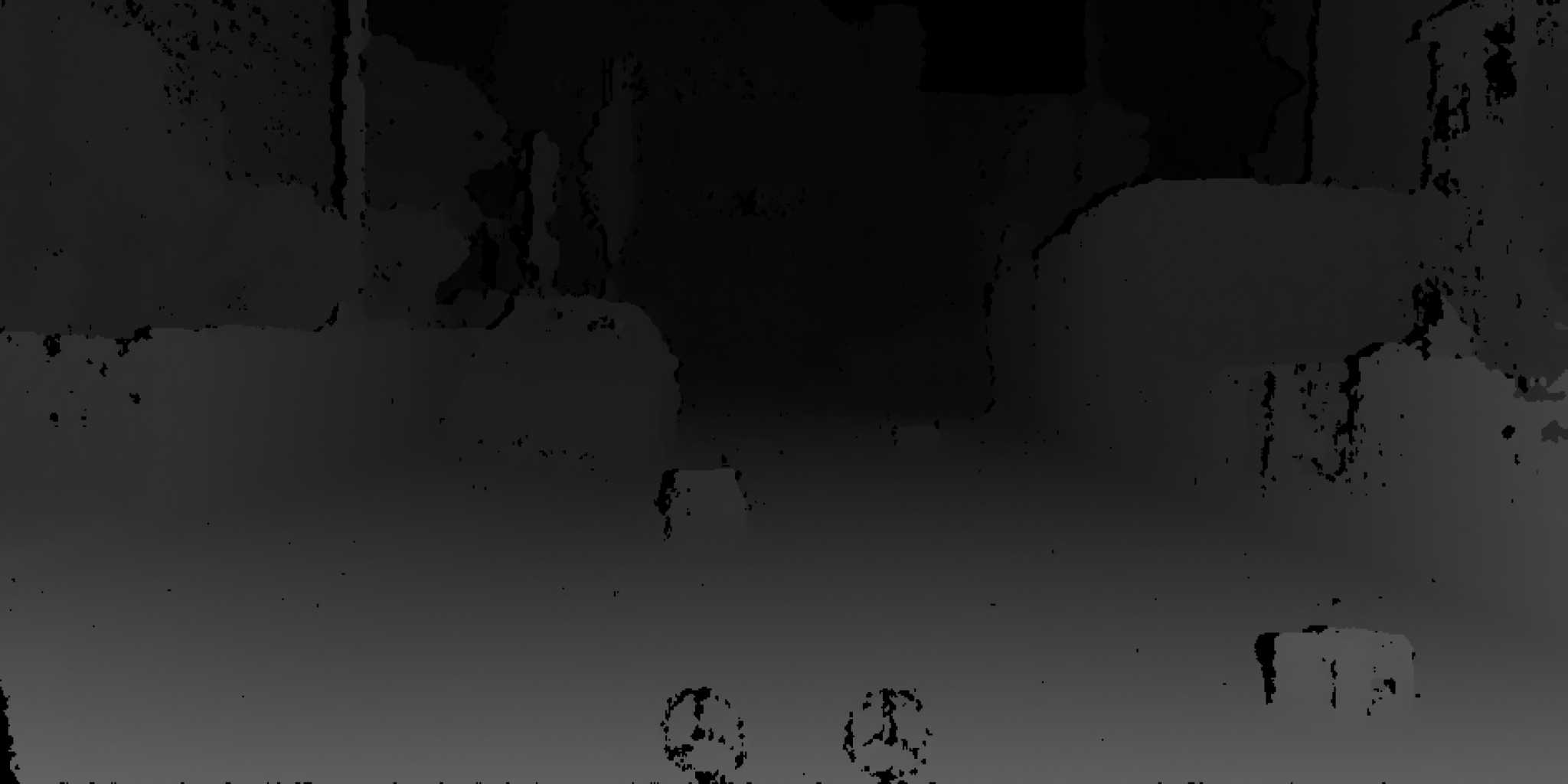}\vspace{0pt}
        \includegraphics[width=1\linewidth]{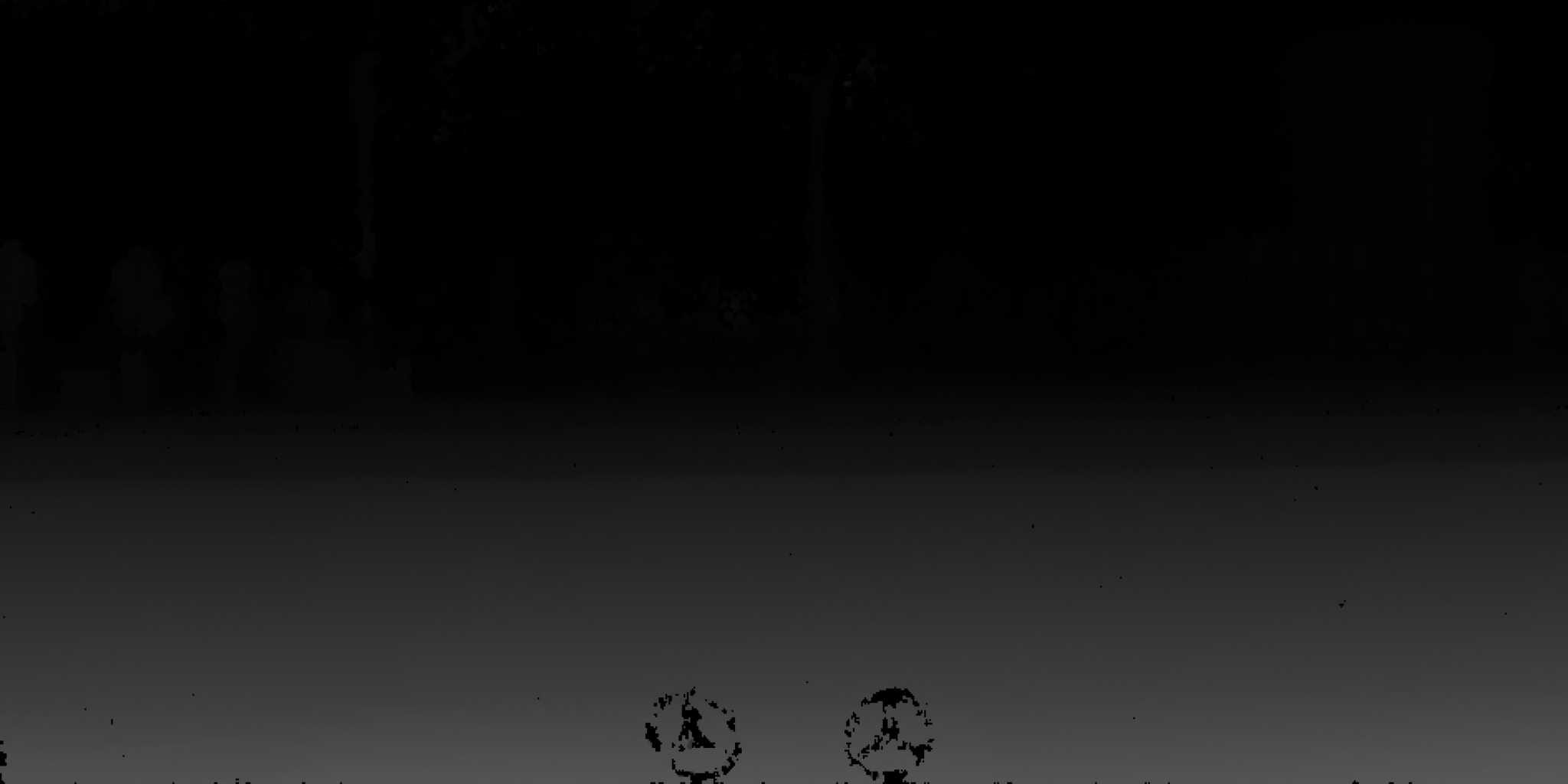}
    \end{minipage}}\hspace{-6pt}
    \subfigure[GT]{
    \begin{minipage}[b]{0.235\linewidth}
    \includegraphics[width=1\linewidth]{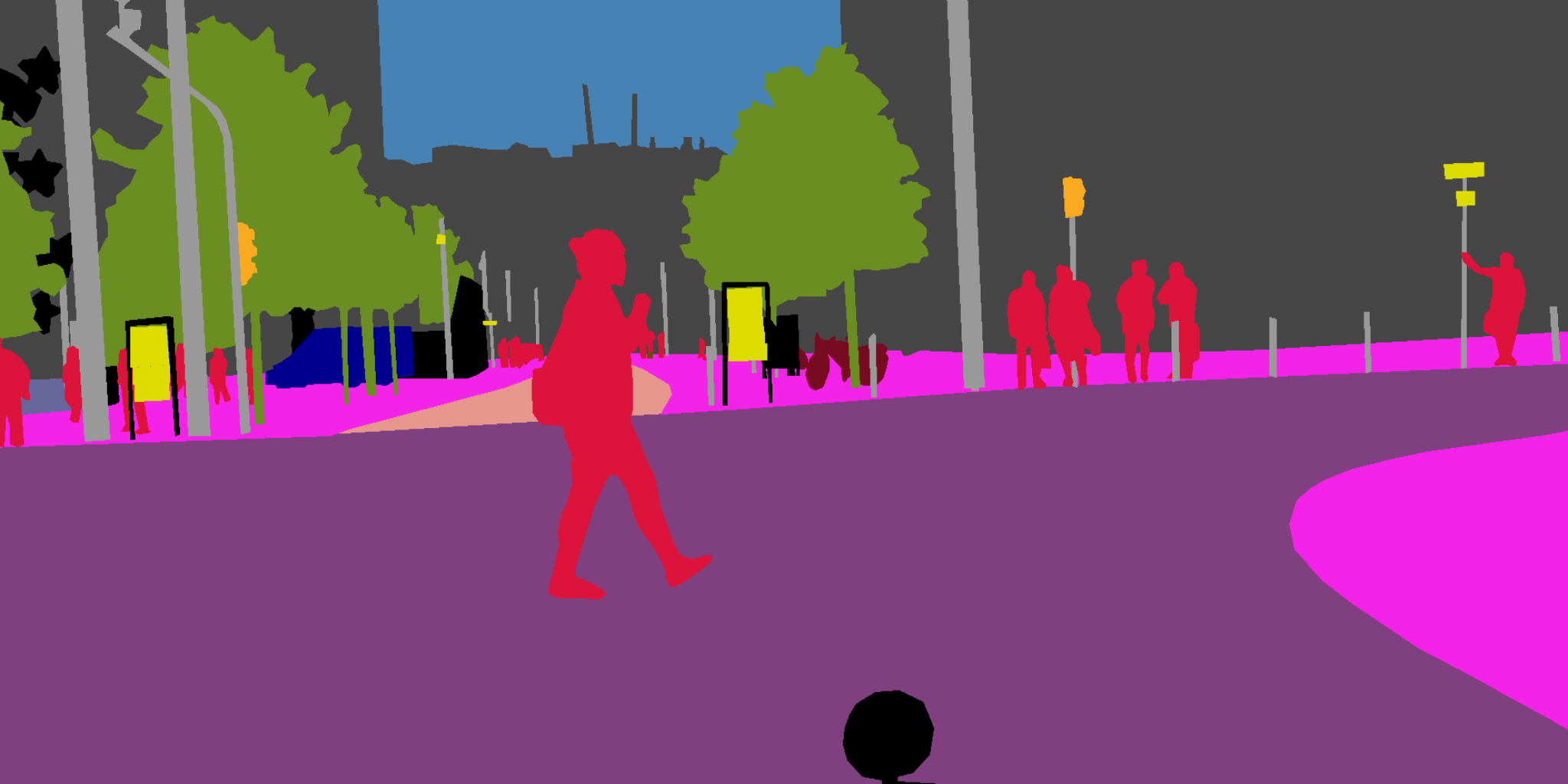}\vspace{0pt}
    \includegraphics[width=1\linewidth]{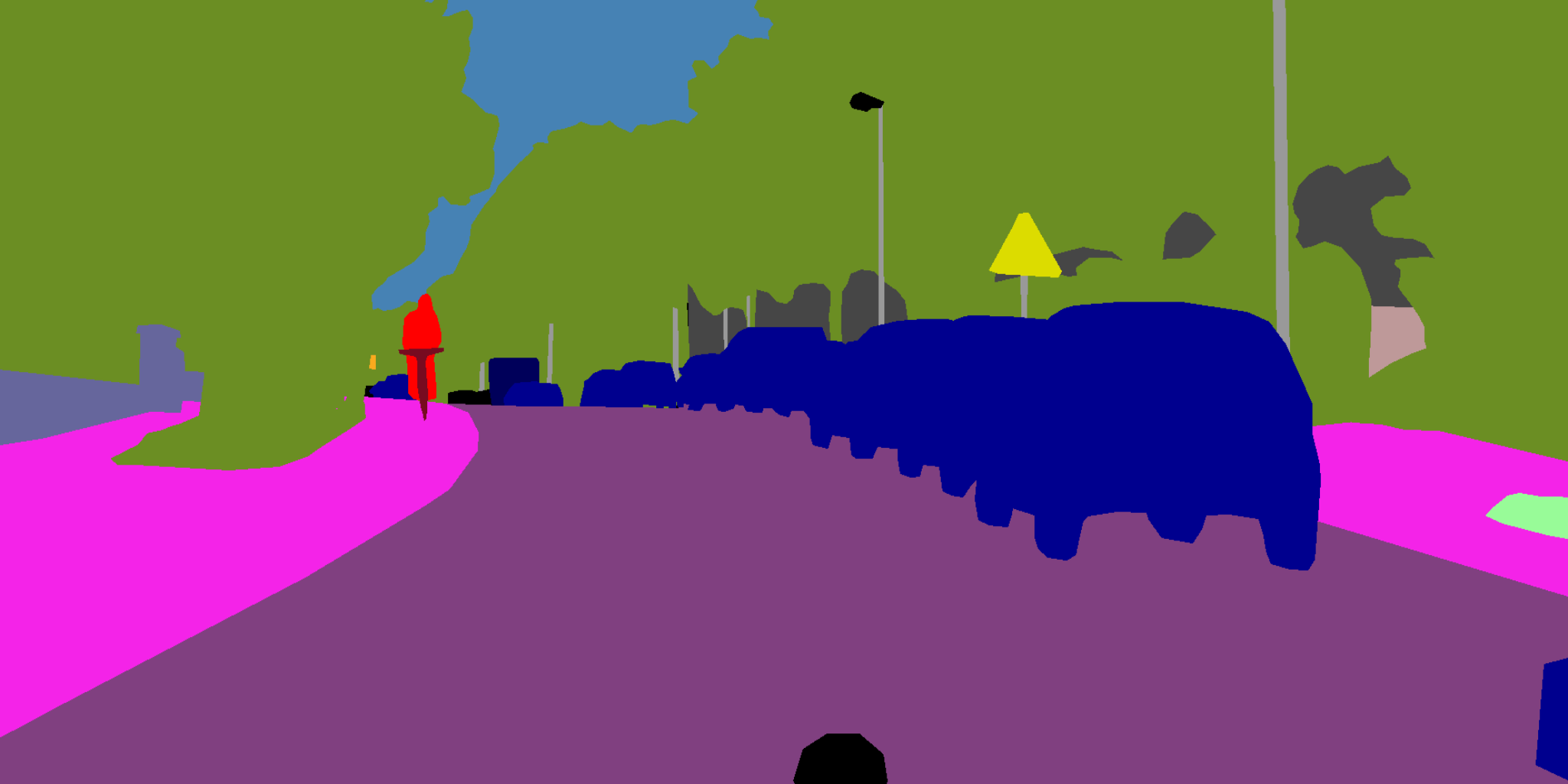}\vspace{0pt}
    \includegraphics[width=1\linewidth]{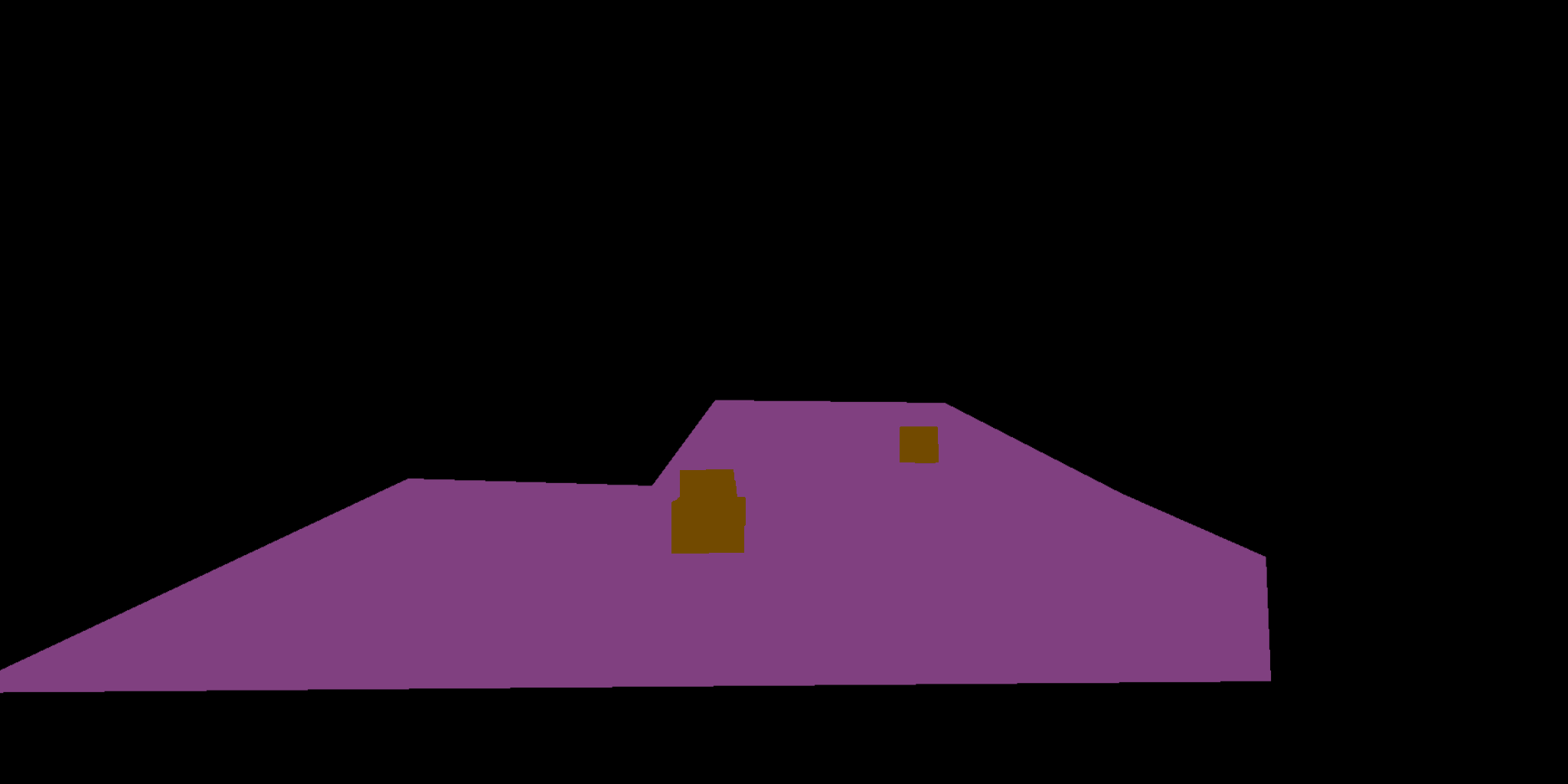}\vspace{0pt}
    \includegraphics[width=1\linewidth]{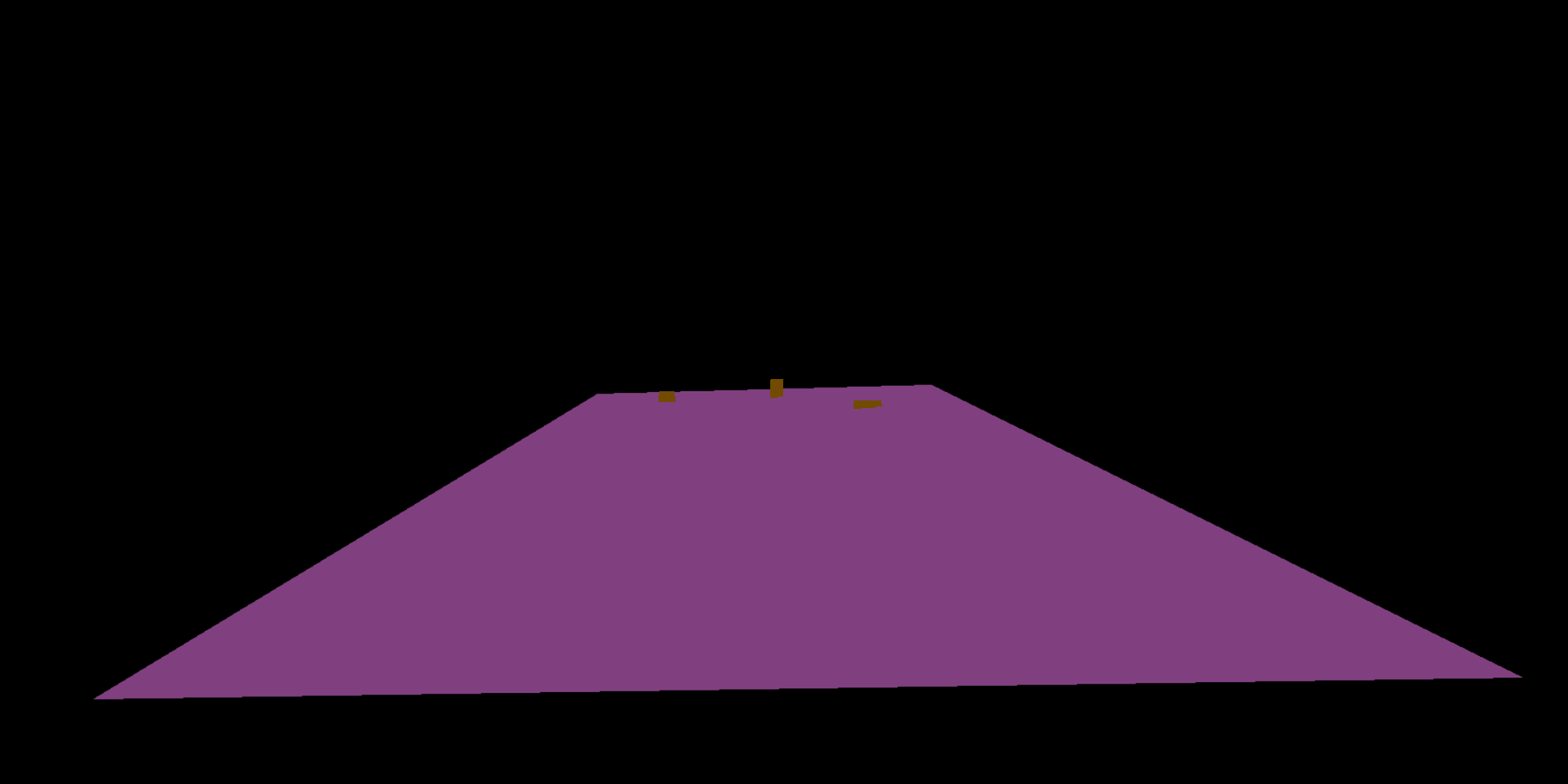}
    \end{minipage}}\hspace{-6pt}
    \subfigure[Results]{
    \begin{minipage}[b]{0.235\linewidth}
        \includegraphics[width=1\linewidth]{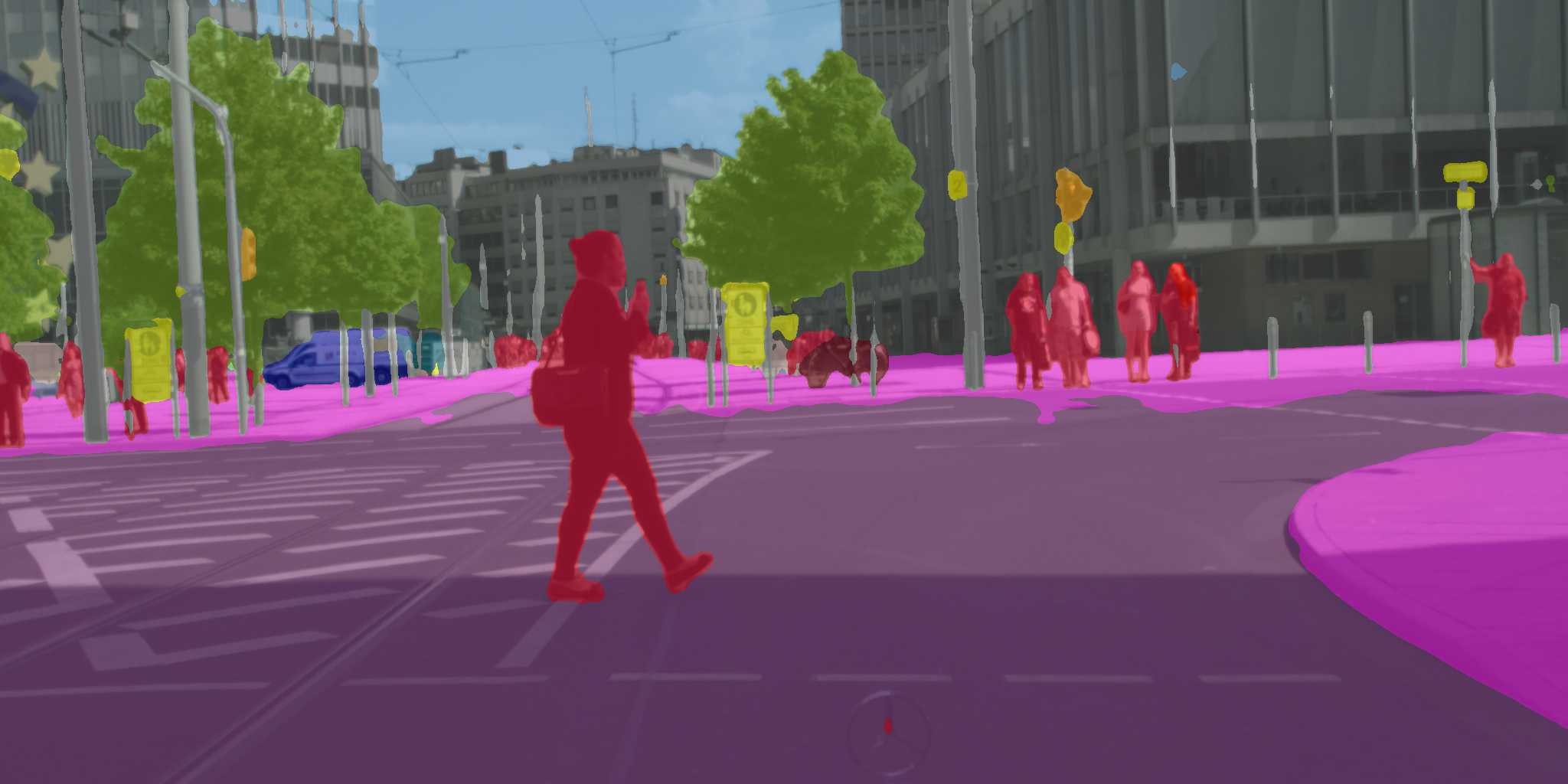}\vspace{0pt}
        \includegraphics[width=1\linewidth]{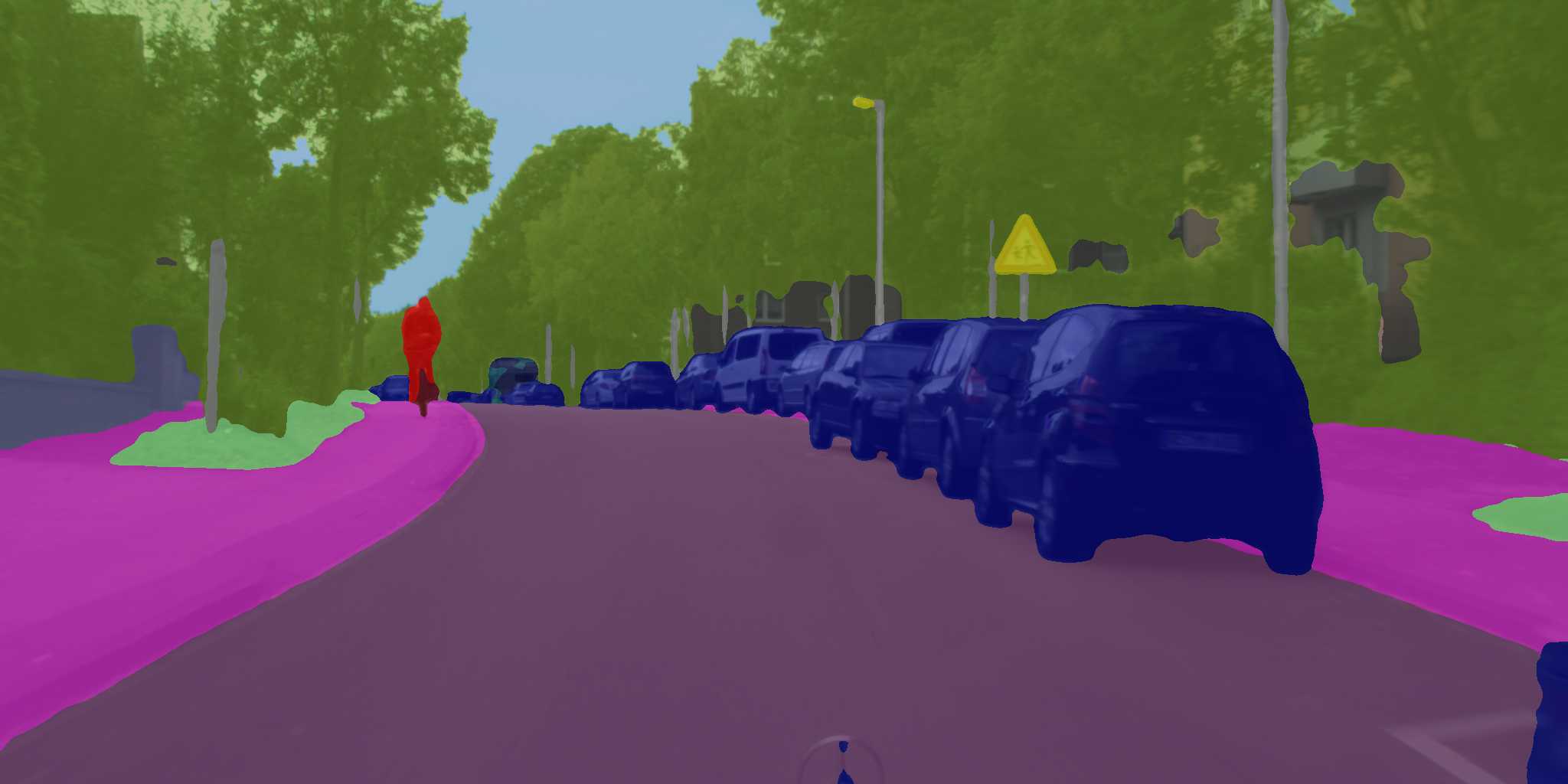}\vspace{0pt}
        \includegraphics[width=1\linewidth]{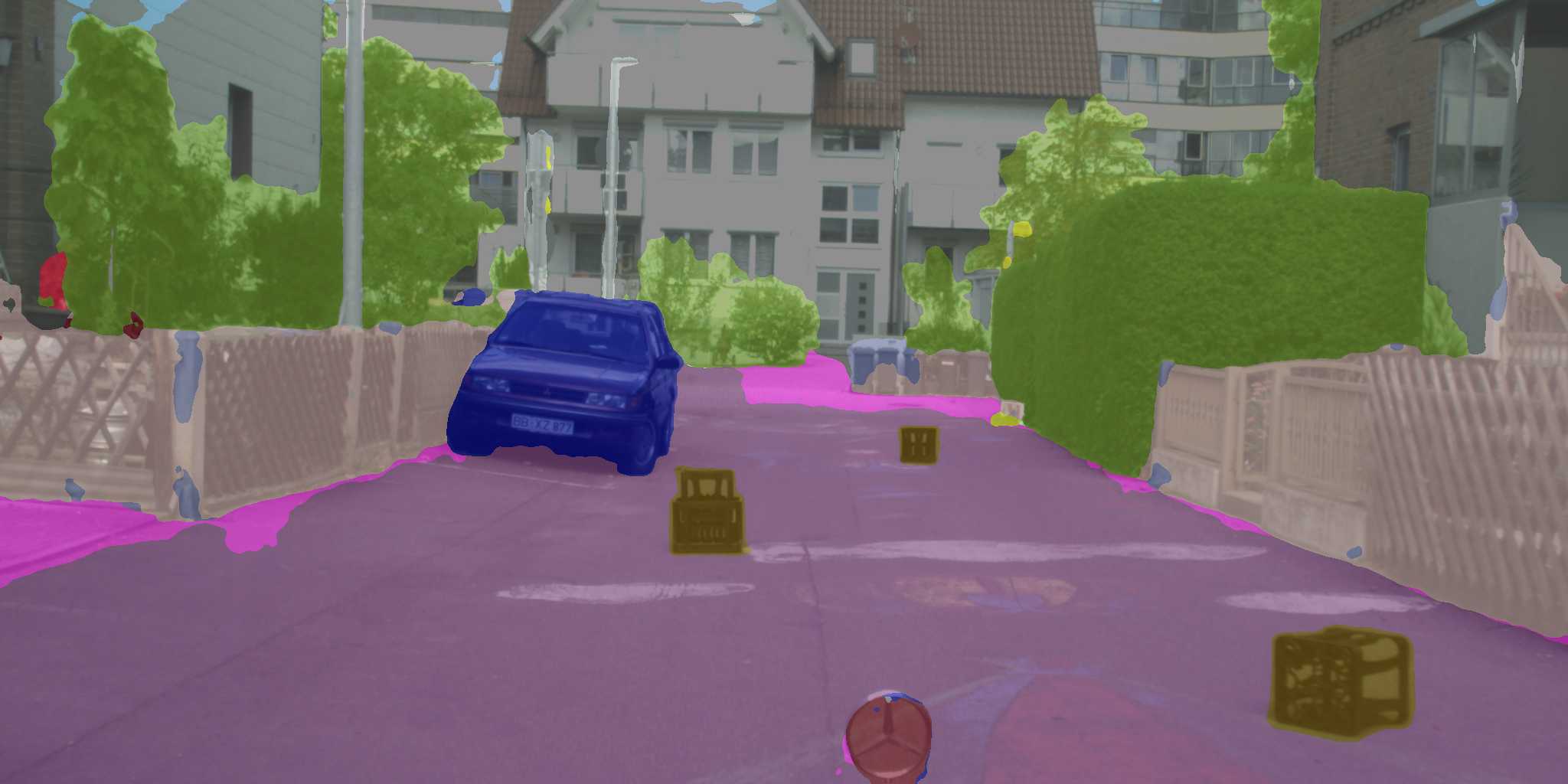}\vspace{0pt}
        \includegraphics[width=1\linewidth]{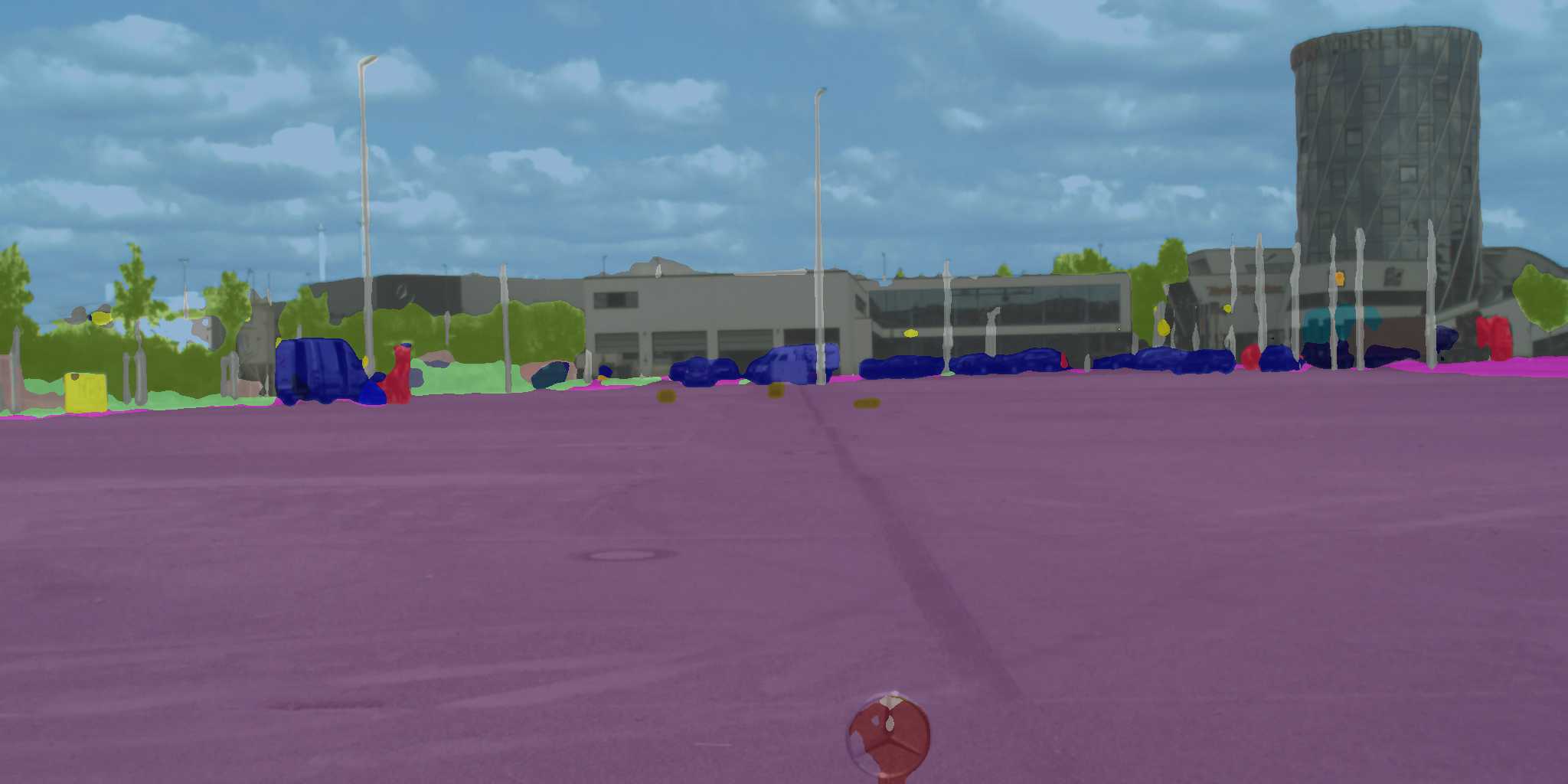}
    \end{minipage}}
    \caption[RFNet results]
    { \label{fig:RFNet_results}
    Predictions with additional unexpected obstacle class from RFNet.}
\end{figure}

\textbf{Numerical Performance Comparison.}
Based on our multi-dataset training, we create a benchmark to compare our RFNet with the other two real-time networks: ERF-PSPNet~\cite{ERF-PSPNet} (a light-weight network), SwiftNet~\cite{orsic2019defense} (whose network architecture is very similar to our RFNet). The first two networks only take RGB input. Table~\ref{benchmarking} shows IoU of all 20 classes in the new multi-source setting. Our RFNet achieves higher accuracy in most of the classes. Compared to SwiftNet, RFNet improves accuracy remarkably in certain classes like fence, traffic light, terrain, truck, bus, train, and small obstacle, which is benefited from the depth complementary information. Figure \ref{fig:RFNet_results} shows some examples from the validation set of \textit{Cityscpaes} and \textit{Lost and Found}, which demonstrates the excellent segmentation accuracy of our RFNet in various scenarios with or without small obstacles.
%我想解释和选这两个网络对比的原因
%增加不同深度精度对比

\begin{figure}
    \centering
    \subfigure[]{
    \centering
    \begin{minipage}[b]{0.48\textwidth}
    \includegraphics[width=1.0\textwidth]{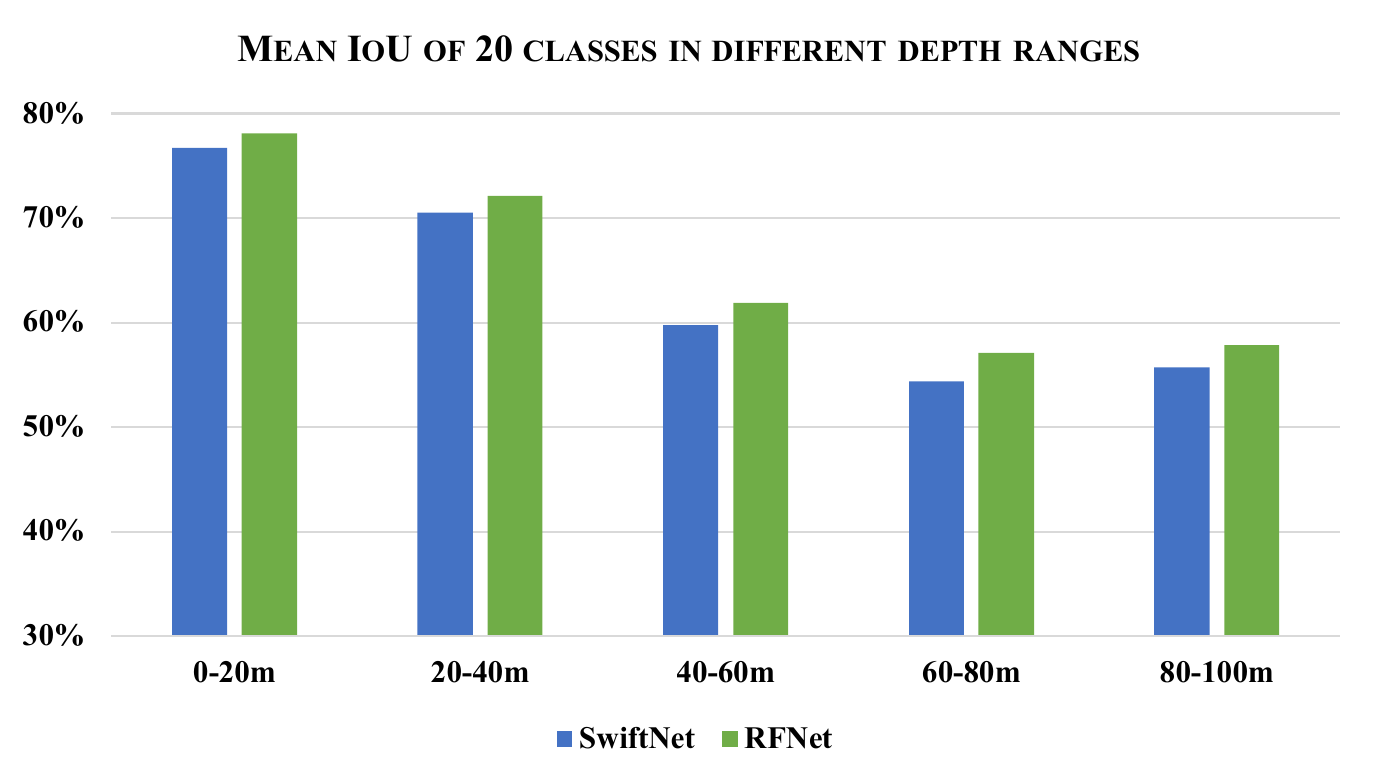}
    \end{minipage}
    }
    %\subfigure[SwiftNet with multi-dataset training]{
    %\centering
    %\begin{minipage}[b]{0.5\textwidth}
    %\includegraphics[width=0.9\textwidth]{./images/first/Swiftnet_multidataset/big.png}
    %\end{minipage}
    %}
    \subfigure[]{
        \centering
        \begin{minipage}[b]{0.48\textwidth}
        \includegraphics[width=1.0\textwidth]{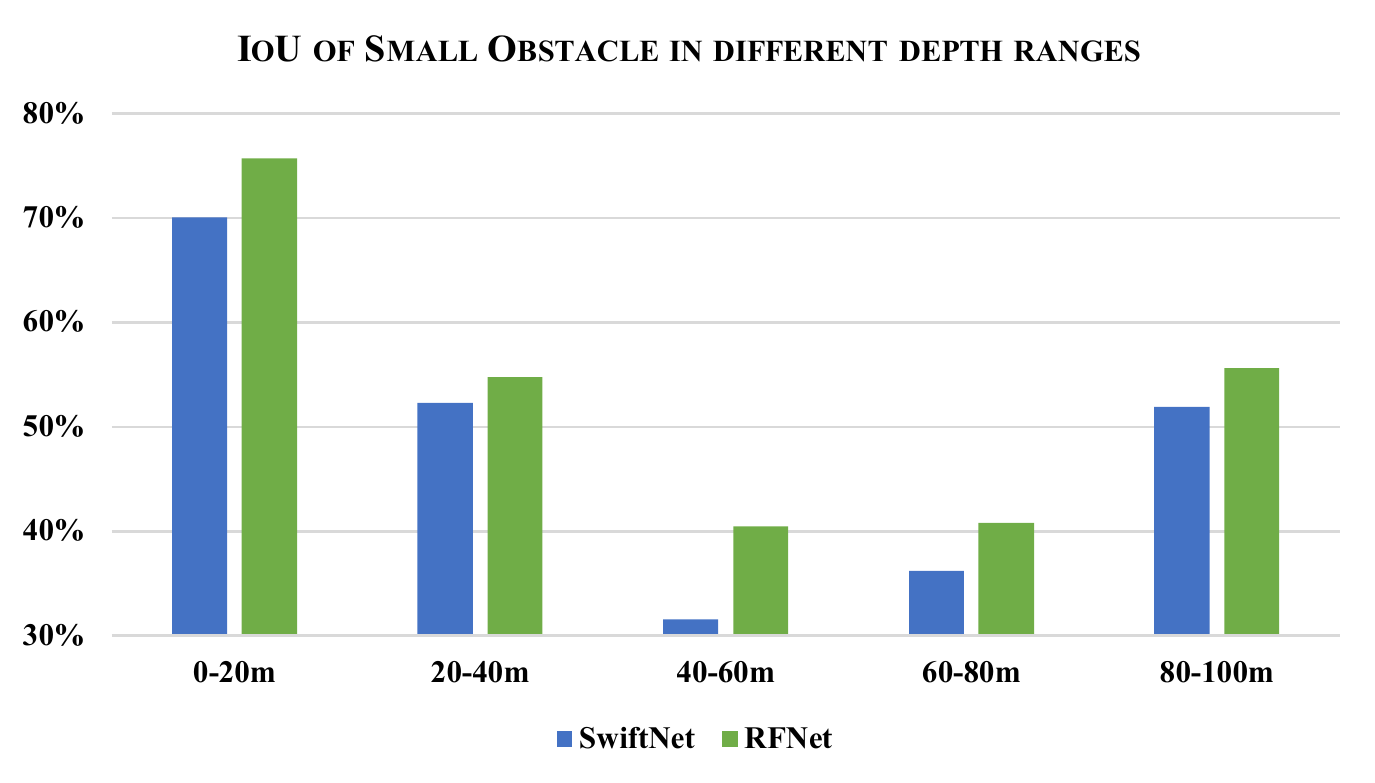}
        \end{minipage}
    }
    \caption{Mean IoU of 20 classes and IoU of Small Obstacle from SwiftNet and RFNet respectively. RFNet improves precision in all depth ranges, especially in close and middle ranges.} 
    \label{fig:bar_graph}
    
\end{figure}

To explore how the proposed RFNet improves precision in different depth ranges, we perform analysis on mean IoU and the IoU of small obstacle in different depth ranges for RFNet and SwiftNet. We calculate the depth value of each pixel from disparity value. The maximum depth value is set to 100 and limited by the quality of disparity image, while all the unmatched pixels are set to 100. Bar graph~\ref{fig:bar_graph} shows that RFNet performs better in all depth ranges in the case of mean IoU of 20 classes. Specifically, RFNet boosts the accuracy of unexpected small obstacle recognition in close and middle ranges remarkably. This is reasonable because disparity images derived from semi-global matching algorithm have higher accuracy at close range, and contribute more to the prediction than pixels with greater depth values.
%解释一下为什么RFNet在中近距离效果更好:SGM计算的视差图在近距离有更高的准确度，因此为ss精度提升贡献更多。在远距离情况下，视差图的贡献更小。
%small obstacle 作为类别（class）统一大写

In Table~\ref{comparison} we also compare our RFNet with other state-of-the-art networks on the \textit{Cityscapes} validation set. The column of speed reports the inference speed of a full resolution image   (2048$\times$1024) on a single RTX 2080Ti.
%添加解释，RFNet, SwiftNet, ERF-PSPNet三者在同样的硬件平台取得的结果,这样够吗？
Specifically, ERF-PSPNet and SwiftNet are implemented on the same hardware.
Compared to mainstream RGB semantic segmentation networks, our RFNet achieves better results while maintaining a real-time performance, which proves that exploiting depth information helps improving accuracy.
In the table, we also list some other RGB-D fusion networks: LDFNet~\cite{hung2019incorporating} and RFBNet based on ERFNet~\cite{deng2019rfbnet}. Our RFNet is both more accurate and faster than these multimodal networks. Overall, rare multi-modal semantic segmentation methods meet the real-time prediction speed, while our method achieves the highest accuracy on the validation set of \textit{Cityscapes} to the best of our knowledge that meets both demands including real-time inference, highly qualified accuracy, and capacity to leverage complementary features in cross-modal imagery.

\begin{table}[htbp]
    \centering
    \caption{Comparison of semantic segmentation methods on the validation set of Cityscapes.}
    \label{comparison}
    \begin{tabular}{l|c|c|c}
    \multicolumn{1}{c|}{\textbf{Network}} & \textbf{Multimodal}       & \textbf{mIoU(\%)} & \multicolumn{1}{l}{\textbf{Speed (FPS)}} \\ \hline \hline
    FCN8s~\cite{long2015fully}   &  \xmark  &  65.3\%  &  2.0 $^{\mathrm{*}}$  \\
    DeepLabV2-CRF~\cite{chen2017deeplab}  & \xmark  &  70.4\%  &  n/a  \\ 
    ENet~\cite{paszke2016enet}   &  \xmark  &  58.3\%  &  76.9 $^{\mathrm{*}}$  \\
    ERFNet~\cite{romera2017erfnet}    &       \xmark     & 65.8\%            & 20.8                    \\
    ERF-PSPNet~\cite{ERF-PSPNet}     &     \xmark    & 64.1\%            & 20.4                    \\
    SwiftNet~\cite{orsic2019defense}      &    \xmark    & 72.0\%           & 41.0                            \\
    VGG-D (ScaleInvariant)~\cite{krevso2016convolutional}  & \cmark & 64.4\% & n/a \\
    LDFNet~\cite{hung2019incorporating}    & \cmark                & 68.5\%           & 18.4                                    \\
    GoogLeNet (NiN-2)~\cite{schneider2017multimodal} & \cmark  &  69.1\%  &  n/a  \\ %直接加这个会不会有点突兀？
    RFBNet (ERFNetEnc)~\cite{deng2019rfbnet}     & \cmark                & 72.0\%            & n/a                                     \\ \hline
    RFNet (Ours)                           & \cmark                & 72.5\%            & 22.2                   \\
    \multicolumn{4}{l}{$^{\mathrm{*}}$ Speed on half resolution images.}
    \end{tabular}
\end{table}

%放一些RFNet inference出来的图片

%统计不同深度的精度

\begin{figure}[t]
    \centering
    \subfigure[RGB feature maps]{
    \begin{minipage}[b]{0.32\linewidth}
    \includegraphics[width=1\linewidth]{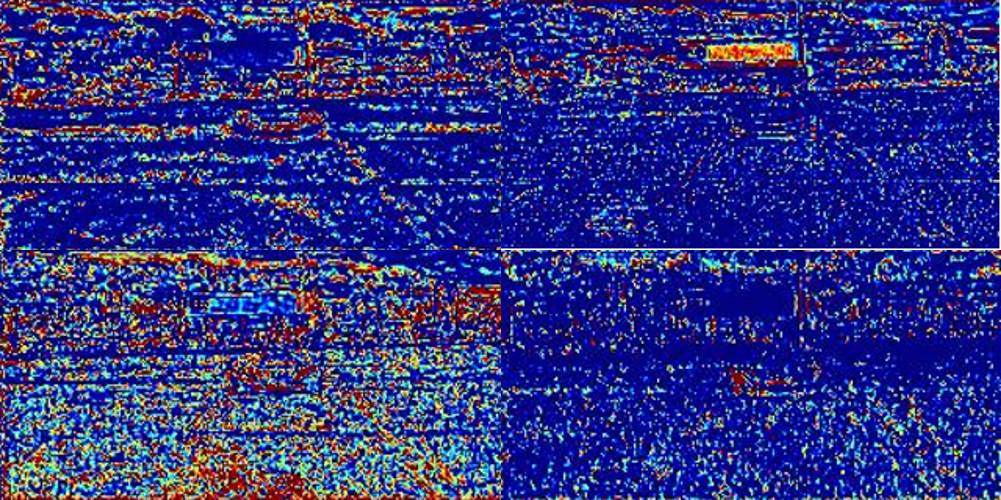}\vspace{0pt}  %控制垂直距离
    \end{minipage}}\hspace{-6pt} 
    \subfigure[Depth feature maps]{
    \begin{minipage}[b]{0.32\linewidth}
    \includegraphics[width=1\linewidth]{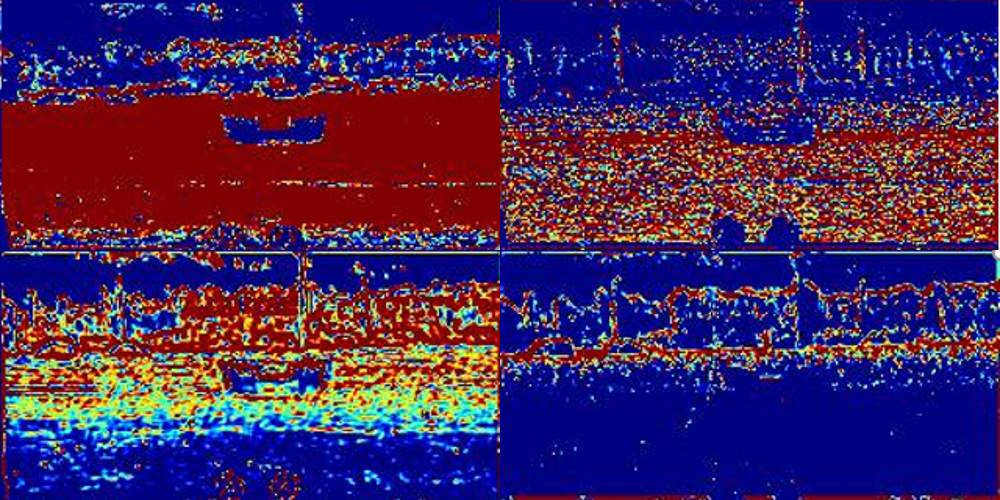}\vspace{0pt}  %控制垂直距离
    \end{minipage}}\hspace{-6pt}
    \subfigure[Merged feature maps]{
    \begin{minipage}[b]{0.32\linewidth}
    \includegraphics[width=1\linewidth]{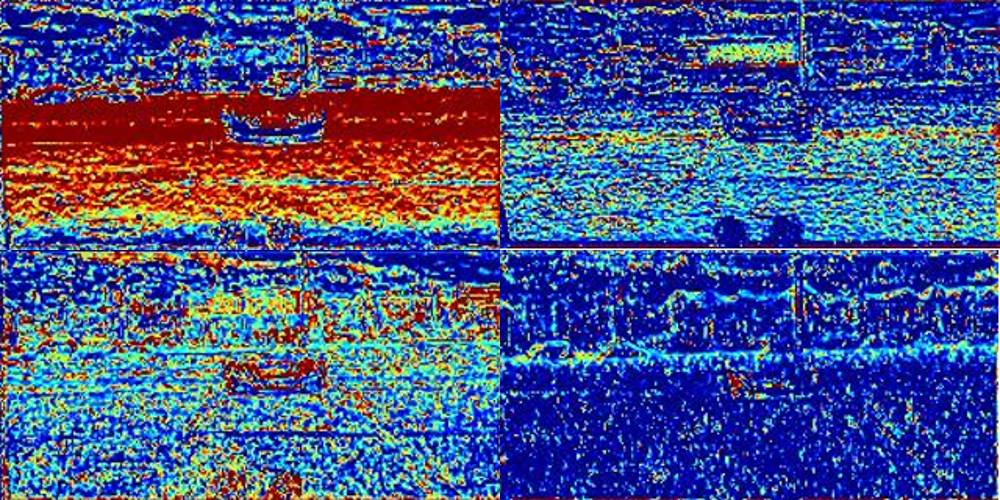}\vspace{0pt}  %控制垂直距离
    \end{minipage}}\hspace{-6pt}
    % \caption[mix directly]
    \caption{Visualization of feature maps from the second block of RFNet.}
    \label{fig:feature_maps}
\end{figure}

\begin{figure*}
    \centering
    \subfigure[RGB]{
    \begin{minipage}[b]{0.19\linewidth}
        \includegraphics[width=1\linewidth]{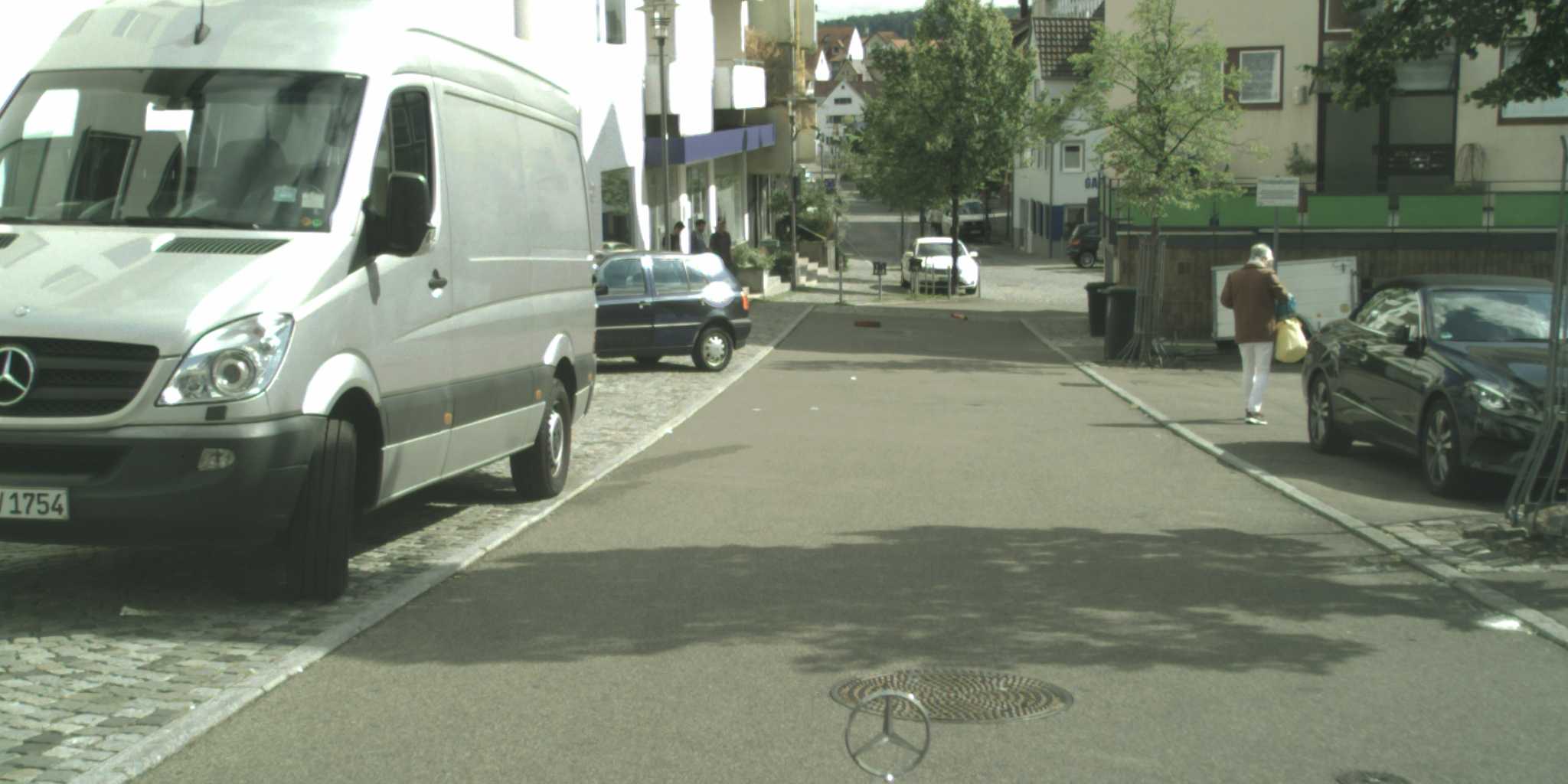}\vspace{1pt}  %控制垂直距离
        \includegraphics[width=1\linewidth]{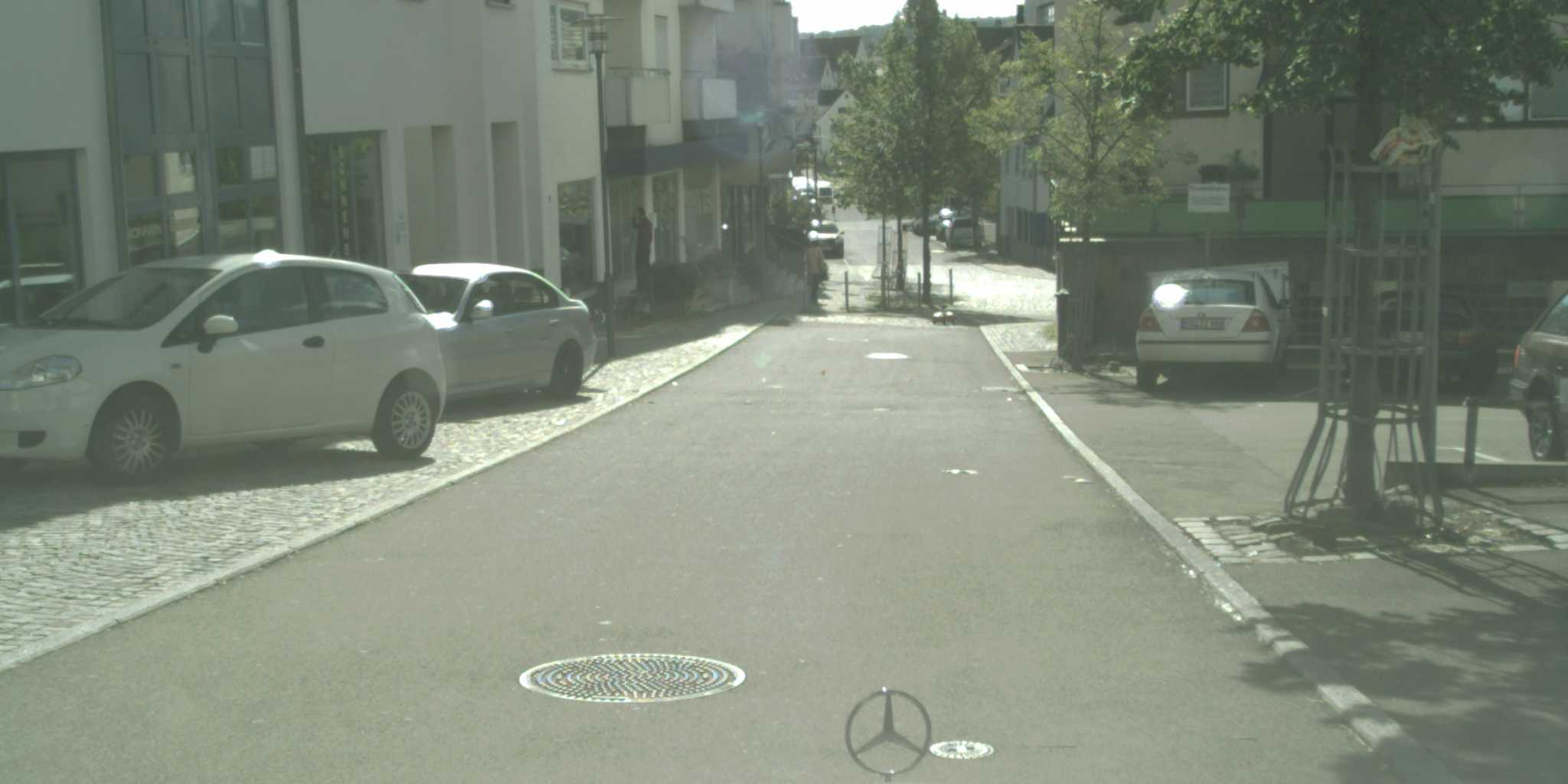}\vspace{1pt}
        \includegraphics[width=1\linewidth]{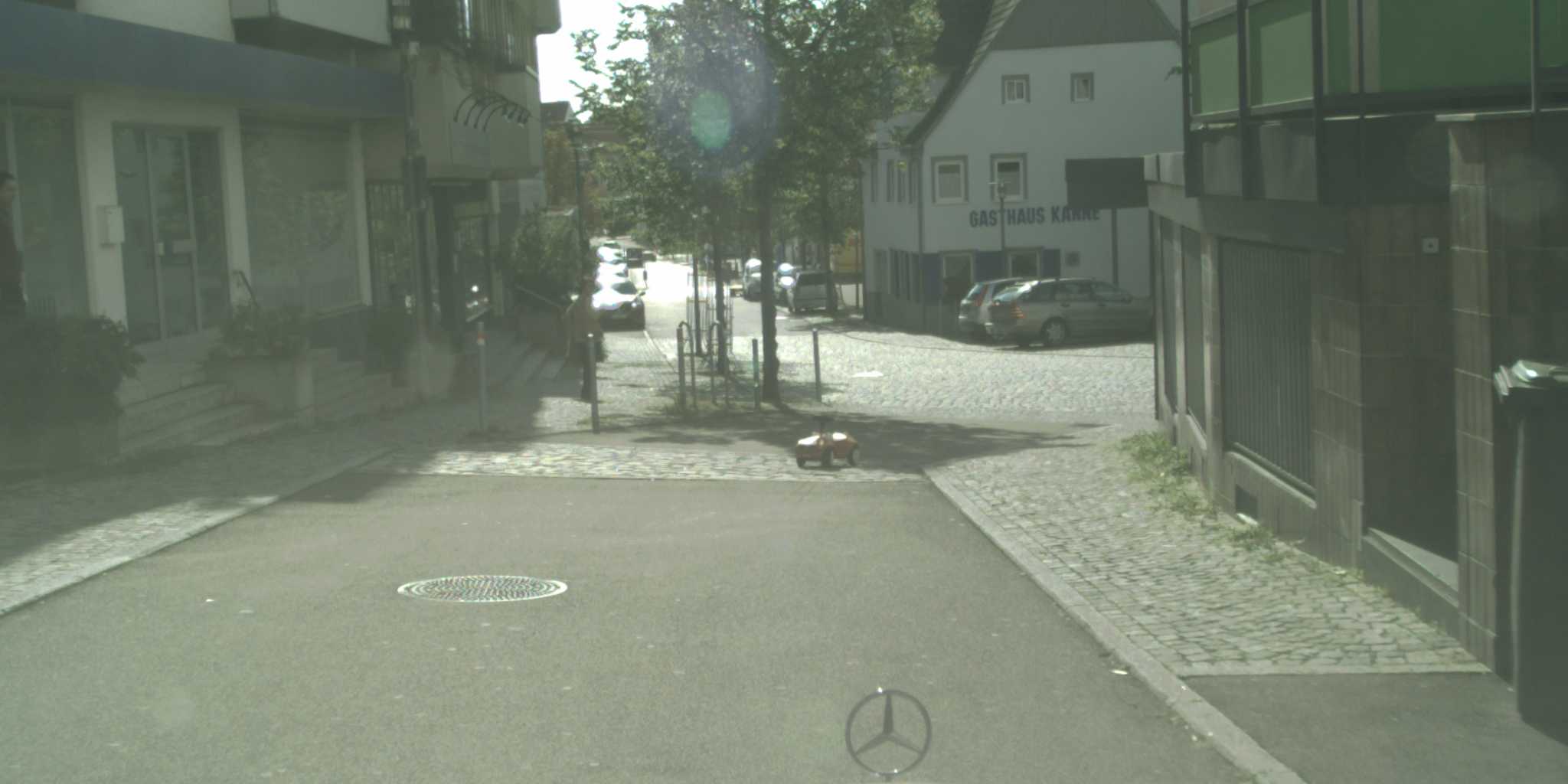}\vspace{1pt}
        \includegraphics[width=1\linewidth]{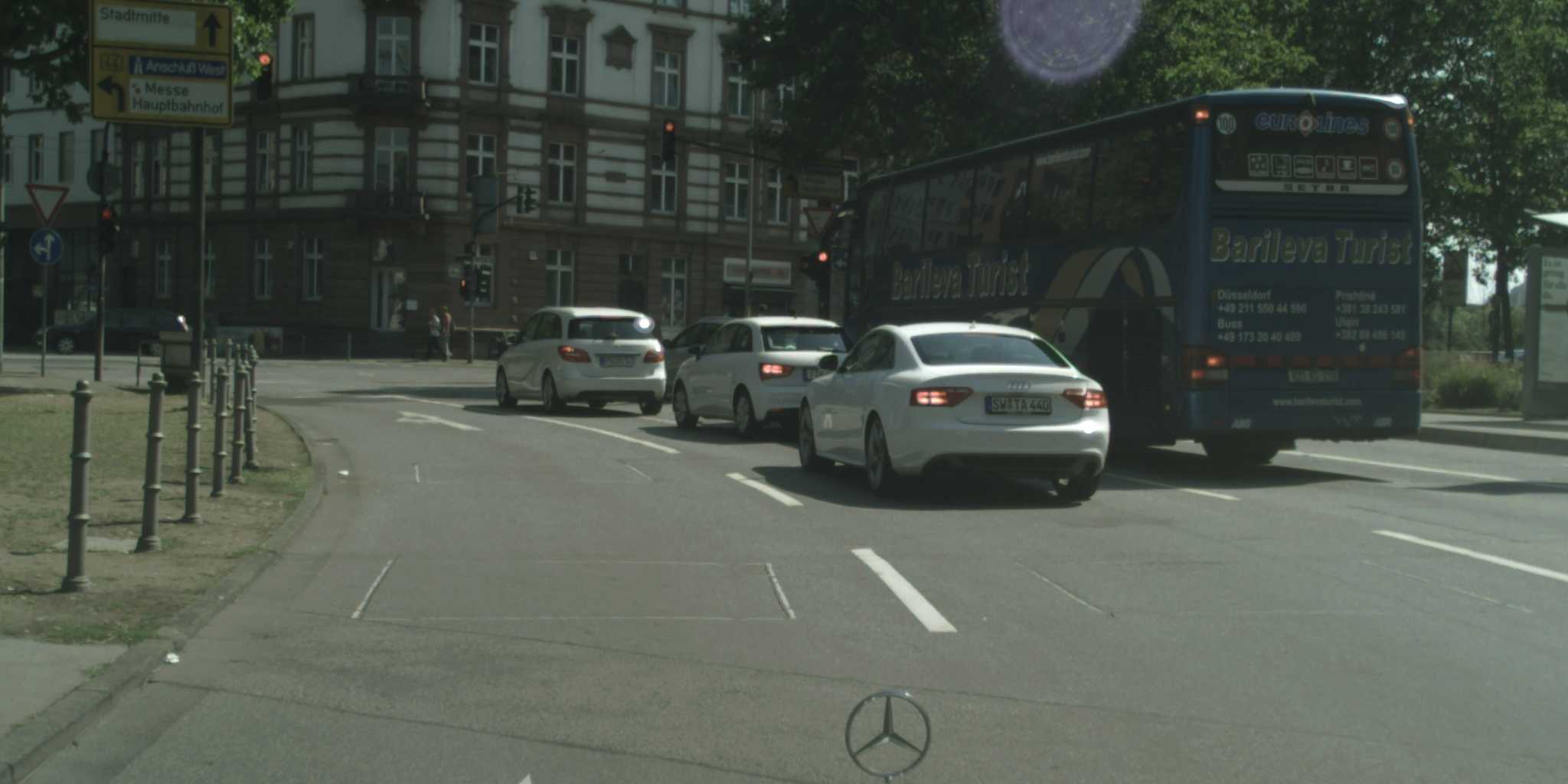}\vspace{1pt}
        \includegraphics[width=1\linewidth]{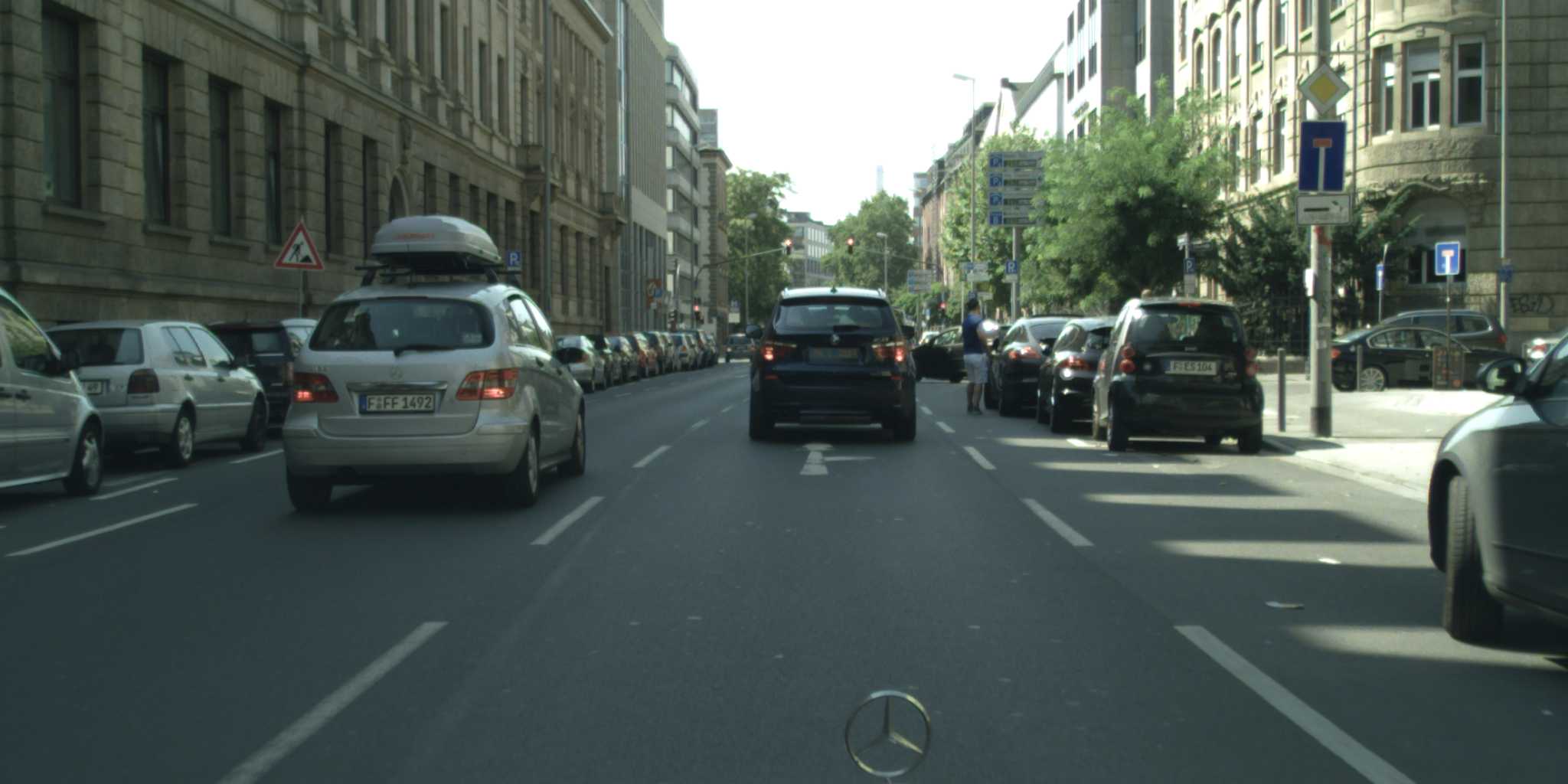}\vspace{1pt}
        \includegraphics[width=1\linewidth]{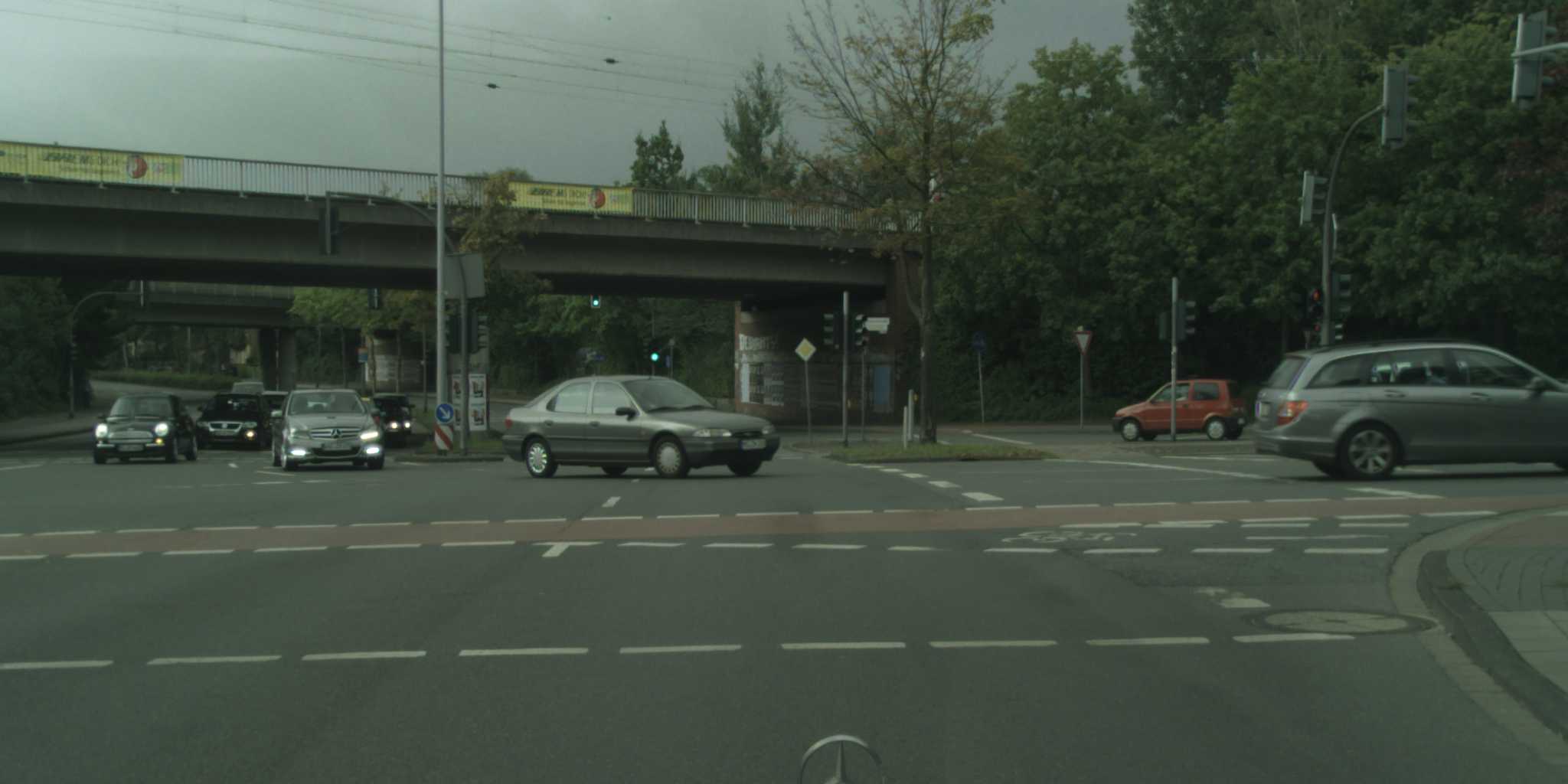}\vspace{1pt}
    \end{minipage}}\hspace{-4pt}  % 控制subfigure之间横向距离
    \subfigure[Disparity]{
    \begin{minipage}[b]{0.19\linewidth}
        \includegraphics[width=1\linewidth]{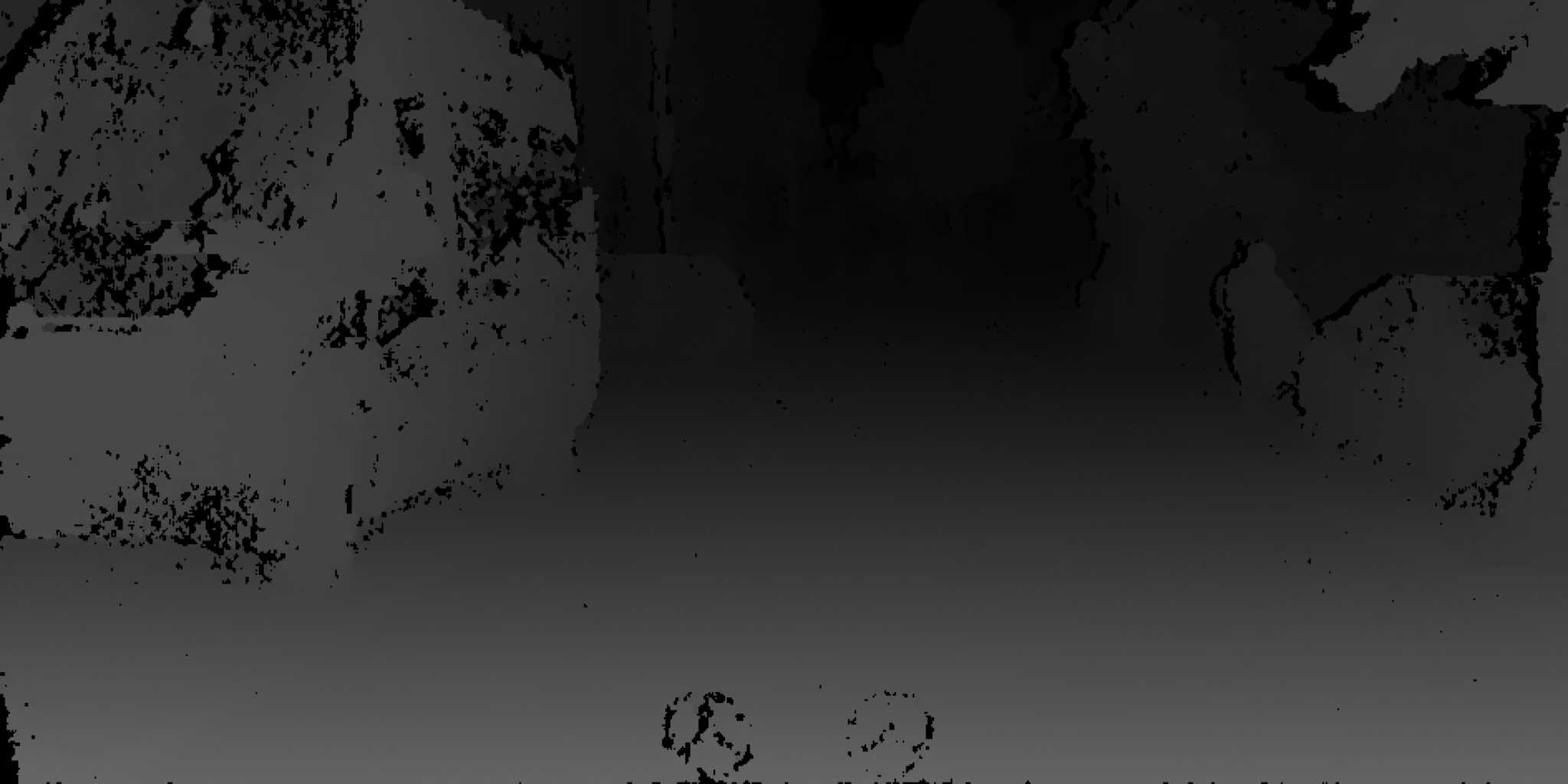}\vspace{1pt}  %控制垂直距离
        \includegraphics[width=1\linewidth]{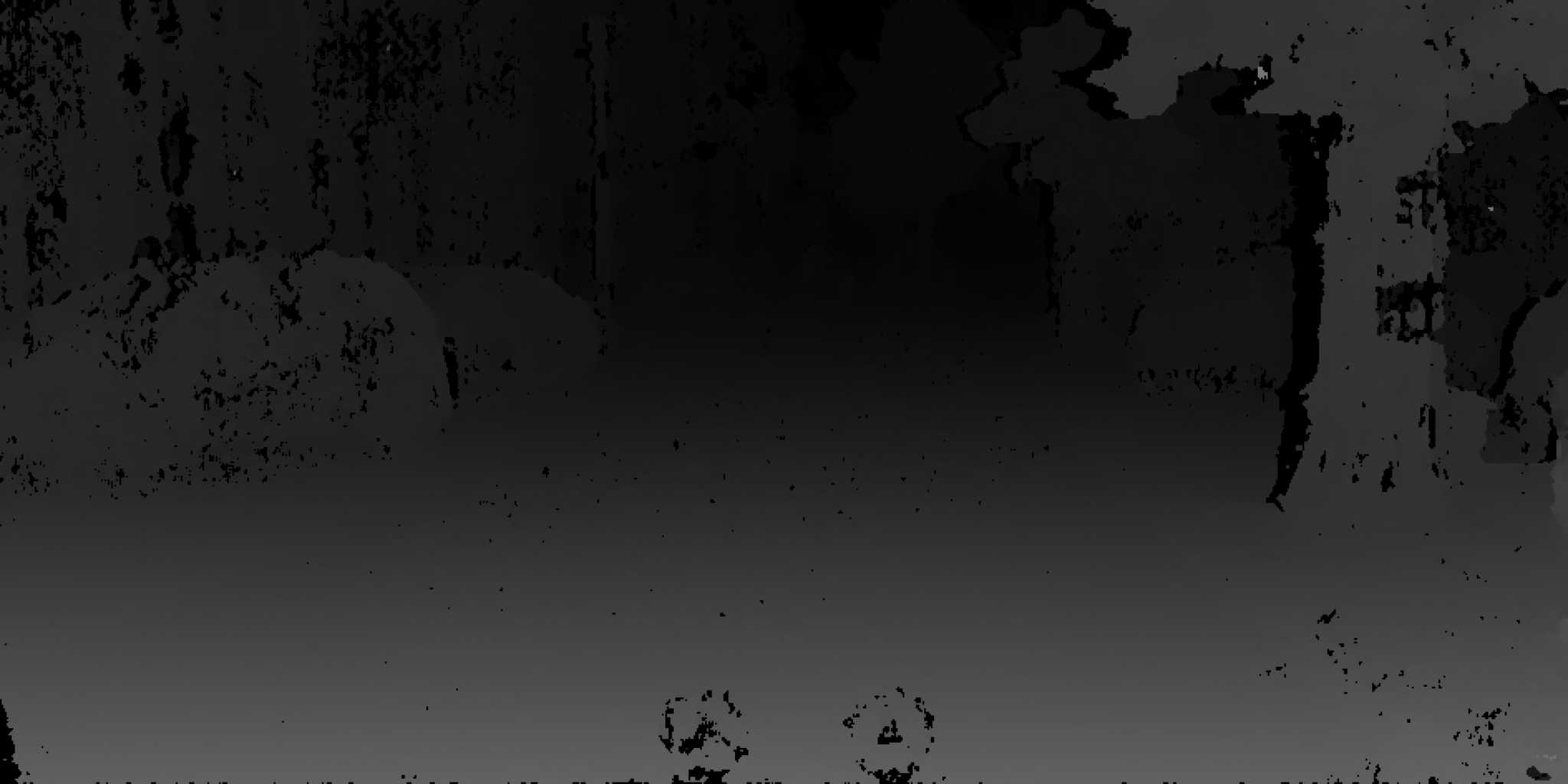}\vspace{1pt}
        \includegraphics[width=1\linewidth]{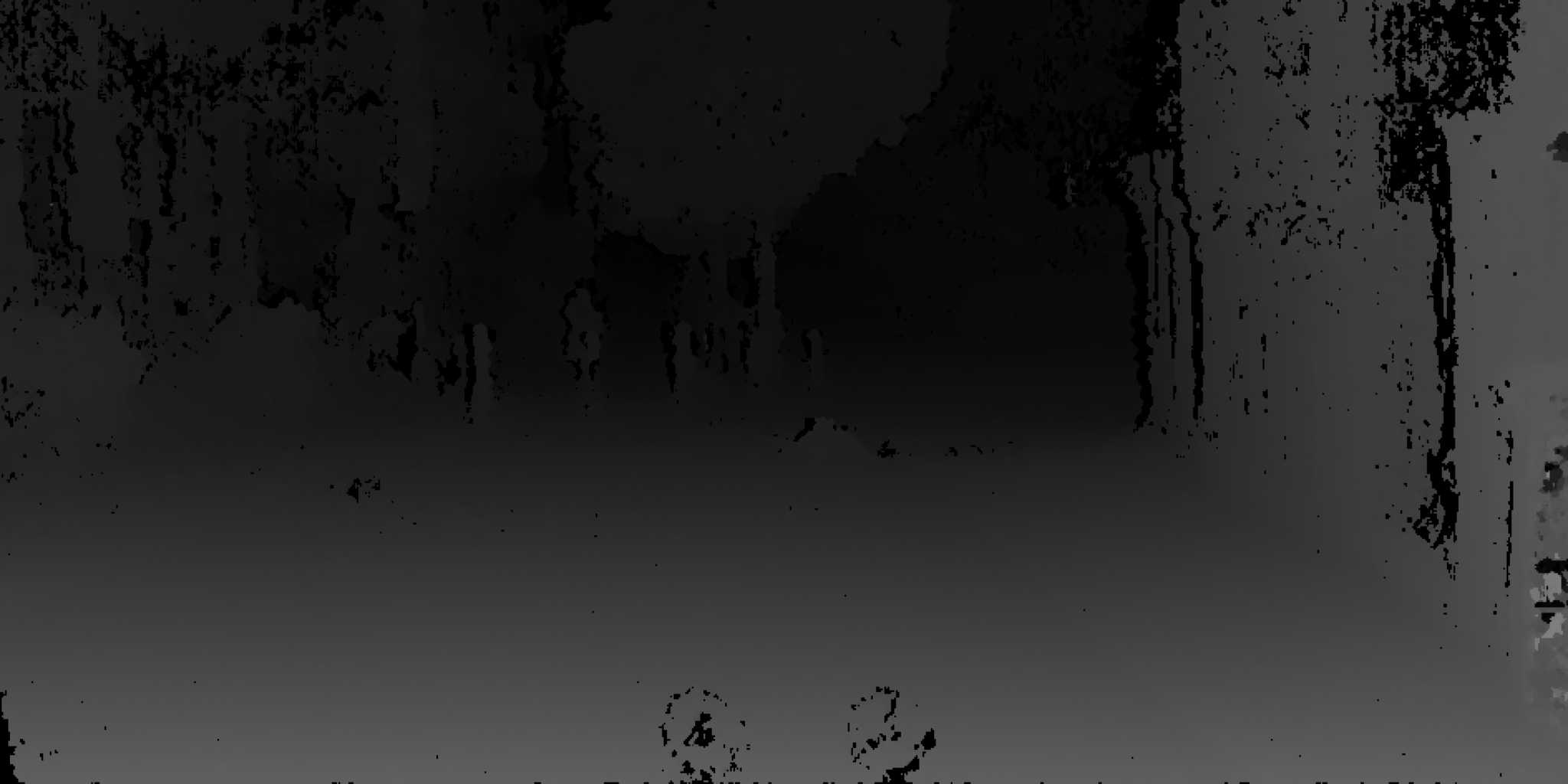}\vspace{1pt}
        \includegraphics[width=1\linewidth]{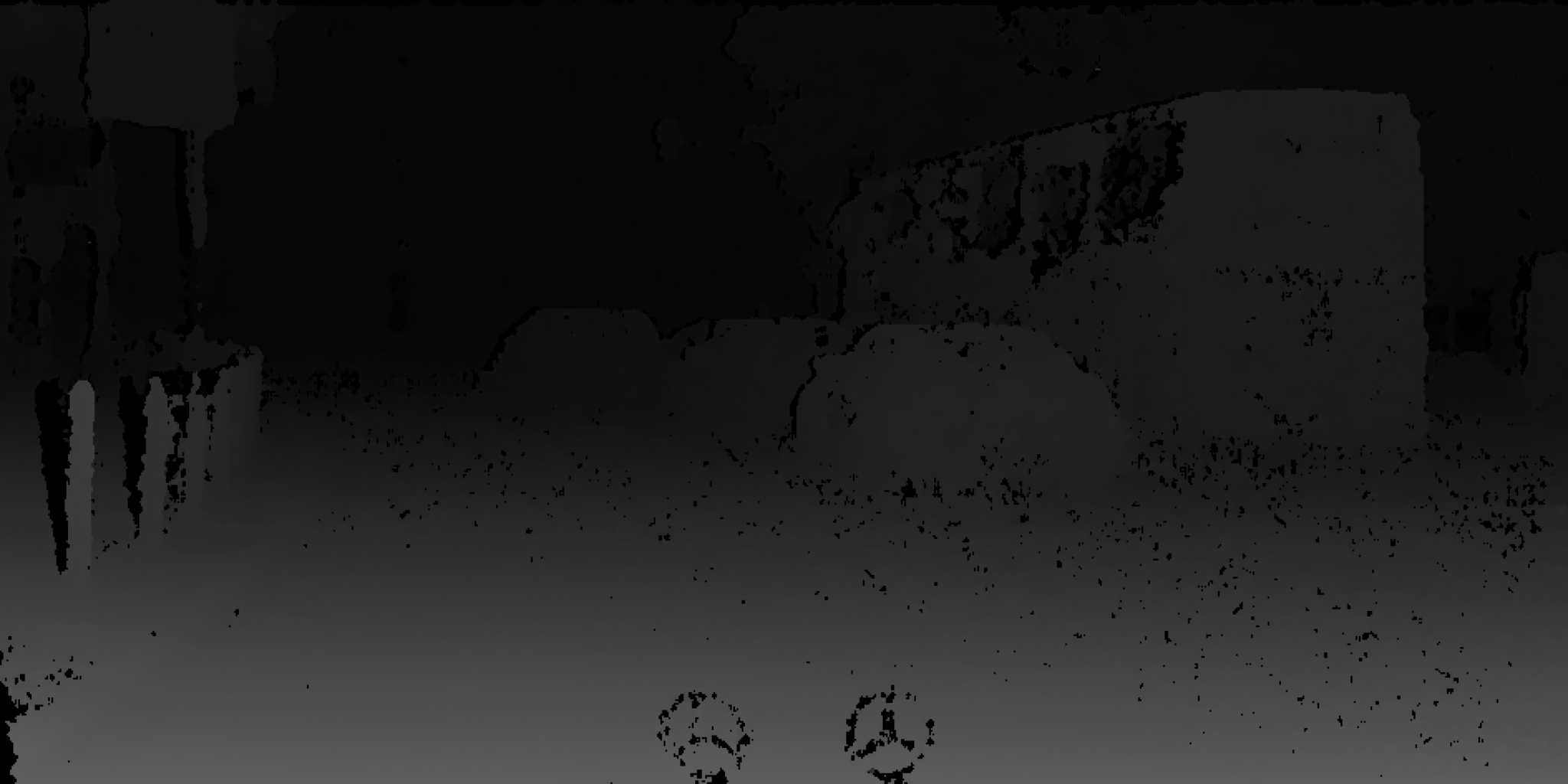}\vspace{1pt}
        \includegraphics[width=1\linewidth]{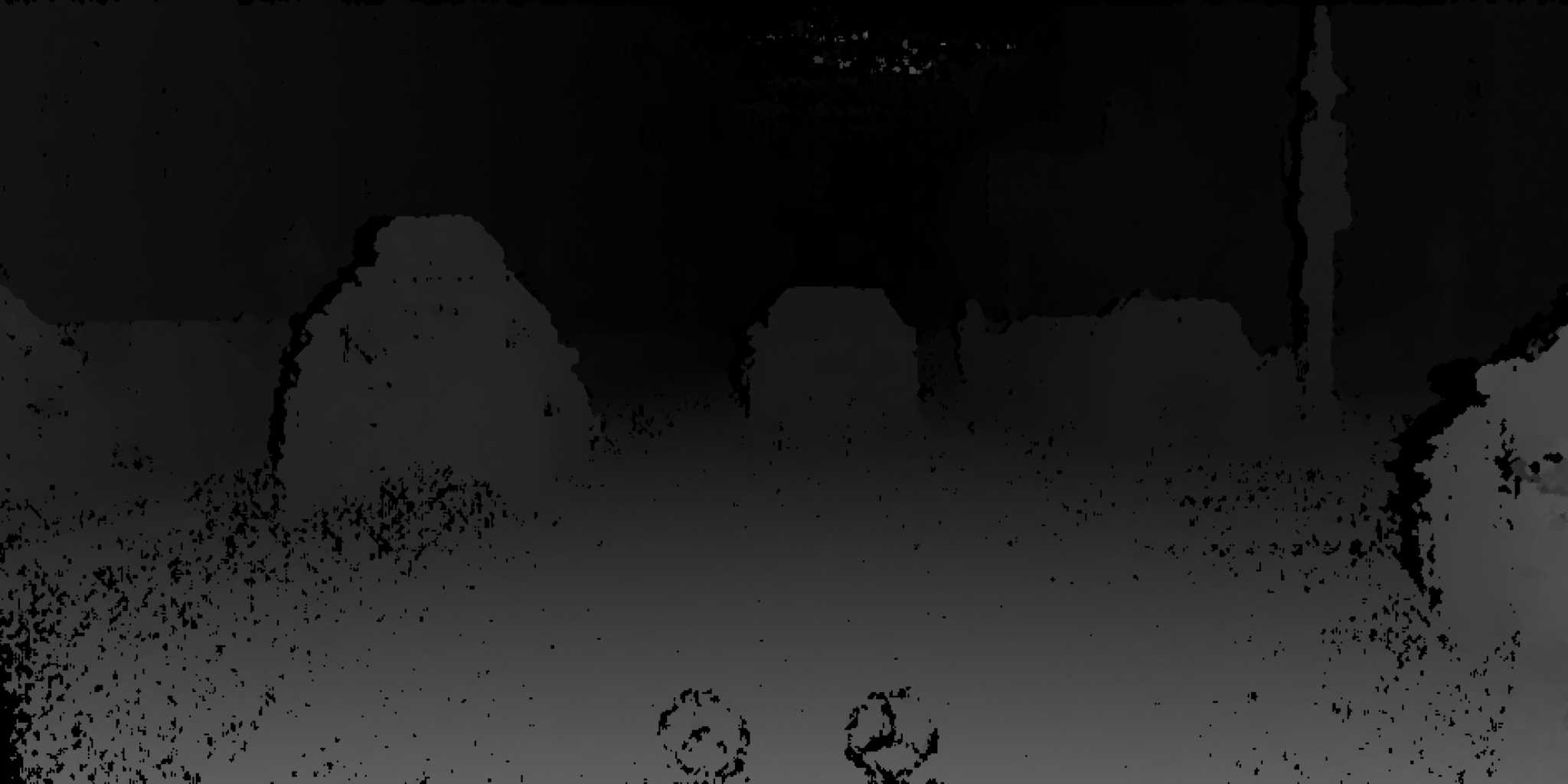}\vspace{1pt}
        \includegraphics[width=1\linewidth]{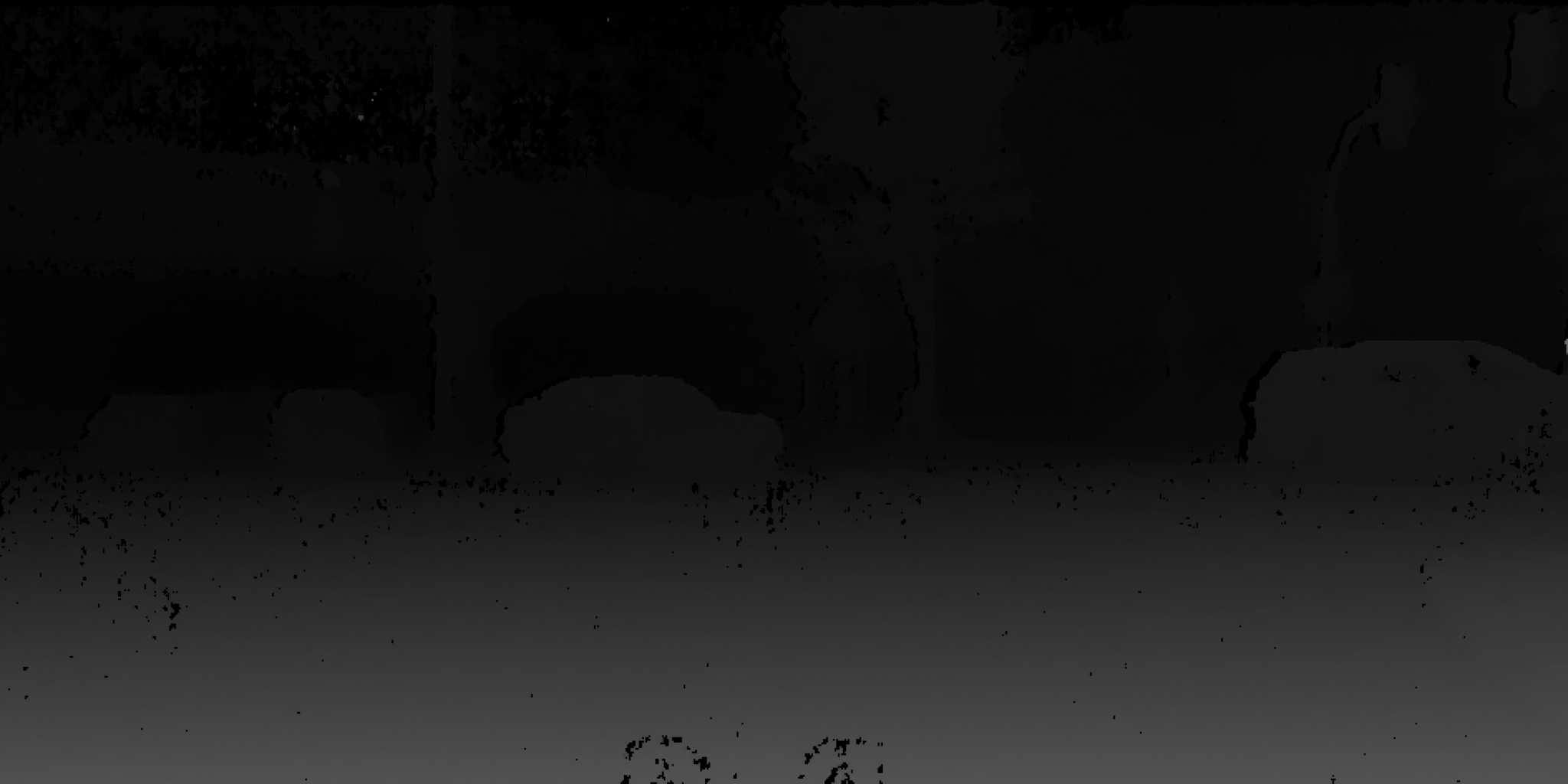}\vspace{1pt}
    \end{minipage}}\hspace{-4pt}  % 控制subfigure之间横向距离
    \subfigure[GT]{
    \begin{minipage}[b]{0.19\linewidth}
        \includegraphics[width=1\linewidth]{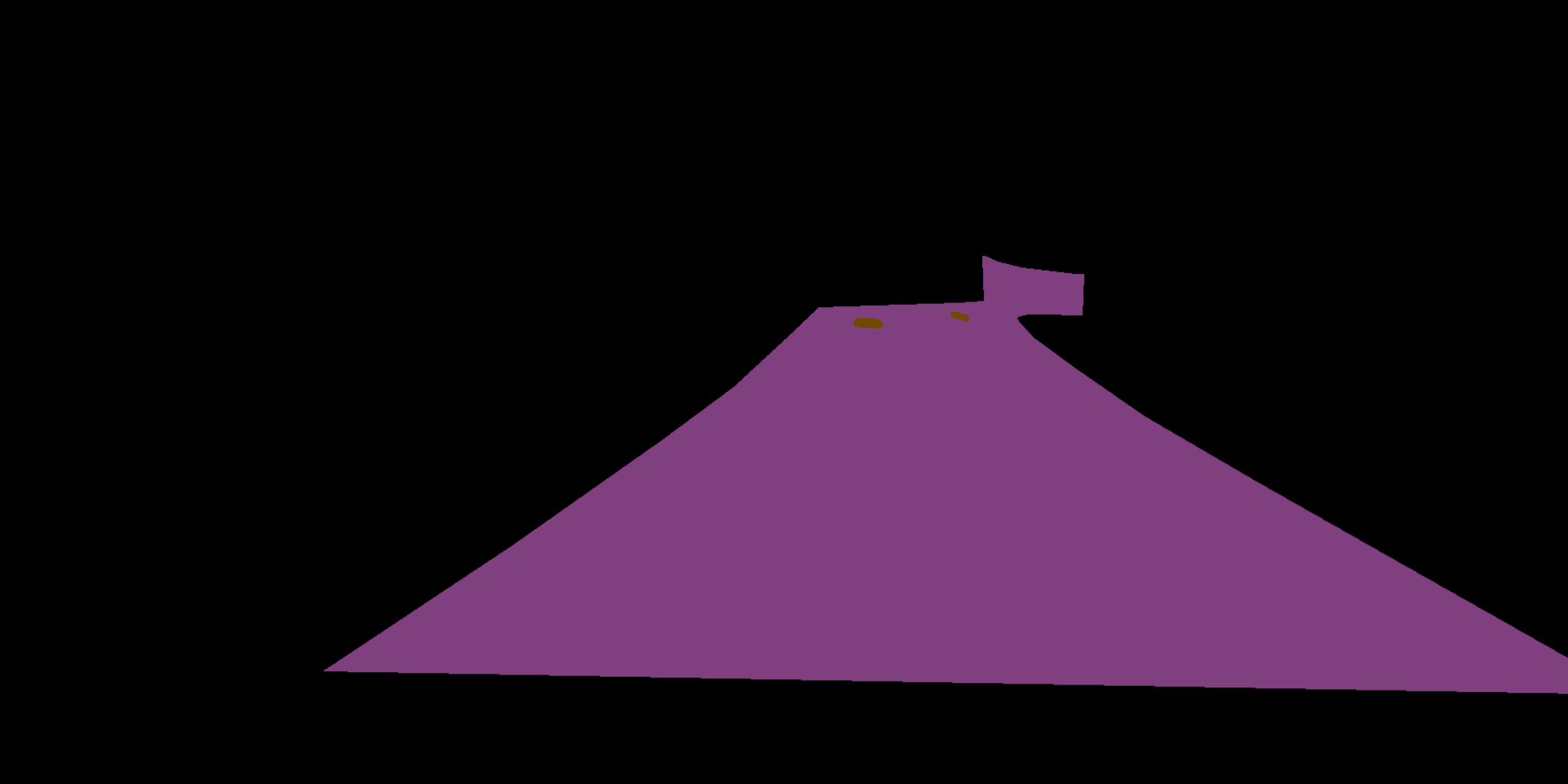}\vspace{1pt}
        \includegraphics[width=1\linewidth]{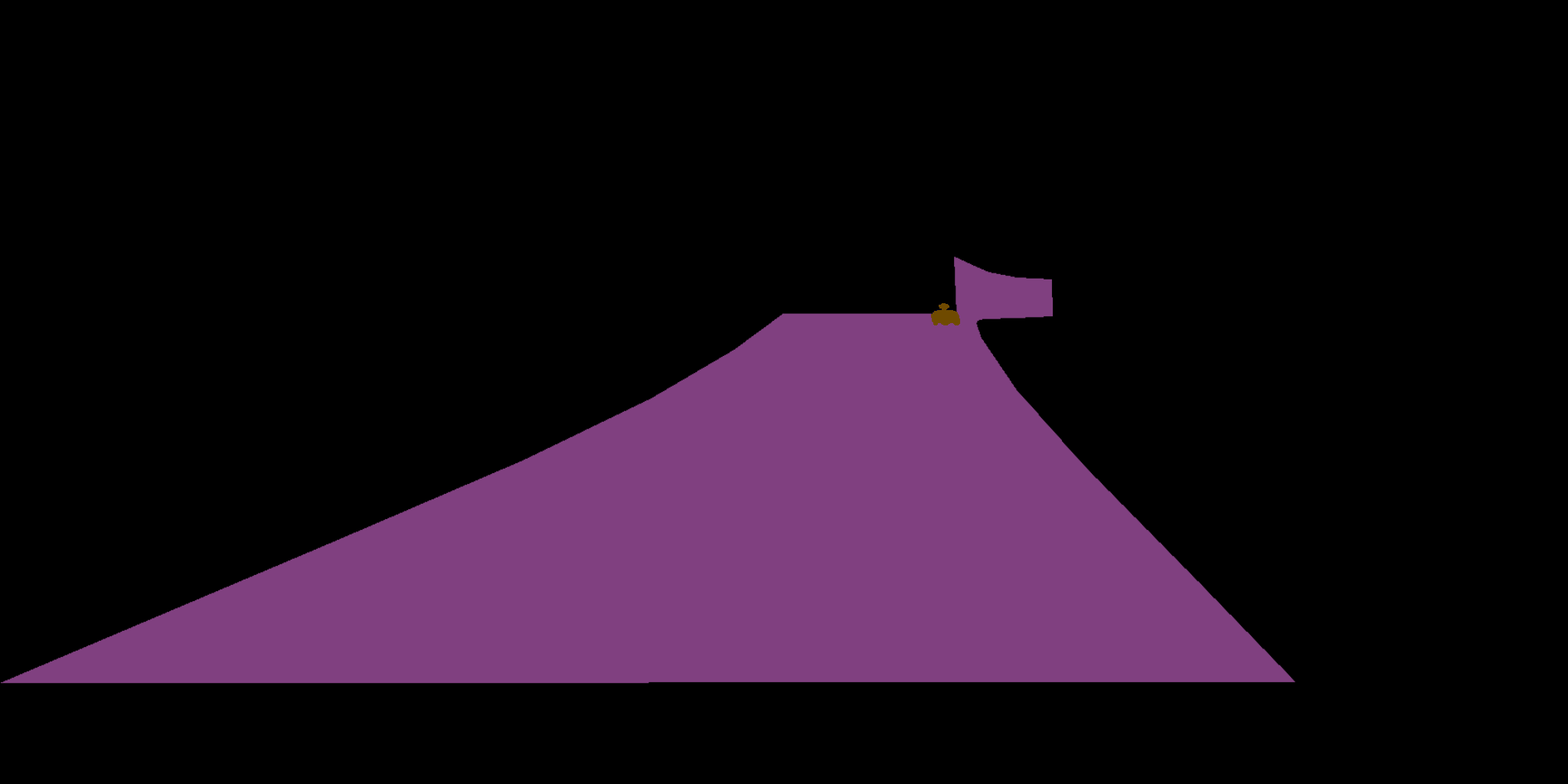}\vspace{1pt}
        \includegraphics[width=1\linewidth]{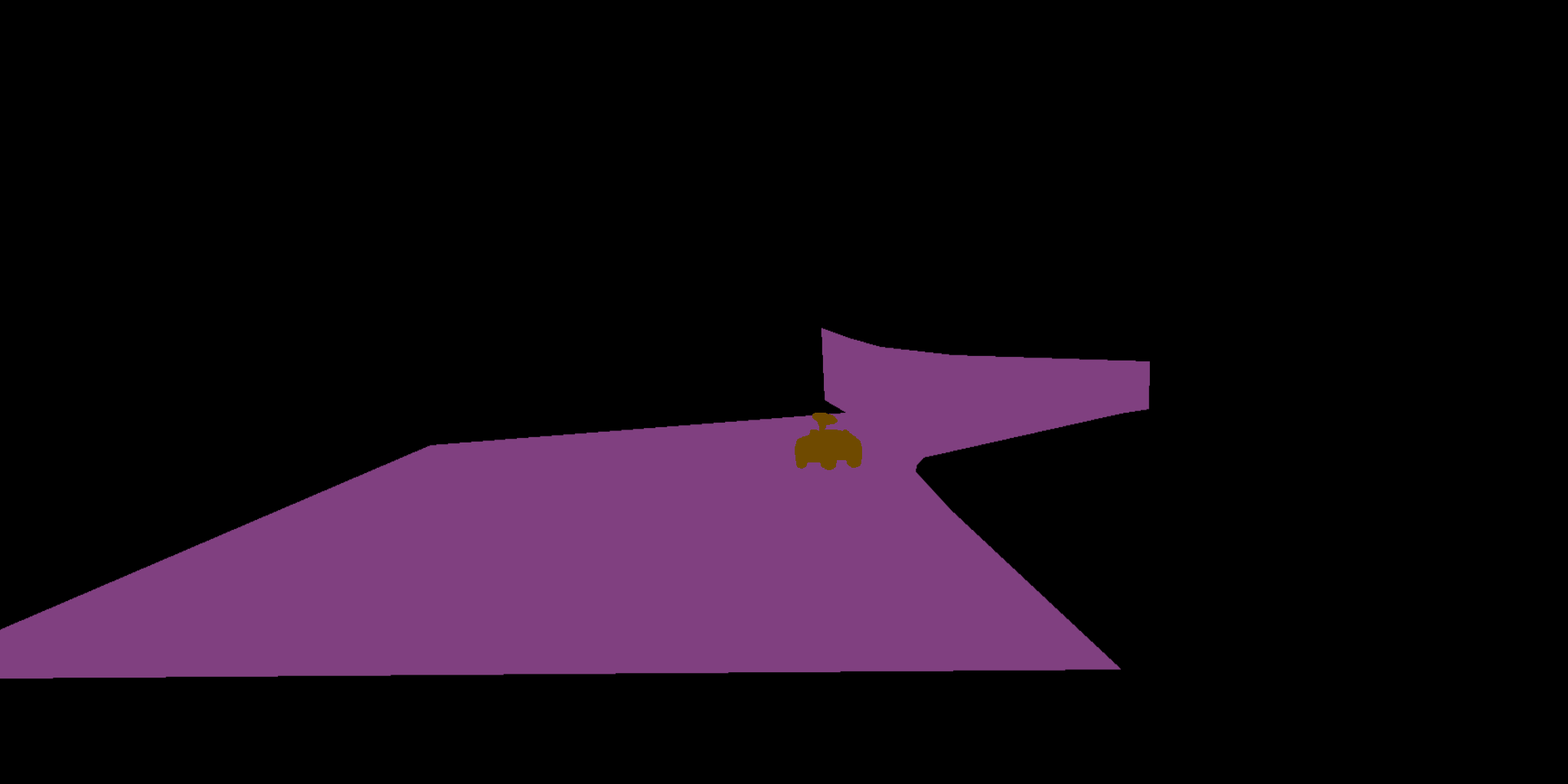}\vspace{1pt}
        \includegraphics[width=1\linewidth]{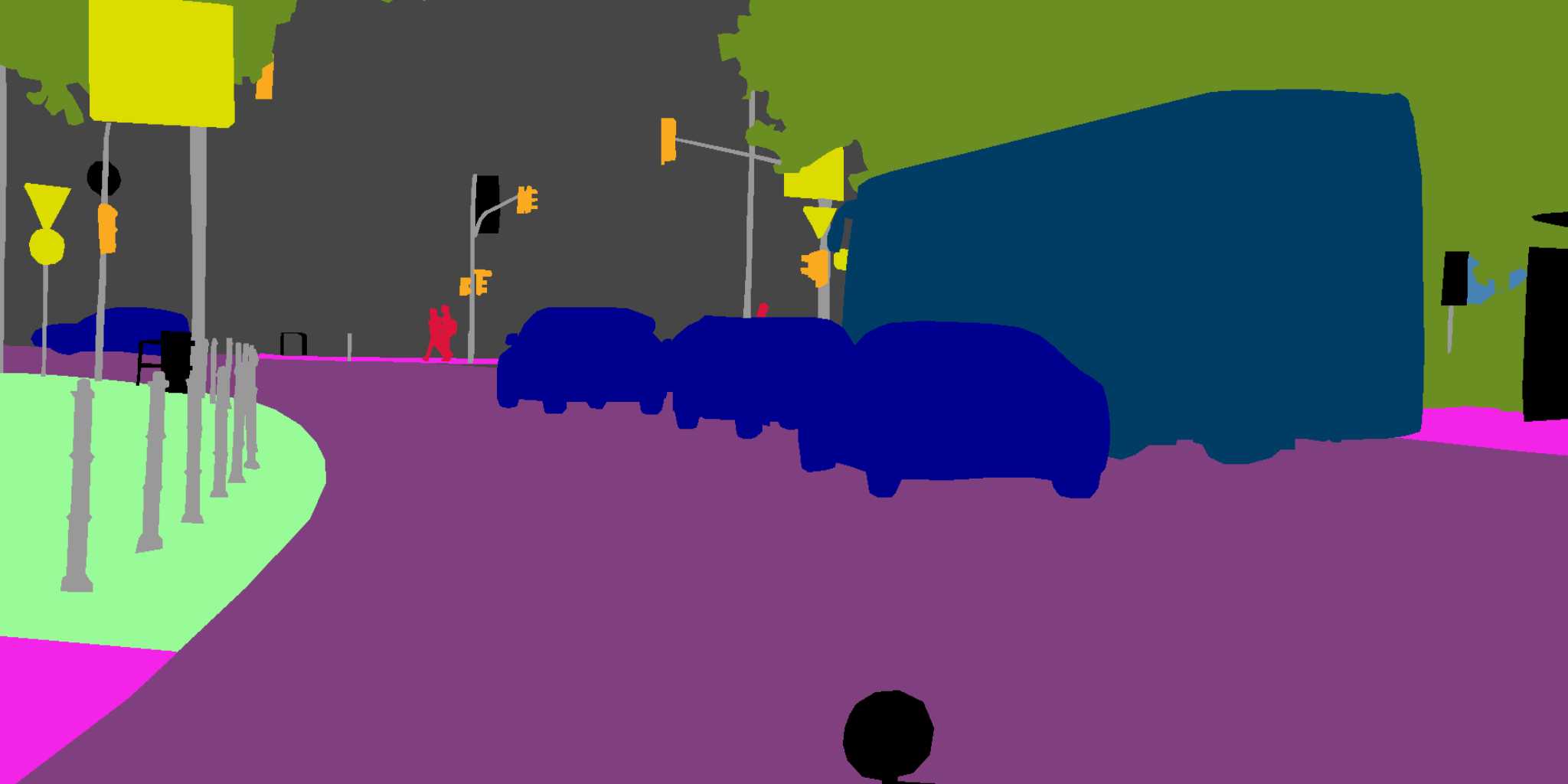}\vspace{1pt}
        \includegraphics[width=1\linewidth]{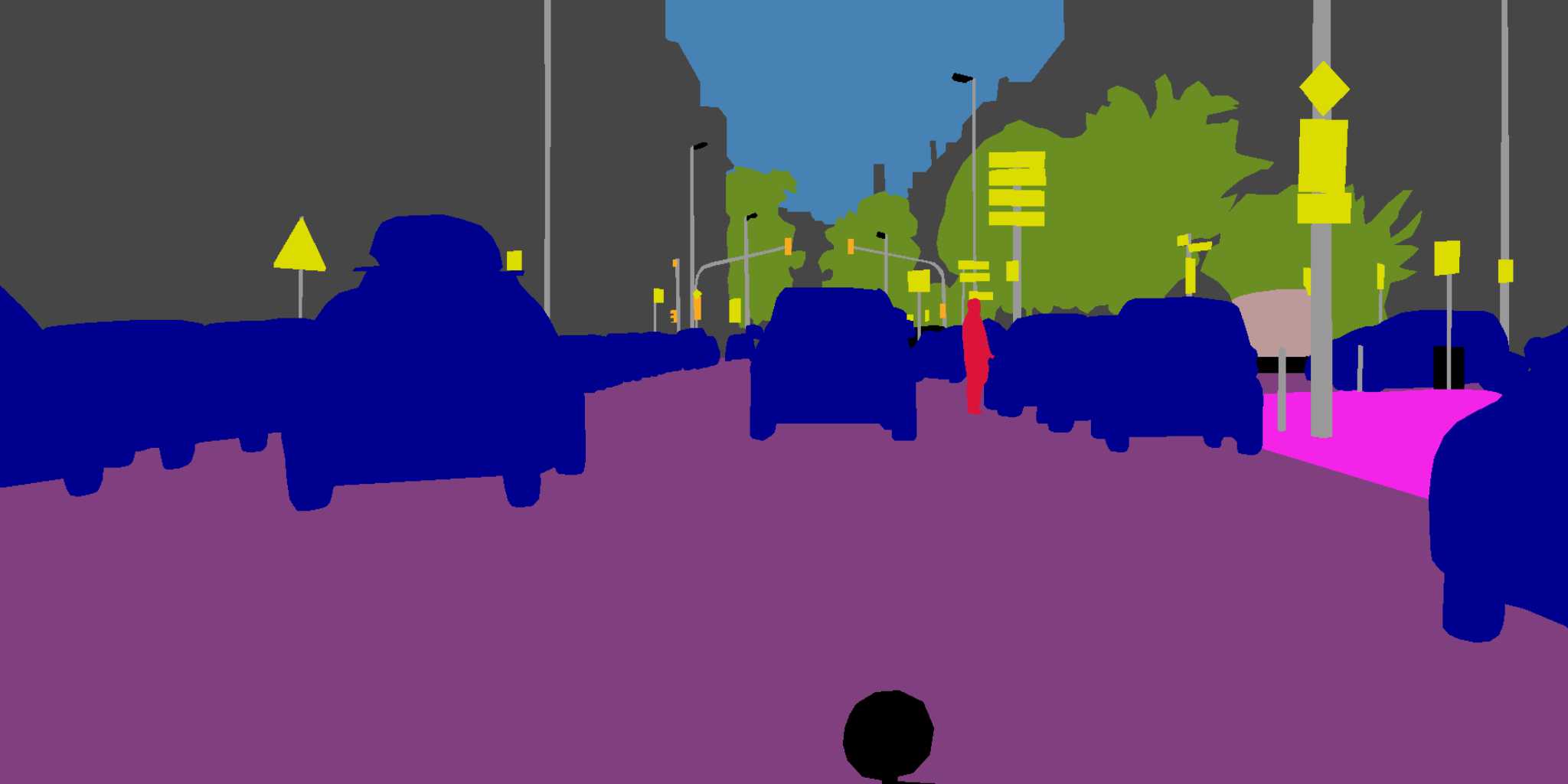}\vspace{1pt}
        \includegraphics[width=1\linewidth]{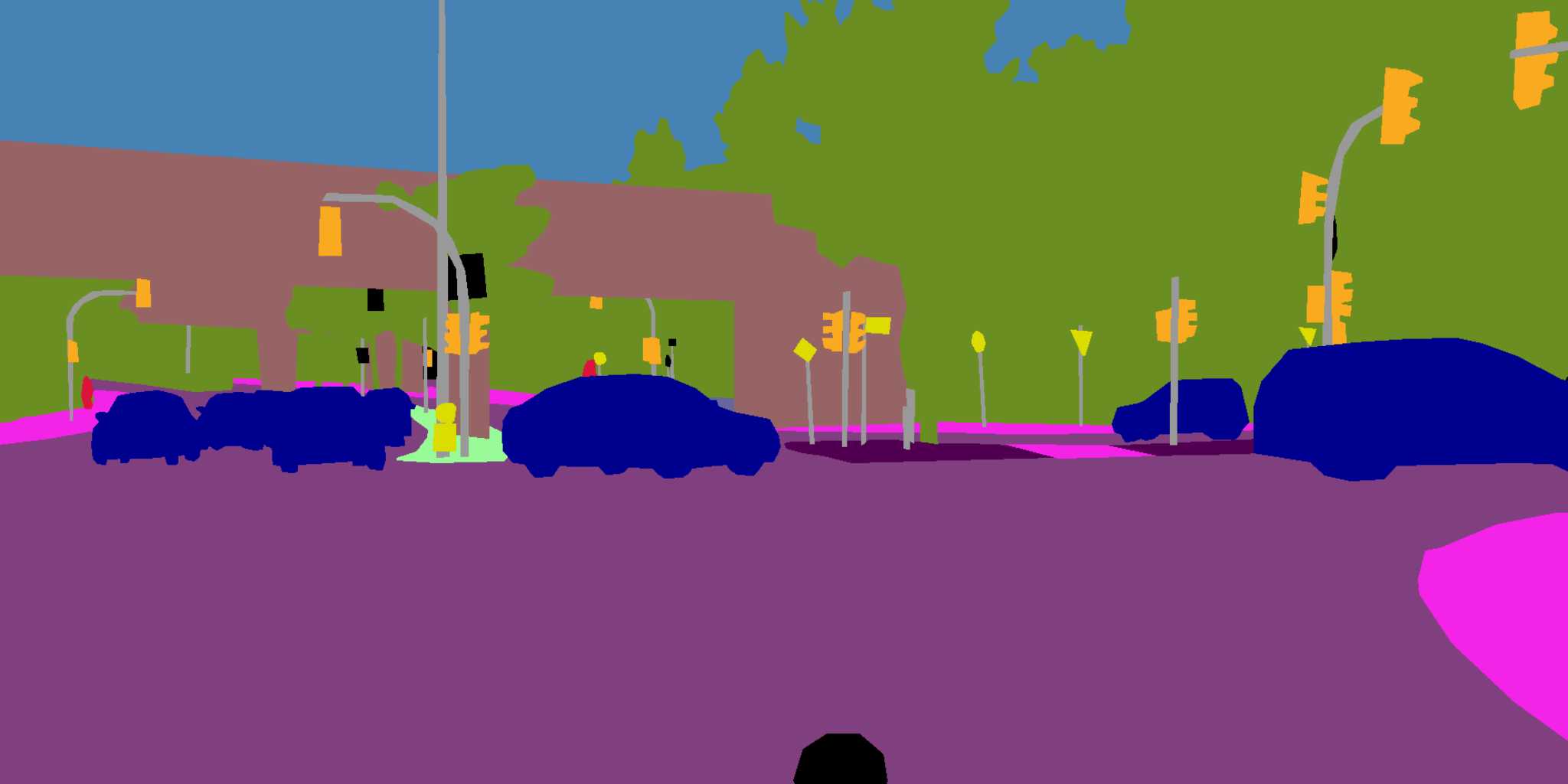}\vspace{1pt}
    \end{minipage}}\hspace{-4pt}  % 控制subfigure之间横向距离
    \subfigure[SwiftNet]{
    \begin{minipage}[b]{0.19\linewidth}
        \includegraphics[width=1\linewidth]{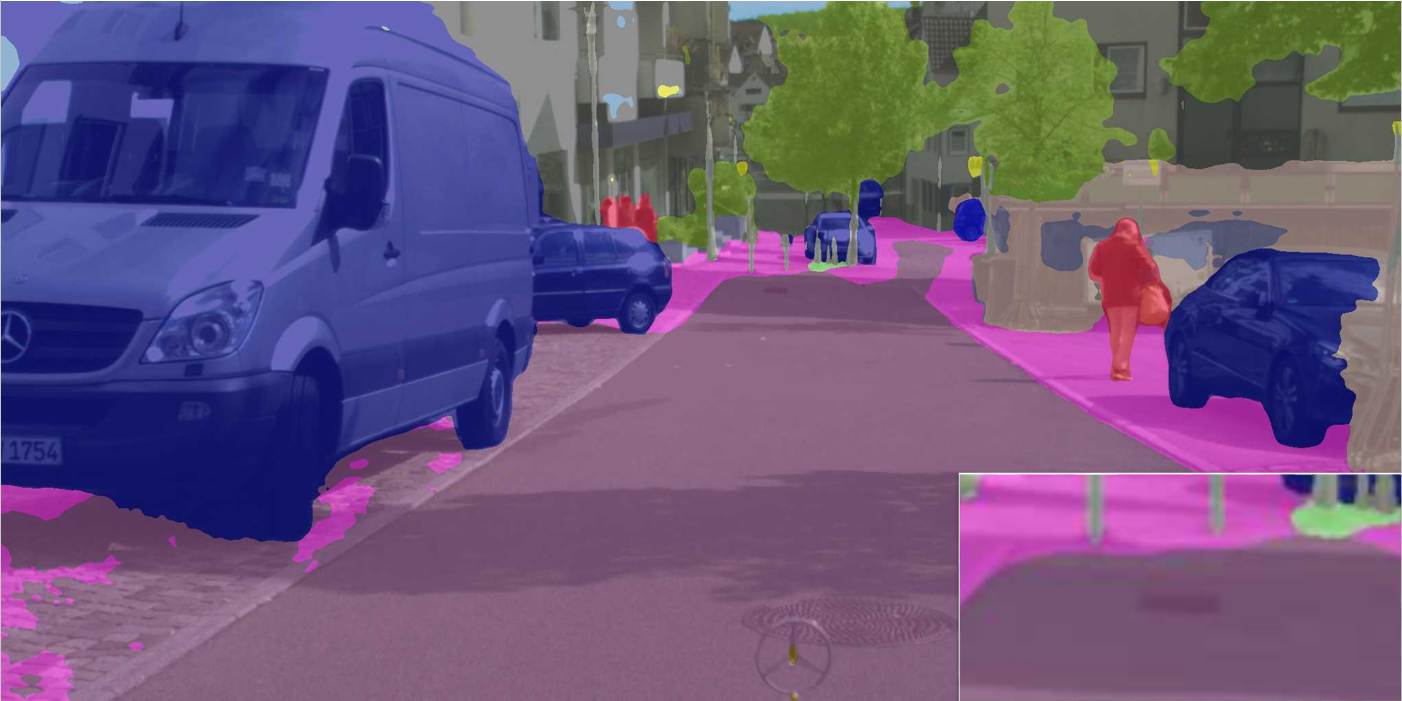}\vspace{1pt}
        \includegraphics[width=1\linewidth]{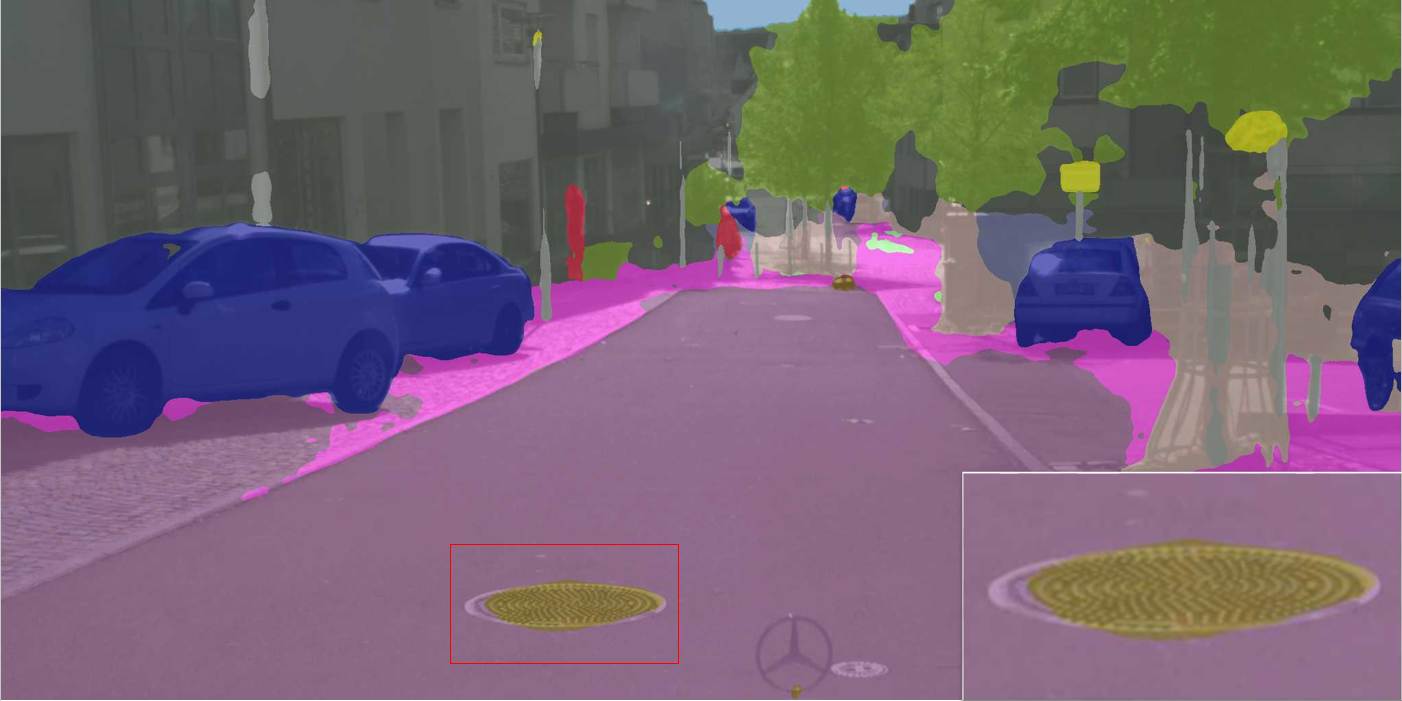}\vspace{1pt}
        \includegraphics[width=1\linewidth]{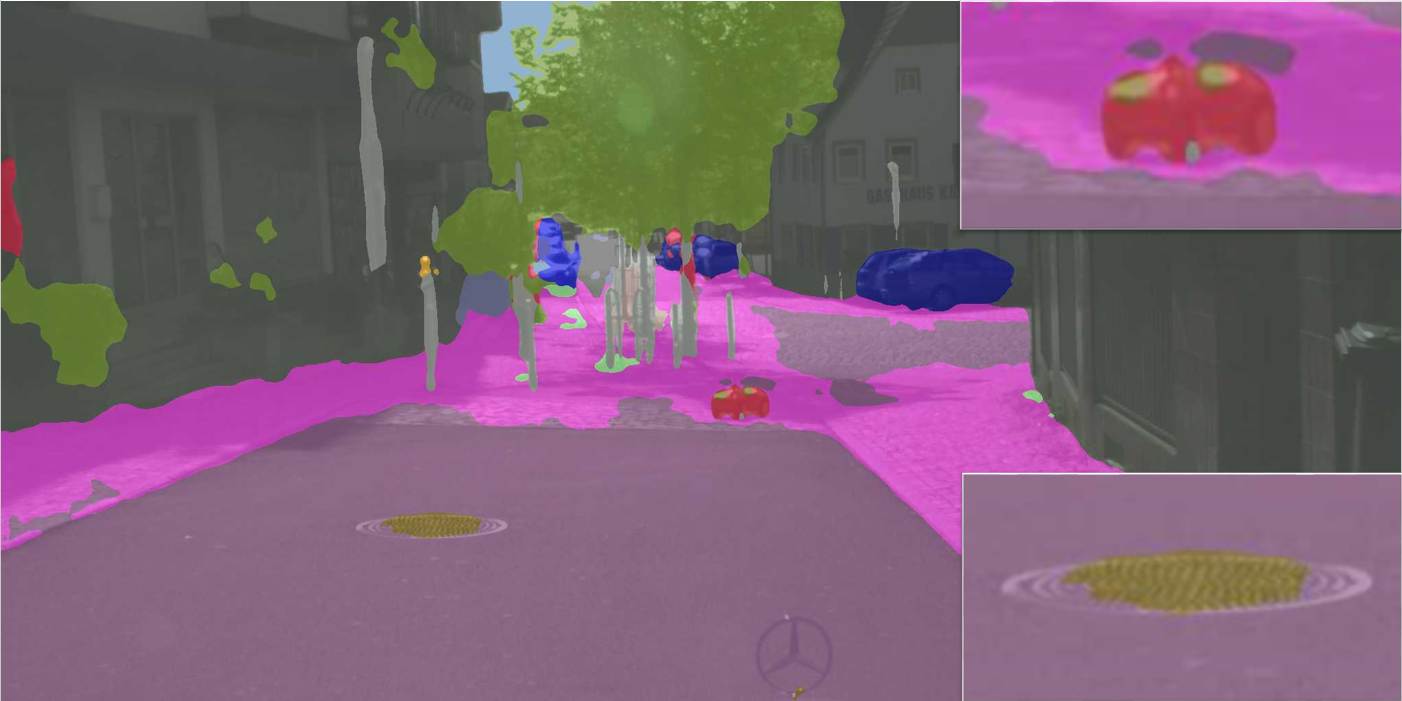}\vspace{1pt}
        \includegraphics[width=1\linewidth]{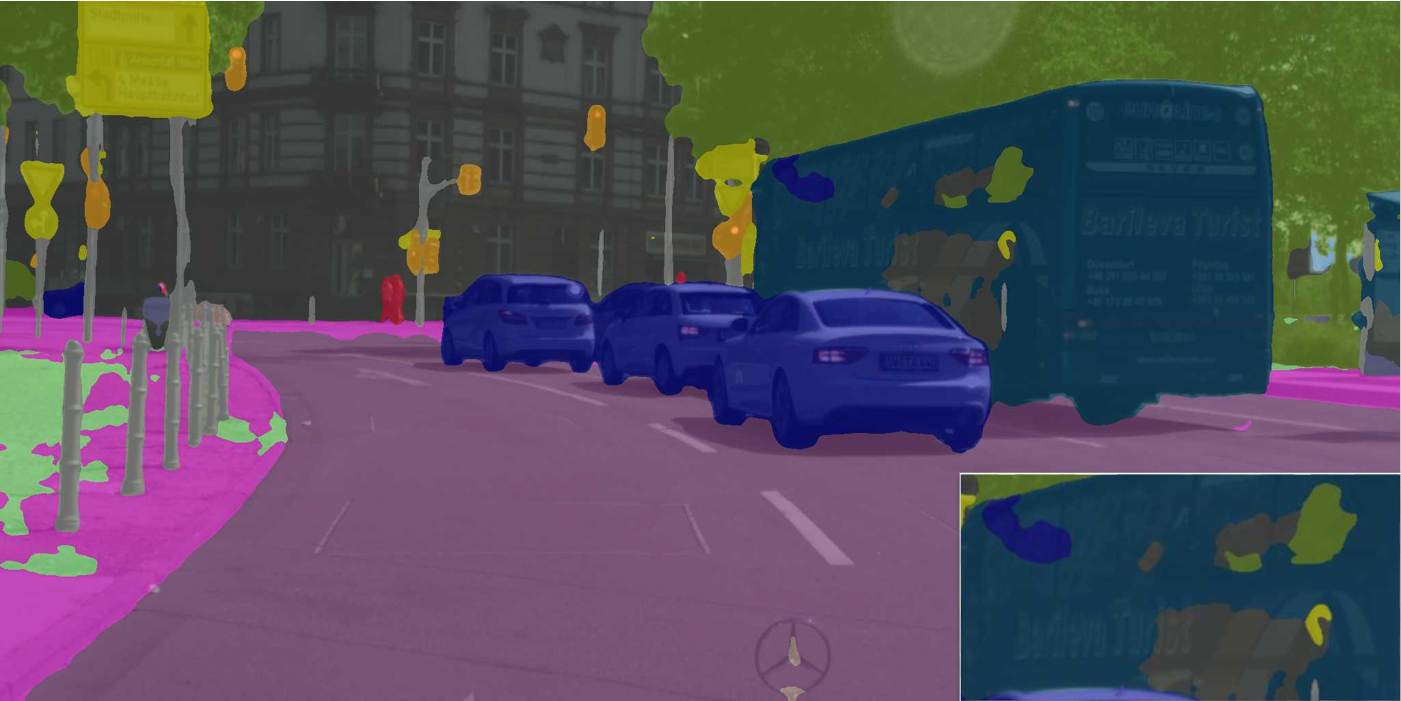}\vspace{1pt}
        \includegraphics[width=1\linewidth]{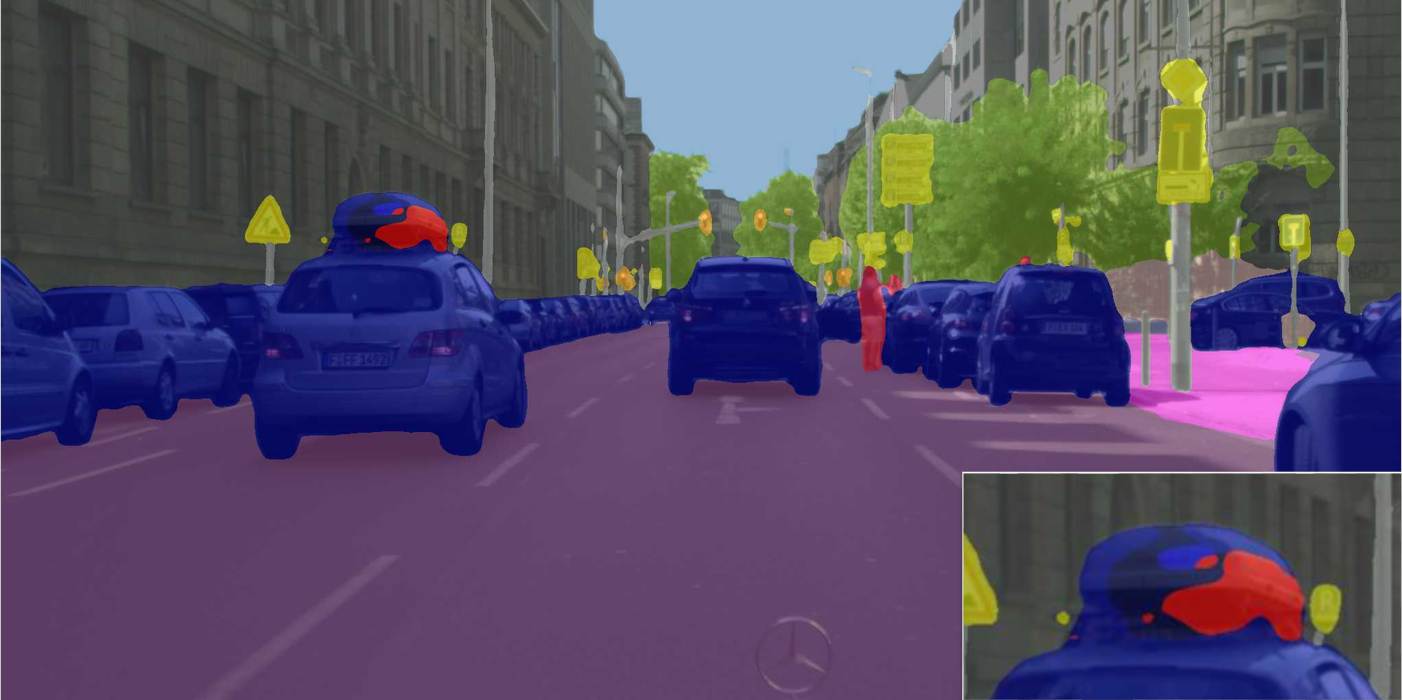}\vspace{1pt}
        \includegraphics[width=1\linewidth]{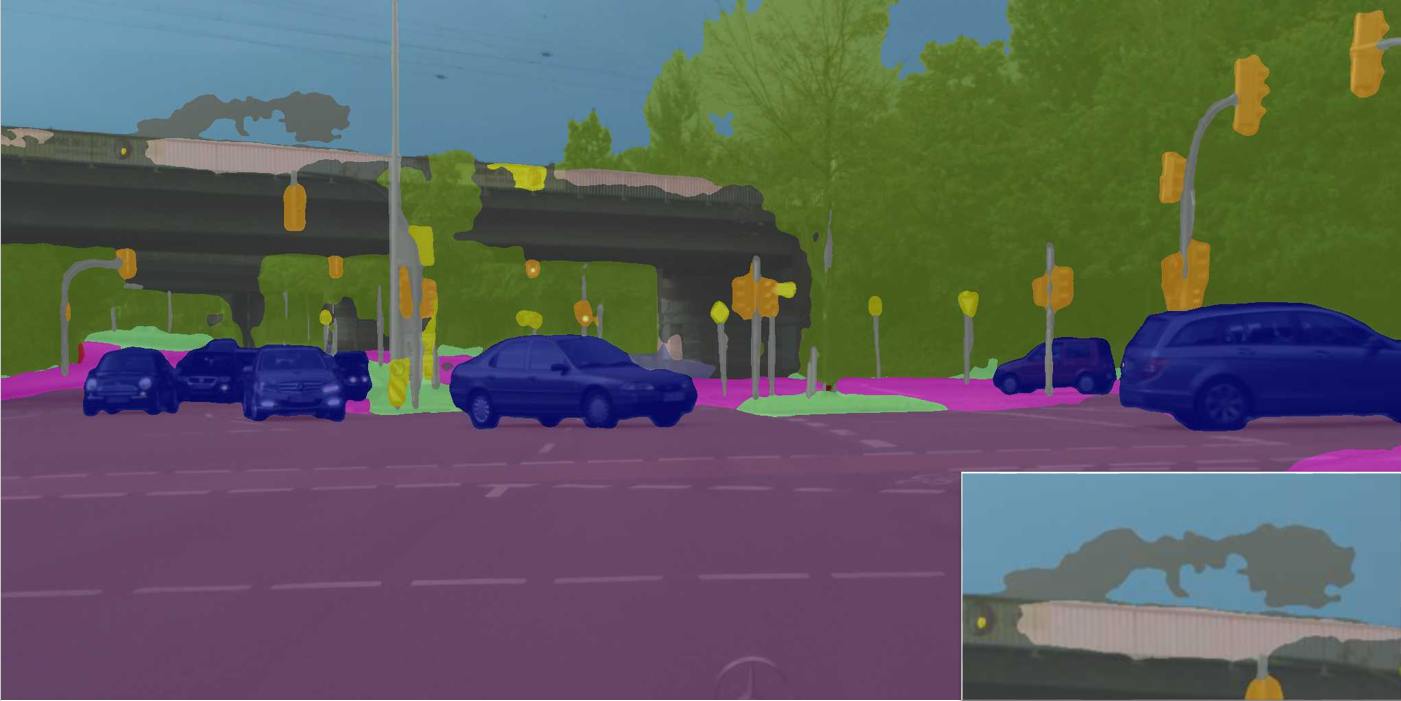}\vspace{1pt}
    \end{minipage}}\hspace{-4pt}  % 控制subfigure之间横向距离
    \subfigure[RFNet]{
    \begin{minipage}[b]{0.19\linewidth}
        \includegraphics[width=1\linewidth]{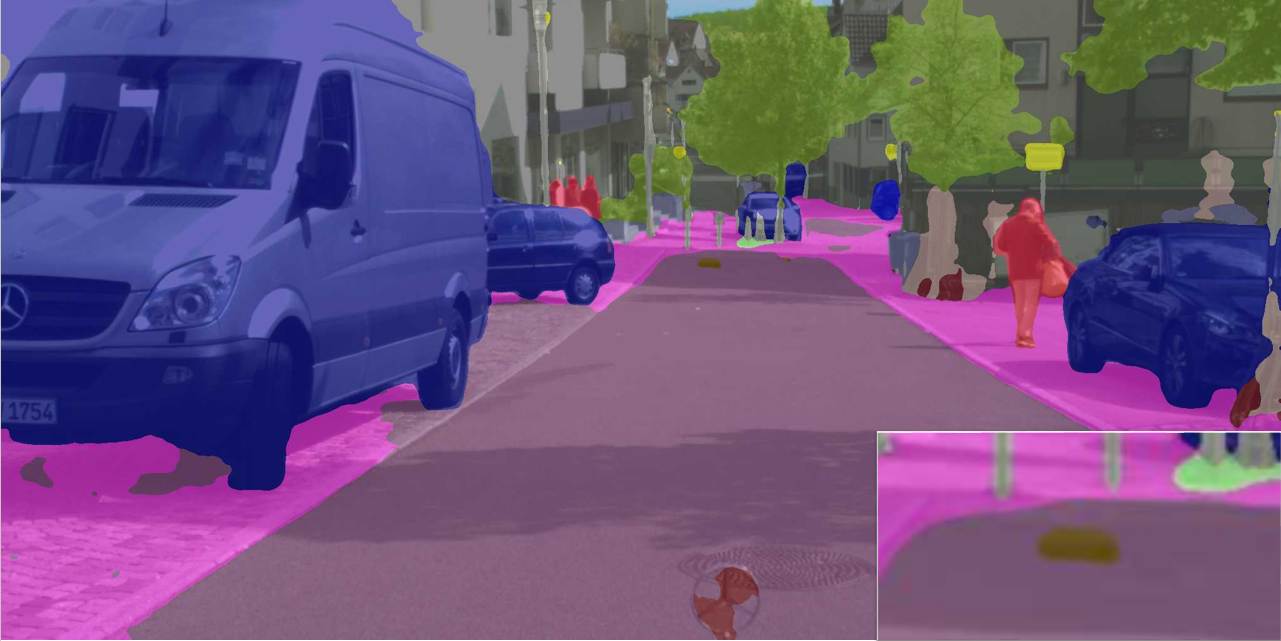}\vspace{1pt}
        \includegraphics[width=1\linewidth]{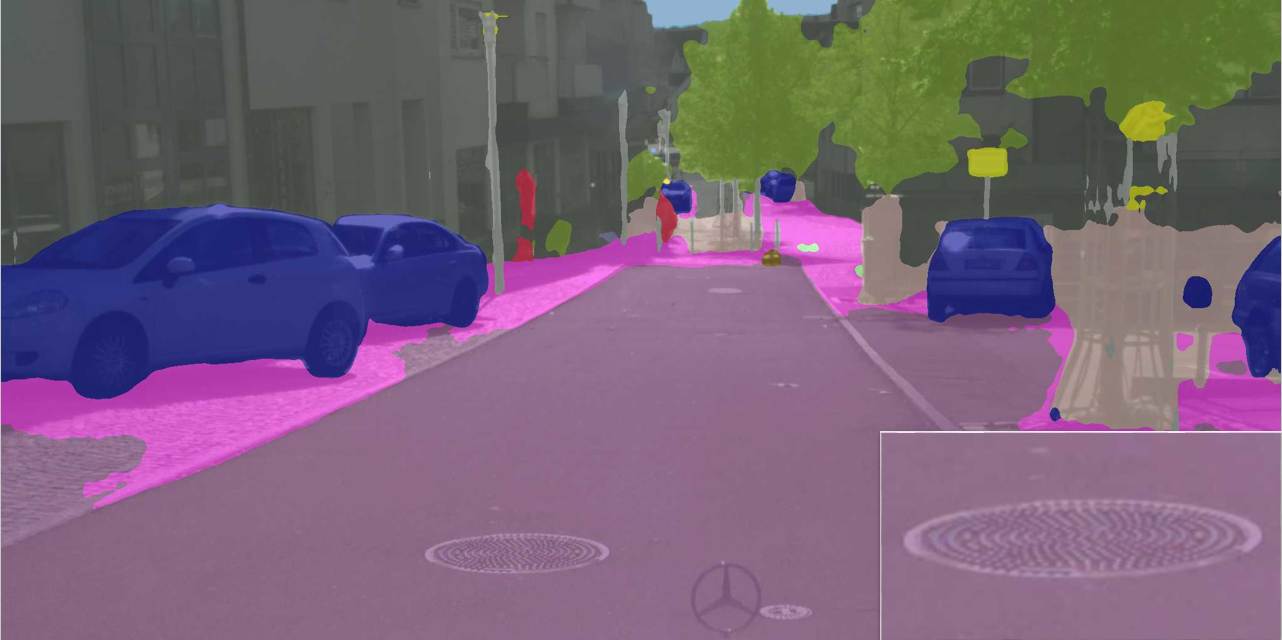}\vspace{1pt}
        \includegraphics[width=1\linewidth]{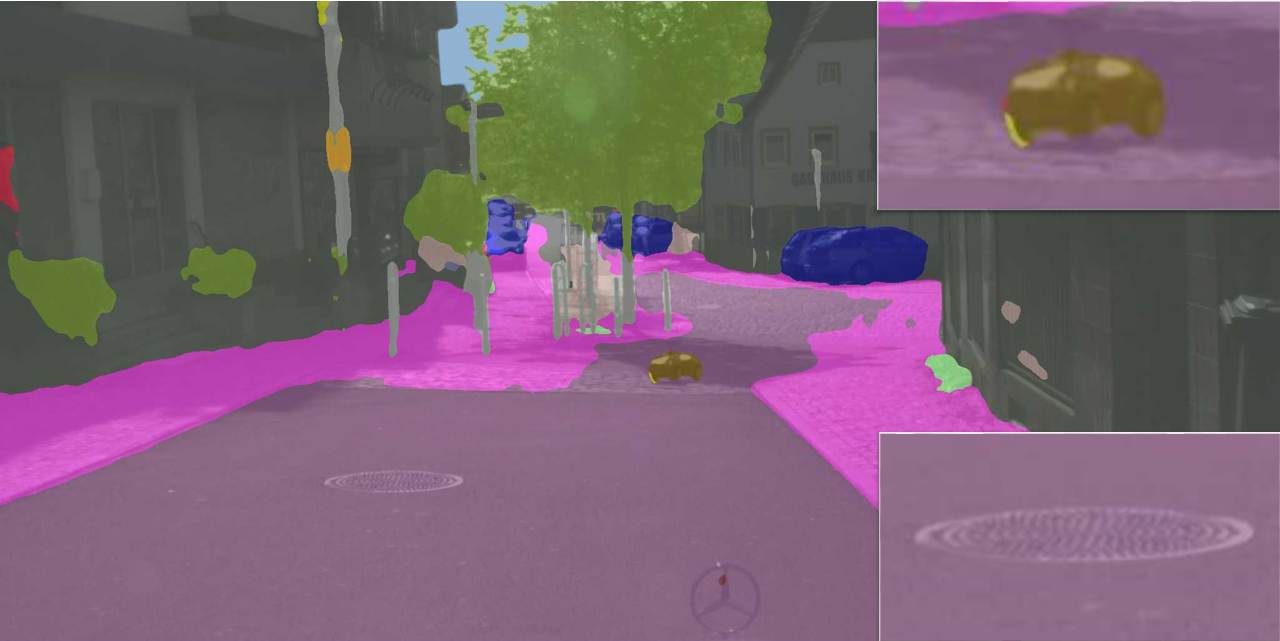}\vspace{1pt}
        \includegraphics[width=1\linewidth]{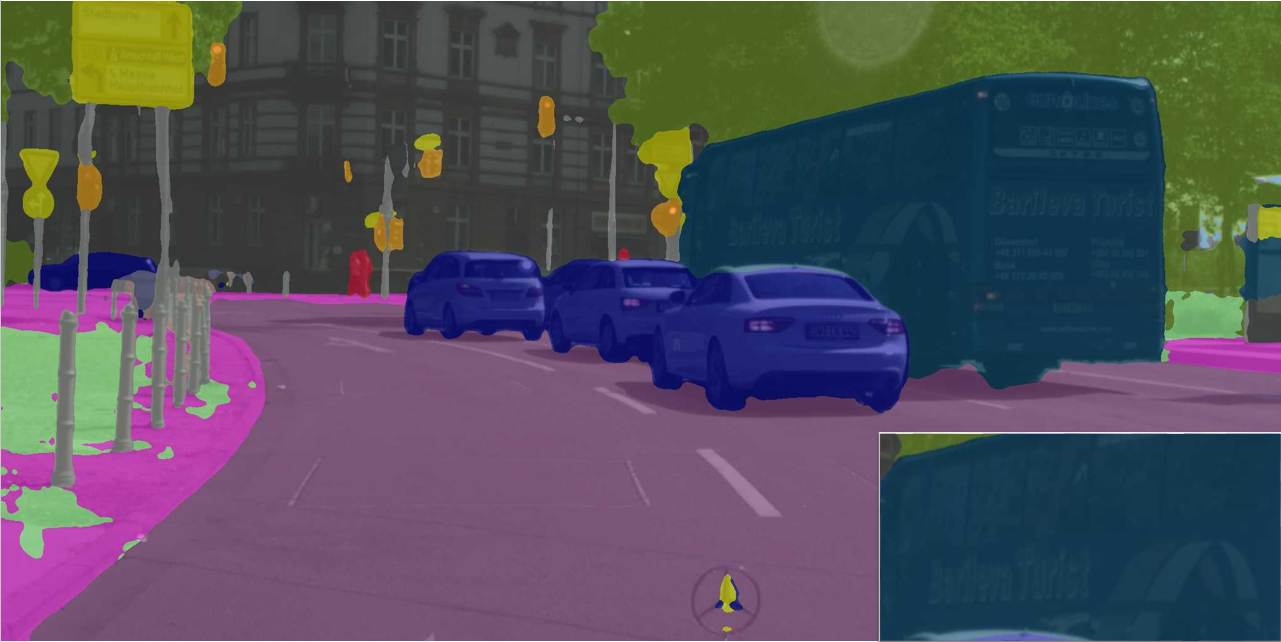}\vspace{1pt}
        \includegraphics[width=1\linewidth]{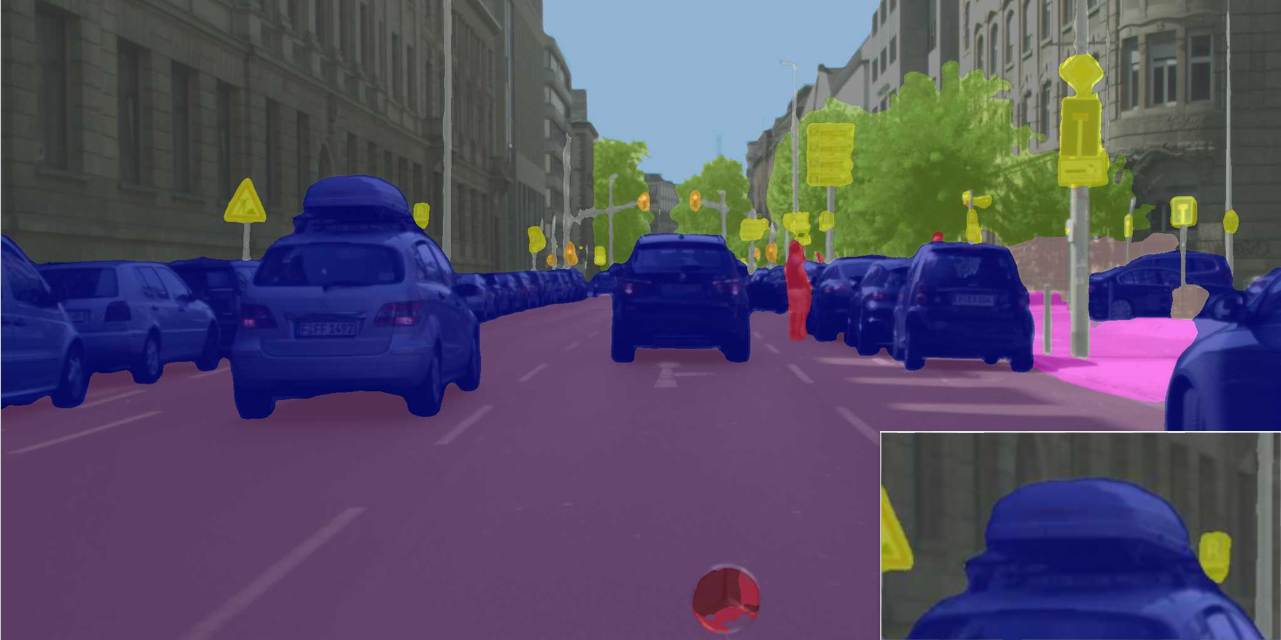}\vspace{1pt}
        \includegraphics[width=1\linewidth]{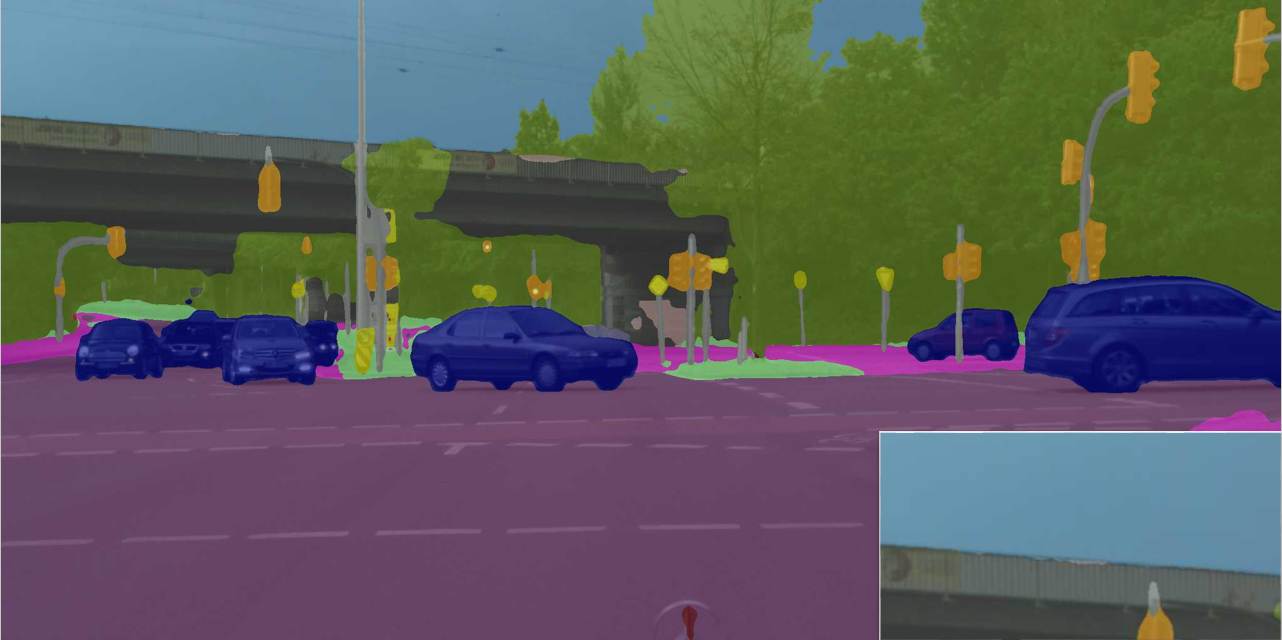}\vspace{1pt}
    \end{minipage}}
    % \caption[contrast_results]
    
    \caption{Qualitative semantic segmentation results from SwiftNet and the proposed RFNet that exploits both RGB and depth information.}
    \label{fig:contrast_results}
\end{figure*}

\textbf{Qualitative Performance Study.}
We present the qualitative examples in Figure~\ref{fig:contrast_results}. In this paper, the main purpose of exploiting depth information is to enhance the segmentation accuracy in classes which are difficult for the RGB method, including small obstacles. The appearance and surface texture of the small obstacles are not fixed, and it is easy to be confused with graffiti, manhole covers, zebra crossings on the road. Comparatively, in the depth map where texture is ignored, the contour of small obstacles is clear. Graffiti, manhole covers are flat, making it part of the road surface in the depth map. All these features of depth maps enable to reduce the chance of false alarm in detecting small obstacles. We compare our RFNet with SwiftNet\cite{orsic2019defense}, which has a similar network architecture with RFNet, where Figure~\ref{fig:contrast_results} shows representative contrast results from the two networks. As a purely RGB-based method, SwiftNet fails to predict some small obstacles on the road and predicts manhole cover as small obstacle. RFNet correctly detect small obstacles and classifies the manhole as part of the road, which demonstrates the superiority of our method for safety-critical road sensing. RFNet also performs better in large-scale objects like bus and truck because contours of these classes are much clearer in depth maps compared to RGB images.

Furthermore, for the input image from Figure~\ref{fig:1}, Figure~\ref{fig:feature_maps} shows the feature maps after the second block from RFNet, in which the first two are feature maps from RGB and depth branch respectively, and the merged feature maps are from the output part of the AFC module. As it can be clearly seen, compared to RGB feature maps, small obstacle is much more clear and manhole cover disappears in depth feature maps, while the feature map after AFC module takes the advantages of both branches. In summary, the AFC module enables RFNet to effectively exploit the depth features in a complementary way, improving the accuracy of obstacle detection evidenced by both numerical and qualitative results.

\section{CONCLUSION}
In this study, we propose RFNet, a real-time fusion network for RGB-D semantic segmentation on road-driving images. With the designed AFC module, RFNet exploits complementary depth information effectively and significantly improves the accuracy over purely RGB-based methods. With the presented multi-source training strategy, RFNet can also detect unexpected small obstacles, enriching the recognizable classes required to face the real world with unforeseen hazards. More importantly, RFNet operates at 22Hz with full resolution \textit{Cityscapes} images and 41.6Hz with half resolution on a single Nvidia GTX2080Ti GPU, which makes it ideally suitable for autonomous driving applications. Our RFNet outperforms state-of-the-art RGB-D fusion methods in terms of accuracy and speed. In the future, we plan to further streamline RFNet and deploy it to portable TPU devices with robustness augmented. The source code of our RFNet is available at {\tt https://github.com/AHupuJR/RFNet}.

%\addtolength{\textheight}{-12cm}
\bibliographystyle{IEEEtran}
\bibliography{IEEEabrv,reference}

\end{document}